\documentclass{article}

\usepackage[preprint]{neurips_2026}

\usepackage[utf8]{inputenc} 
\usepackage[T1]{fontenc}    
\usepackage{hyperref}       
\usepackage{url}            
\usepackage{booktabs}       
\usepackage{amsfonts}       
\usepackage{nicefrac}       
\usepackage{microtype}      
\usepackage{xcolor}         
\usepackage{graphicx}
\usepackage{xltabular}
\usepackage{amsmath} 
\usepackage{multirow}
\usepackage{pdflscape}

\title{Universal CT Representations from Anatomy to Disease Phenotype through Agglomerative Pretraining}

\author{%
\begin{minipage}{0.95\textwidth}
\centering
Yuheng Li$^{1,\dagger}$,
Yuan Gao$^{2,\dagger}$,
Haoyu Dong$^{3}$,
Yuxiang Lai$^{4}$,
Shansong Wang$^{2}$,
Mojtaba Safari$^{2}$,
James E. Baciak$^{5,*}$,
Xiaofeng Yang$^{1,2,4,*}$
\vspace{0.7em}
{\small\normalfont
\\
$^{1}$Wallace H. Coulter Department of Biomedical Engineering, 
Georgia Institute of Technology and Emory University, Atlanta, GA 30332, USA
\\
$^{2}$Department of Radiation Oncology and Winship Cancer Institute, 
Emory University, Atlanta, GA 30322, USA
\\
$^{3}$Department of Electrical and Computer Engineering, 
Duke University, Durham, NC 27705, USA
\\
$^{4}$Department of Computer Science and Informatics, 
Emory University, Atlanta, GA 30322, USA
\\
$^{5}$Department of Materials Science \& Engineering, Nuclear Engineering Program, 
University of Florida, Gainesville, FL 32611, USA
\\
}
\vspace{0.7em}
{\small\normalfont
$^{\dagger}$These authors contributed equally to this work.\\
$^{*}$Corresponding authors: 
\texttt{jebaciak@mse.ufl.edu}, 
\texttt{xiaofeng.yang@emory.edu}
}
\end{minipage}
}

\begin{document}

\maketitle

\begin{abstract}
Computed tomography (CT) is a central to three-dimensional medical imaging, yet CT-based artificial intelligence remains fragmented across task-specific models for segmentation, classification, registration, and report analysis. Here we present FlexiCT, a family of CT foundation models trained by agglomerative continual pretraining on 266,227 CT volumes from 56 publicly available datasets, forming a large-scale public resource for CT representation learning. FlexiCT uses agglomerative pretraining across three stages: two-dimensional axial pretraining, three-dimensional anatomical pretraining and report-guided semantic alignment. This training strategy supports slice-level, volume-level and vision-language analysis. Across five downstream task families (segmentation, classification, registration, vision-language understanding and clinical retrieval), FlexiCT matches or exceeds prior task-specific approaches on multiple benchmarks. Its embeddings further organize CT scans along gradients associated with various tumor stages, suggesting that CT foundation models can capture imaging features relevant to disease phenotype characterization. Project page and code are available at: https://ricklisz.github.io/flexict.github.io and https://github.com/ricklisz/FlexiCT.
\end{abstract}

\section{Introduction}

\label{sec:introduction}
Computed tomography (CT) is the workhorse of modern diagnostic imaging, with more than 90 million examinations performed annually in the United States alone and growing use globally~\cite{smith2019trends, brenner2007computed}. CT plays a key role across clinical decision making, supporting emergency triage, oncologic staging, treatment planning and longitudinal monitoring~\cite{power2016computed}. Clinical CT interpretation spans multiple representational levels, from anatomical localization and abnormality detection to disease characterization, severity assessment, and treatment-relevant evaluation. Yet most CT artificial intelligence (AI) models are optimized for only a single level of this hierarchy, addressing anatomical segmentation~\cite{wasserthal2023totalsegmentator,nnunet2021}, abnormality classification~\cite{ardila2019end}, deformable registration~\cite{voxelmorph}, or report-aligned representation learning~\cite{ctclip2024, merlin2024} as separate problems.  This fragmentation has concrete consequences in practice: a representation that is useful for anatomical matching or registration may not support disease characterization, while one aligned to clinical semantics may not preserve the volumetric structure needed for correspondence or retrieval. Clinical CT interpretation moves fluidly across abstraction levels (from anatomy through pathology to severity assessment), but the AI systems intended to support it do not.

Foundation models offer a promising path toward more general CT representations~\cite{moor2023foundation, willemink2022toward}. Recent self-supervised CT pretraining approaches, including VoCo~\cite{voco2024}, CT-FM~\cite{harvardctfm2024}, and SPECTRE~\cite{SPECTRE2024}, have shown that large-scale learning on CT volumes can produce robust anatomy-centric features for segmentation, retrieval, and other dense visual tasks. In parallel, report-aligned models such as CT-CLIP~\cite{ctclip2024} and Merlin~\cite{merlin2024} demonstrate that paired CT-report data can inject clinical semantics into learned embeddings, while broader radiology foundation models such as Curia~\cite{curia2024} and lesion-focused biomarker studies~\cite{pai2024foundation} suggest that pretrained imaging features can generalize across modalities and correlate with tumor biology. These efforts parallel progress in pathology~\cite{chen2024towards,vorontsov2024foundation} and ophthalmology~\cite{zhou2023foundation}, where domain-specific foundation models have begun to translate into clinical utility. However, prior CT efforts remain separated by both objective and scope. Anatomy-centric models are rarely evaluated for clinically meaningful severity structure, whereas report-aligned models are typically developed on narrower paired subsets and less validated on volumetric anatomical tasks. Whether a single coherent foundation-model family can unify anatomical understanding, volumetric reasoning, report-aligned semantics, and clinically meaningful severity structure therefore remains unresolved. This question has only recently become tractable, owing to three converging developments: the availability of large-scale public CT datasets with broad anatomical coverage~\cite{wasserthal2023totalsegmentator,ctclip2024,merlin2024}, self-supervised frameworks that operate without label-centric supervisions~\cite{caron2021emerging, dinov2, dinov3}, and paired CT and report datasets of sufficient scale to enable vision and language alignment.

The central challenge is not data scale alone, but how CT representations should be accumulated from learning signals that differ in spatial density, semantic abstraction, and data availability. Dense anatomical information, three-dimensional spatial continuity, and report-grounded disease concepts each provide distinct information, and a single pretraining stage may not capture them equally well. We therefore hypothesize that CT representations built through progressive accumulation, proceeding from slice-level anatomy, to volumetric structure, and finally to report-aligned semantics, will better support transfer across the clinical CT workflow than single-stage alternatives of comparable scale, while also preserving clinically meaningful severity structure in the embedding space. Stage-wise rather than joint training is motivated by both data asymmetry and objective interference: self-supervised learning can exploit the full unlabelled corpus, whereas vision-language alignment is limited to the paired subset, and optimizing dense spatial objectives together with semantic alignment may degrade one or both~\cite{clip2021}. Under this view, each phase addresses a limitation of the previous one, progressively extending the representation from slice-level anatomy to three-dimensional structure and finally to clinically grounded semantics. We evaluate this progression directly using phase ablations rather than treating it as an assumption.

Here we introduce FlexiCT, a family of CT foundation models trained through agglomerative continual pretraining on 266,227 CT volumes from 56 publicly available datasets. Using a shared public corpus and a sequential pretraining strategy, we derive three checkpoints with increasing representational scope: FlexiCT-2D captures slice-level anatomy, FlexiCT-3D adds volumetric understanding, and FlexiCT-3D-VLM aligns visual representations with clinical language. We evaluate FlexiCT across five downstream task families: segmentation, classification, registration, vision-language understanding, and clinical retrieval, and show that it matches or improves on relevant baselines across multiple benchmarks. We also show that training-free cross-modal registration provides a complementary assessment of anatomical representation quality. In addition, FlexiCT volumetric embeddings show structure associated with disease severity, stratifying lung cancer T-stage and renal cell carcinoma grade without staging supervision. These findings suggest that CT foundation models trained through sequential pretraining can encode anatomical and disease-relevant representations within a single transferable model family, with potential utility in retrieval, prompt-based analysis and other annotation-limited settings.
\section{Results}
\label{sec:results}

\begin{figure}[!htbp]
\centering
\includegraphics[width=\textwidth]{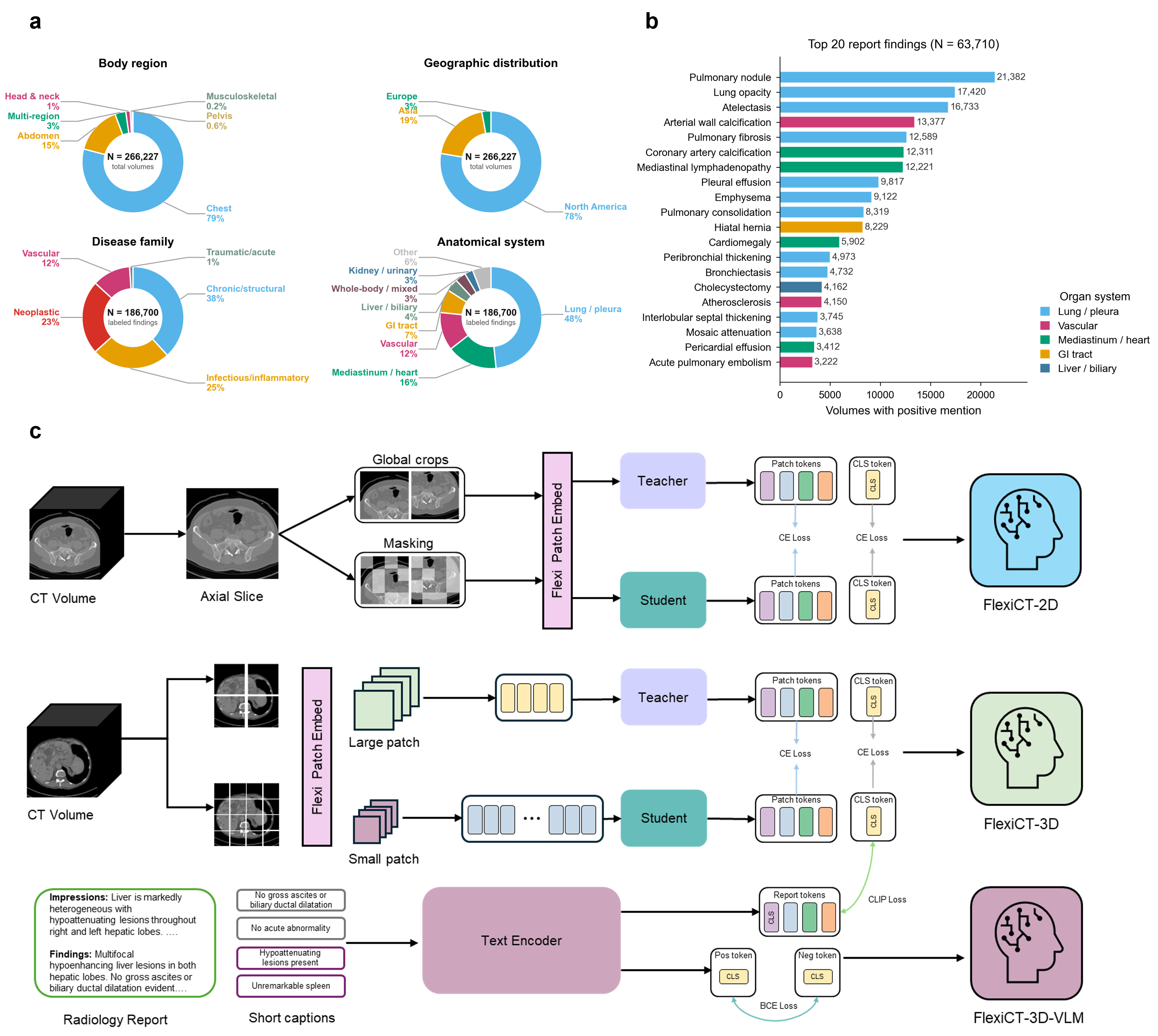}
\caption{\textbf{Dataset statistics and three-stage pretraining strategy of FlexiCT.}
\textbf{a}, Composition of the FlexiCT pretraining dataset. Four donut charts summarise body region (top left; $n = 266{,}227$ volumes), geographic distribution (top right; $n = 266{,}227$), disease family (bottom left; $n = 186{,}700$ volumes with case- or cohort-level labels) and anatomical system (bottom right; same $n$).
\textbf{b}, Frequency of the top 20 clinical concepts in the paired CT-report subset ($n = 63{,}710$ unique volumes from CT-RATE, Merlin and INSPECT). Concepts are derived from reports and metadata and ranked by number of volumes with positive mention.
\textbf{c}, Three-stage agglomerative continual pretraining pipeline. Stage~1 trains FlexiCT-2D on axial CT slices with a DINO-style teacher-student self-supervised objective over patch-embedded tokens. Stage~2 inflates the 2D encoder to 3D and continues teacher-student pretraining on volumetric crops (FlexiCT-3D). Stage~3 (FlexiCT-3D-VLM) decomposes paired radiology reports into short positive and negative caption statements, encodes them with a shared text encoder, and aligns them with 3D image representations through a contrastive loss. Each stage is initialized from the previous stage's teacher, and only the final teacher checkpoint is retained.}

\label{fig:overview}
\end{figure}

We developed FlexiCT, a family of CT foundation models trained through agglomerative continual pretraining on 266,227 CT volumes drawn from 56 public datasets (Fig.~\ref{fig:overview}; Methods). Each training phase addresses a limitation of its predecessor. FlexiCT-2D captures slice-level anatomy but does not model volumetric relationships (Phase~1). FlexiCT-3D adds volumetric understanding but remains limited to visual features (Phase~2). FlexiCT-3D-VLM further aligns visual representations with clinical text, yielding a representation family that spans anatomy, disease, and semantics (Phase~3). 

To test whether this progressive accumulation yields a genuinely general CT representation, we evaluated FlexiCT on 18 benchmarks across five downstream task families: segmentation, registration, classification, tumor phenotype retrieval, and vision--language understanding. Our evaluation logic mirrors hierarchy of clinical CT interpretation. We first assessed dense anatomical understanding through segmentation, cross-modal registration. We then tested whether frozen features capture pathological texture by evaluating disease classification under varying label budgets. Next, we used phenotype retrieval analyses to examine whether volumetric embeddings reflected tumor severity without staging supervision. Finally, we evaluated whether report-aligned representations enable zero-shot disease classification and semantic retrieval using clinical language prompts. This progression, from anatomical localization to disease characterization to language-grounded reasoning, parallels the cognitive workflow of a radiologist and provides a structured framework for assessing whether agglomerative pretraining produces representations that transfer across the clinical CT pipeline.

\subsection{A single representation transfers across organs and lesions}
\label{sec:results_seg}

\begin{figure}[htbp!]
\centering
\includegraphics[width=\textwidth]{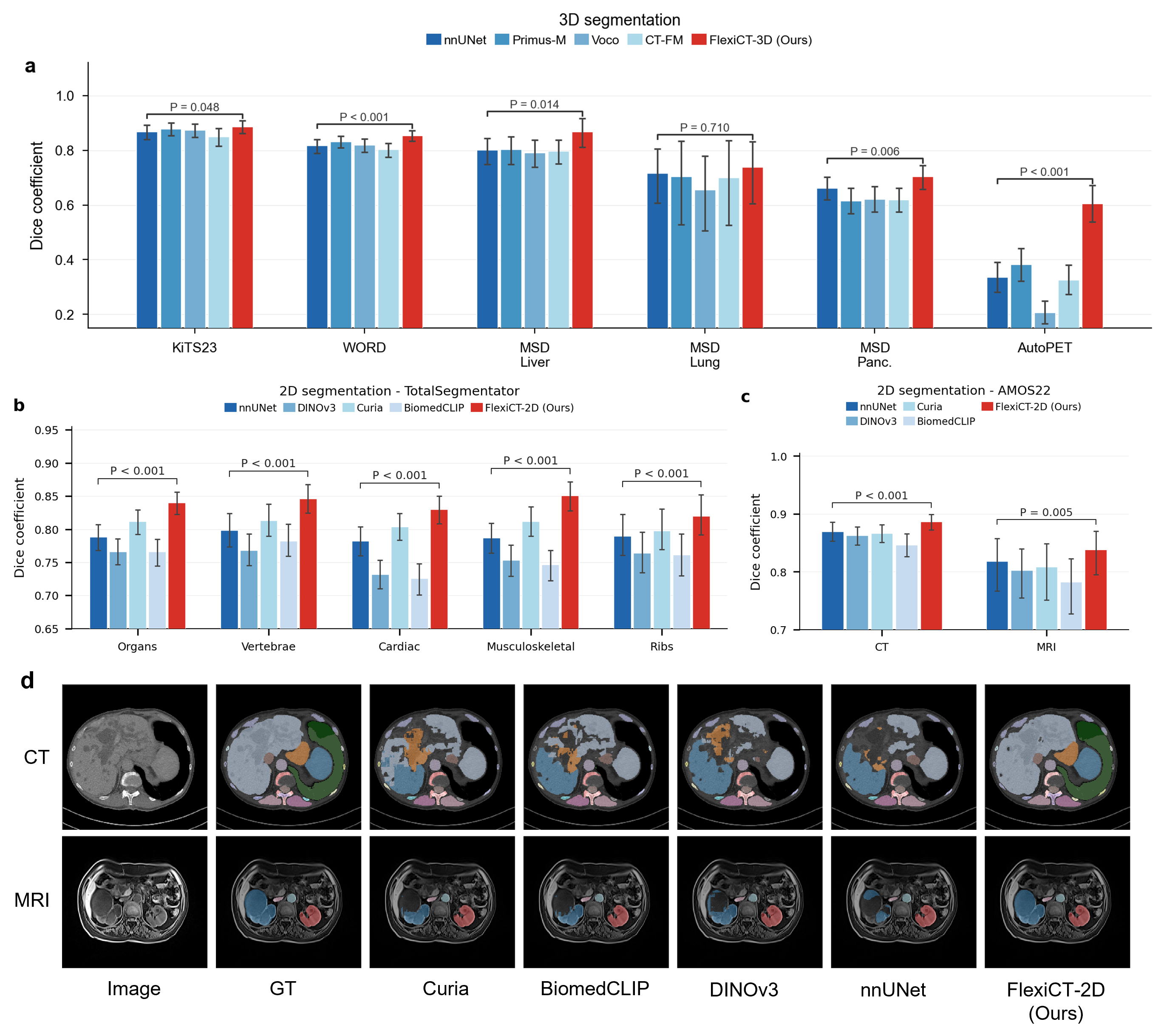}
\caption{\textbf{FlexiCT outperforms foundation models across 3D and 2D segmentation benchmarks.}
\textbf{a}, Volumetric segmentation Dice coefficient on six abdominal, thoracic and whole-body benchmarks (KiTS23, WORD, MSD Liver, MSD Lung, MSD Pancreas, and AutoPET), comparing nnU-Net, Primus-M, VoCo, CT-FM and FlexiCT-3D (red). 
\textbf{b}, Slice-level segmentation Dice coefficient on TotalSegmentator (104 anatomical classes partitioned into five groups: organs, vertebrae, cardiac, musculoskeletal, ribs), comparing nnU-Net, DINOv3, Curia, BiomedCLIP and FlexiCT-2D (red). Per-class Dice scores are aggregated by case.
\textbf{c}, Slice-level segmentation Dice coefficient on AMOS22 (15 abdominal structures) grouped by modality (CT, MRI), with the same five methods and ordering as panel \textbf{b}.
\textbf{d}, Representative qualitative segmentation on AMOS22. Columns show the input image, ground-truth segmentation and predictions from nnU-Net, DINOv3, Curia, BiomedCLIP and FlexiCT-2D, with organ masks overlay. Top row, CT; bottom row, MRI.
In \textbf{a}--\textbf{c}, bars denote mean Dice across validation cases and error bars indicate 95\% BCa bootstrap CIs ($n = 10{,}000$ resamples). Brackets report two-sided $P$ values from permutation tests with Holm--Bonferroni correction, comparing FlexiCT with nnUNet in each benchmark.}
\label{fig:segmentation}
\end{figure}

Agglomerative pretraining produces dense anatomical features that generalize across organ systems, lesion types, and spatial dimensionalities (Fig.~\ref{fig:segmentation}). At the slice level, FlexiCT-2D achieved a Dice coefficient of 0.879 on AMOS22~\cite{amos2022} (averaged across CT and MR; per-modality Dice in Fig.~\ref{fig:segmentation}c), outperforming Curia (0.857)~\cite{curia2024}, DINOv3 (0.853)~\cite{dinov3}, BiomedCLIP (0.836)~\cite{biomedclip2023}, and a matched nnU-Net baseline (0.861). On TotalSegmentator~\cite{wasserthal2023totalsegmentator}, which covers 104 anatomical structures, the advantage was maintained (0.842 versus 0.811 for Curia, 0.793 for nnU-Net, 0.762 for DINOv3, and 0.762 for BiomedCLIP). Because AMOS22 includes both CT and MR acquisitions, the comparable performance across modalities suggest that Phase~1 captures modality-invariant organ structure rather than contrast-specific texture.

For volumetric segmentation, FlexiCT-3D matched or exceeded all CT foundation model baselines across six benchmarks spanning abdominal organs, thoracic structures, and tumors. On KiTS23~\cite{kits2023}, FlexiCT-3D achieved an average Dice of 0.887 across the kidney, mass, and tumor classes, outperforming nnU-Net (0.867)~\cite{nnunet2021}, Primus-M (0.878)~\cite{primus2024}, VoCo (0.875)~\cite{voco2024}, and CT-FM (0.850)~\cite{harvardctfm2024}. On AutoPET~\cite{autopet2022}, FlexiCT-3D achieved 0.605, exceeding the next-best foundation model, Primus-M (0.382), by 22.3 absolute points. Because all baselines were re-evaluated under the same preprocessing pipeline, this margin reflects a genuine performance difference. The magnitude nonetheless warrants cautious interpretation pending independent replication on additional PET-CT cohorts. Across the Medical Segmentation Decathlon (MSD) benchmarks for liver, lung, pancreas, FlexiCT-3D achieved an average Dice of 0.770, outperforming all foundation model baselines (Primus-M 0.707, VoCo 0.689, CT-FM 0.706). On WORD~\cite{word2022}, FlexiCT-3D achieved a Dice of 0.854.

These results demonstrate that Phase~2 benefits from Phase~1 initialization. The 3D backbone inherits slice-level anatomical knowledge while gaining the volumetric context necessary for three-dimensional clinical tasks. A matched Phase~2 ablation supports this interpretation for volumetric segmentation. Under the same Phase~2 recipe and compute budget, initializing from the Phase~1 checkpoint improved Dice over random initialization on WORD (0.854 versus 0.829; Supplementary Table~\ref{tab:supp_ablation_phase1}). From a clinical perspective, current segmentation workflows require separate models for each anatomical region and lesion type, which imposes substantial engineering overhead during deployment. A single pretrained backbone that, when pairs with a lightweight decoder, matches or exceed organ-specific and lesion-specific pipelines across six benchmarks would substantially reduce this overhead and accelerate translation of CT segmentation into routine practice.

\begin{figure}[htbp!]
    \centering
    \includegraphics[width=\textwidth]{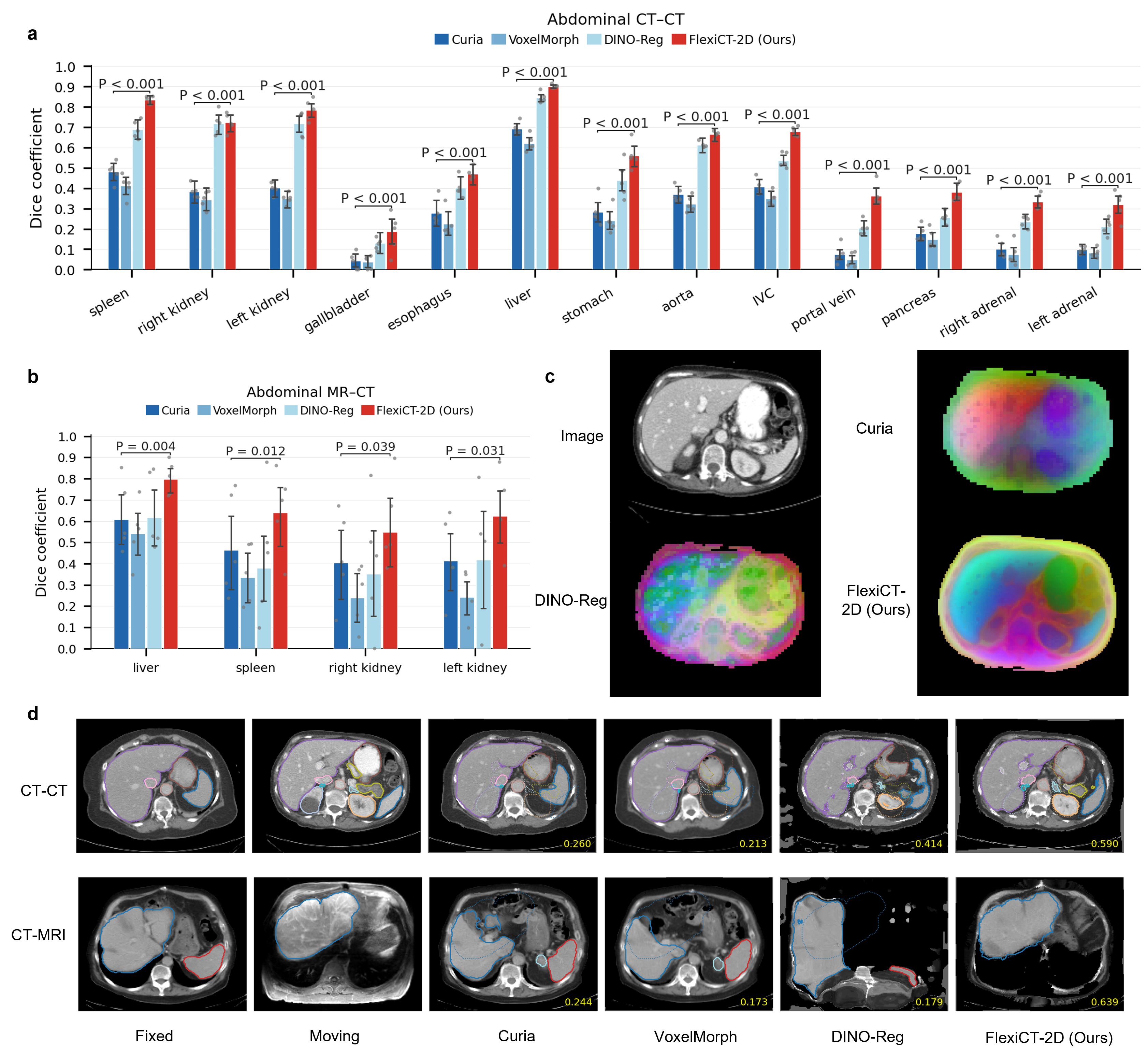}
    \caption{\textbf{FlexiCT-2D enables training-free intra- and cross-modal abdominal registration.}
    \textbf{a}, Per-organ Dice similarity coefficient on the Learn2Reg abdominal CT--CT task across 13 organs ($n = 45$ registration pairs across 5-fold cross-validation), comparing VoxelMorph, Curia, DINO-Reg and FlexiCT-2D (red). Curia, DINO-Reg and FlexiCT-2D share the same ConvexAdam optimisation framework and differ only in the feature backbone; VoxelMorph is a supervised learning baseline.
    \textbf{b}, Per-organ Dice on the Learn2Reg abdominal MR--CT task across 4 organs ($n = 19$ registration pairs across 5-fold cross-validation), computed identically to \textbf{a}.
    \textbf{c}, Principal component analysis (PCA) of patch-level features on a common axial abdominal CT slice. Top-left, input CT slice (reference). Remaining three tiles show PCA-to-RGB feature maps from Curia, DINO-Reg and FlexiCT-2D.
    \textbf{d}, Representative qualitative registration results. Each row shows one case, arranged left-to-right as moving (unregistered source), fixed (target) and the warped moving image produced by each method. For the method tiles, solid contours show the warped moving segmentation and dashed contours the fixed target segmentation; mean dice across organs is annotated in the lower-right corner. Top row, CT--CT; bottom row, MR--CT.
    In \textbf{a} and \textbf{b}, bars denote mean DSC and error bars indicate 95\% BCa bootstrap CIs ($n = 10{,}000$ resamples); 5-fold scatter is overlaid in grey. Brackets report two-sided $P$ values from permutation tests with Holm--Bonferroni correction, comparing FlexiCT-2D with Curia.
    }
\label{fig:registration}
\end{figure}

\subsection{Anatomical pretraining yields emergent cross-modal spatial correspondence}
\label{sec:results_reg}

Segmentation accuracy alone does not prove that a representation encodes genuine spatial anatomy, because a model could memorize organ appearances without learning their geometric relationships. Cross-modal registration provides a more stringent test. If features capture true anatomical structure, they should support spatial correspondence across imaging modalities without any deformation supervision (Fig.~\ref{fig:registration}). Unlike segmentation leveraging local texture cues, registration requires dense, spatially coherent feature maps that preserve geometry across modality-specific intensity distributions, making it a suitable probe of whether self-supervised features encode genuine structure.

We first assessed intra-modal spatial correspondence. On CT to CT registration using the training-free ConvexAdam framework~\cite{dinoreg2024} (see Methods~\ref{sec:methods_downstream}), FlexiCT-2D achieved an average Dice of 0.565, substantially outperforming DINO-Reg (0.278), Curia (0.299), and VoxelMorph~\cite{voxelmorph} (Fig.~\ref{fig:registration}a). This margin is particularly notable because the three feature-based encoders (Curia, DINO-Reg, FlexiCT-2D) share a comparable architecture and are applied in an identical training-free pipeline, while VoxelMorph represents a supervised learning-based approach; the performance gap therefore reflects differences in the quality of learned spatial features rather than architectural advantages.

Cross-modal registration of CT to MR provides an even more demanding test. FlexiCT-2D achieved an average CT to MR Dice of 0.654 across four abdominal organs, compared to 0.476 for Curia~\cite{curia2024} and 0.443 for DINO-Reg (Fig.~\ref{fig:registration}b). Organ-wise permutation tests showed higher Dice for FlexiCT-2D across all four organs, but after correction for multiple comparisons only the liver and spleen remained statistically significant (liver $P = 0.004$, spleen $P = 0.012$, right kidney $P = 0.039$, left kidney $P = 0.031$; corrected significance $alpha$ = 0.0125). The largest absolute gains were observed for the spleen (0.641 versus 0.380 for DINO-Reg) and left kidney (0.624 versus 0.418), organs whose shape variability and positional variation make cross-modal alignment particularly demanding. On the liver, FlexiCT-2D achieved 0.797 with a 95th-percentile Hausdorff distance (HD95) of 6.19~mm, compared to 15.65~mm for DINO-Reg and 11.71~mm for Curia. This boundary precision is within the millimetre-scale range relevant to CT--MR fusion for liver and upper-abdominal treatment planning, where multimodal registration informs target and organ-at-risk delineation. We note that this should not be interpreted as a universal radiotherapy tolerance, as acceptable registration error depends on treatment site and modality. 

That emergence of spatial correspondence in the absence of any registration loss, deformation field supervision, or multi-modal training data provides strong evidence that Phase~1 learns genuine anatomical structure rather than modality-specific texture. This capability has direct clinical relevance. Cross-modal registration underpins workflows in radiation therapy targeting, where CT provides electron density for dose calculation and MR provides soft-tissue contrast for tumor delineation. It is also central to surgical navigation, where pre-operative MR must be aligned to intra-operative CT for real-time guidance. A foundation model that inherently encodes modality-invariant spatial structure could reduce the need for task-specific registration algorithms and the associated training data and engineering effort required to deploy them in these settings.

\subsection{Label-efficient disease classification from frozen representations}
\label{sec:results_cls}

\begin{figure}[!htbp]
\centering
\includegraphics[width=\textwidth]{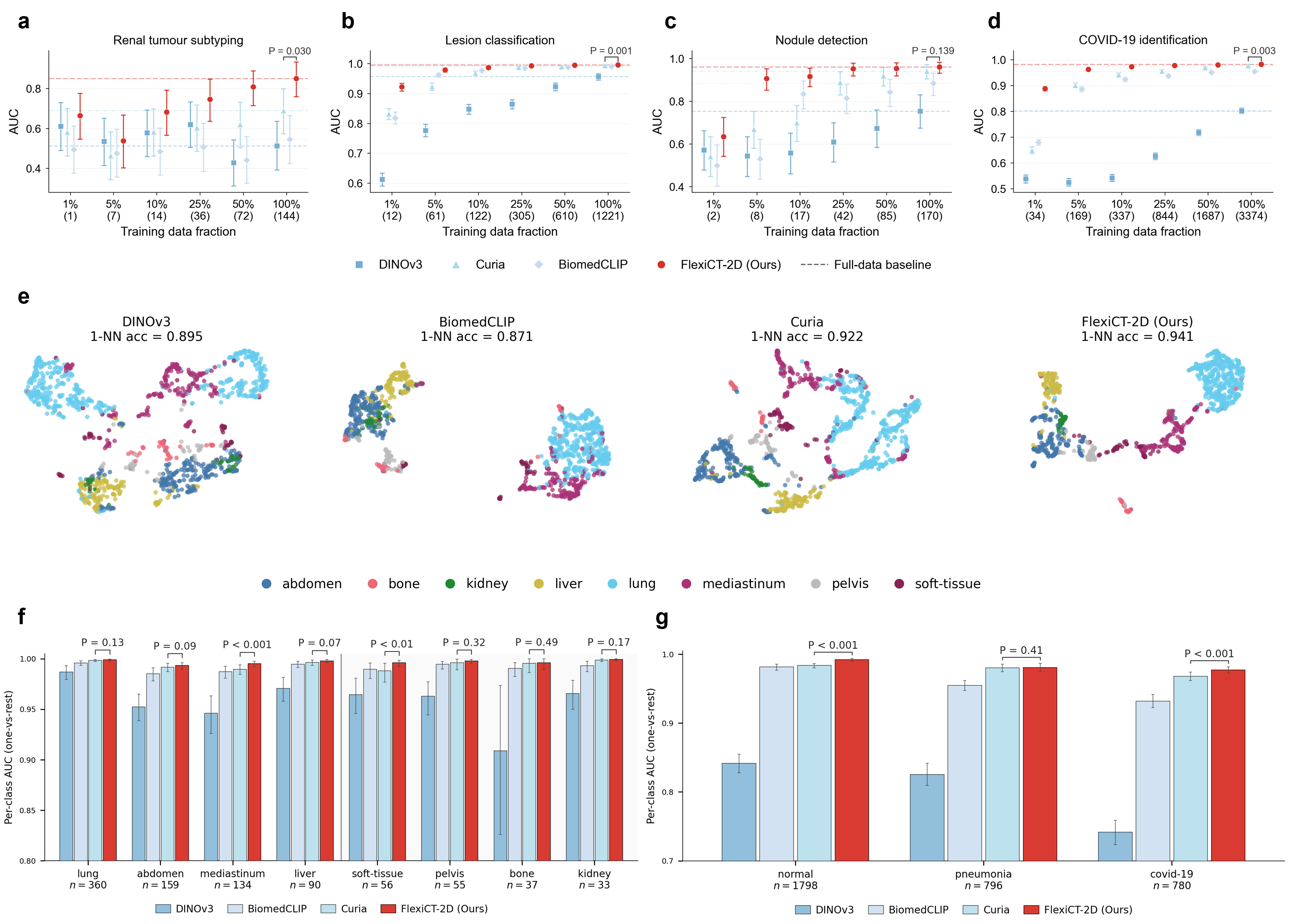}
\caption{\textbf{FlexiCT-2D enables label-efficient disease classification from frozen features.}
\textbf{a}--\textbf{d}, Label-efficiency curves for frozen pretrained encoders trained for: renal tumor subtyping (KiTS; \textbf{a}), universal lesion classification (Deep-Lesion; \textbf{b}), pulmonary nodule detection (Luna16; \textbf{c}) and COVID-19 identification (Covidx-CT; \textbf{d}). X-axis labels give training-sample counts; dashed lines mark each model's full-data AUC ($n = 144$, $1{,}221$, $170$, $3{,}374$ for \textbf{a}--\textbf{d}).
\textbf{e}, UMAPs of L2-normalized Deep-Lesion test features ($n = 1{,}221$ slices for DINOv3, BiomedCLIP, Curia and FlexiCT-2D, coloured by lesion site; titles show leave-one-out 1-nearest-neighbour accuracy in feature space.
\textbf{f}, \textbf{g}, Per-class one-versus-rest AUCs at the 100\% training fraction for Deep-Lesion (\textbf{f}) and Covidx-CT (\textbf{g}); shaded Deep-Lesion classes have $n < 90$.
Markers and bars show mean AUC with 95\% bootstrap CIs ($n = 10{,}000$ resamples). Brackets compare FlexiCT-2D with Curia using two-sided paired permutation tests. AUC, area under the receiver-operating-characteristic curve; UMAP, uniform manifold approximation and projection}
\label{fig:classification}
\end{figure}

The segmentation and registration results establish that Phase~1 encodes dense anatomical structure.  We next asked whether the same frozen features also capture pathological representations to support clinical decision-making (Fig.~\ref{fig:classification}).

Using the full training set, a linear classifier on frozen features achieved the highest area under the receiver operating characteristic curve (AUC) on all four disease benchmarks. FlexiCT-2D reached 0.997 (95\% confidence interval (CI): 0.995-0.998) on Deep-Lesion~\cite{deeplesion2018}, 0.983 (95\% CI: 0.980-0.987) on Covidx~\cite{covidxct2021}, 0.961 (95\% CI: 0.933-0.983) on Luna16~\cite{luna16}, and 0.851 (95\% CI: 0.760-0.934) on KiTS~\cite{kits2023}. The corresponding values for the next-best baseline, Curia~\cite{curia2024} were 0.994, 0.977, 0.942, 0.690, and the differences favoured FlexiCT-2D in three of the four benchmarks (paired permutation test: Deep-Lesion, $P < 0.001$; Covidx-CT, $P=0.003$; KiTS, $P = 0.032$; LUNA16 showed a non-significant trend, $P=0.139$). FlexiCT-2D also outperformed BiomedCLIP~\cite{biomedclip2023} (0.991, 0.956, 0.884, 0.546), and DINOv3~\cite{dinov3} (0.957, 0.803, 0.755, 0.514), respectively. The advantage over the next-best encoder, Curia, was statistically significant on three of four benchmarks (paired permutation test; Deep-Lesion $P < 0.001$, Covidx $P = 0.003$, KiTS $P = 0.032$), with Luna16 showing a non-significant trend ($P = 0.139$). These four benchmarks span distinct clinical scenarios---renal tumor subtyping (KiTS), universal lesion characterization across body regions (Deep-Lesion), pulmonary nodule detection (Luna16), and viral pneumonia identification (Covidx). The consistent advantage across these tasks indicates that agglomerative pretraining produces disease-relevant features that are not organ-specific. 

Per-class analysis confirmed that these gains are not driven by a single dominant category. On Deep-Lesion, FlexiCT-2D achieved the highest one-versus-rest AUC across all eight anatomical lesion sites including low-prevalence tail classes (Fig.~\ref{fig:classification}f). On Covidx, FlexiCT-2D led on all three diagnostic categories (Fig.~\ref{fig:classification}g). Uniform manifold approximation and projection (UMAP) of frozen Deep-Lesion features (Fig.~\ref{fig:classification}e) further illustrated this organisation. FlexiCT-2D yielded the tightest lesion-site clusters, with a leave-one-out 1-nearest-neighbour accuracy 0.941, compared with 0.922 for Curia, 0.895 for DINOv3, and 0.871 for BiomedCLIP. The learned representations therefore arrange pathological content by anatomical context. The fact that a frozen backbone trained with self-supervised objectives alone discriminates pathologies as diverse as renal tumor subtypes and viral pneumonia, without task-specific fine-tuning, suggests that the representation captures pathological texture alongside anatomical structure.

In clinical practice, however, labelled CT data is often scarce for rare pathologies, emerging diseases or new imaging protocols. The practical value of a foundation model also depends on how quickly useful accuracy can be reached with limited supervision. We next investigated the label efficiency of the four pretrained encoders by varying the fraction of available training data from 1\% to 100\% across all four benchmarks (Fig.~\ref{fig:classification}a--d). For each fraction, we trained a linear classifier on frozen encoder features and evaluated on a fixed held-out test set (see Methods).

FlexiCT-2D achieved the highest AUC at every label budget tested. At 5\% of training data, FlexiCT-2D surpassed the full-data AUC of DINOv3 on three of four benchmarks: Deep-Lesion (0.979 versus 0.957 at 100\%), Luna16 (0.900 versus 0.755), and Covidx (0.964 versus 0.803), representing a 20-fold reduction in required labels. On KiTS, performance was comparable (0.538 versus 0.514). The same 5\% fraction also exceeded BiomedCLIP's full-data performance on Luna16 (0.900 versus 0.884) and Covidx (0.964 versus 0.956), indicating that FlexiCT-2D features with twenty times fewer labels can match what competing encoders achieve with the entire training set. On KiTS, the most challenging benchmark where all models exhibited wide confidence intervals reflecting the small cohort ($n = 95$), FlexiCT-2D at 25\% of data (0.747, 95\% CI: 0.637--0.848) surpassed every baseline at 100\% (Curia: 0.690, BiomedCLIP: 0.546, DINOv3: 0.514). At 1\% of data, the separation was especially pronounced on Deep-Lesion (0.922 versus 0.832 for the next-best encoder, Curia) and Covidx (0.889 versus 0.680 for BiomedCLIP, 0.650 for Curia, 0.539 for DINOv3). This steep degradation of baseline encoders in the low-data regime suggests that FlexiCT produces features with higher intrinsic discriminability to linearly separate disease classes.

These results indicate that the advantage of agglomerative pretraining persists as labels become scarce. Together with the segmentation and registration findings, Phase~1 features support dense spatial tasks, cross-modal correspondence, and label-efficient disease discrimination within a single frozen feature space, a property of direct relevance to clinical settings in which the cost of expert annotation constrains the development of supervised models.

\subsection{Volumetric embeddings organize tumors by clinical severity}
\label{sec:results_retrieval}

\begin{figure}[!htbp]
\centering
\includegraphics[width=\textwidth]{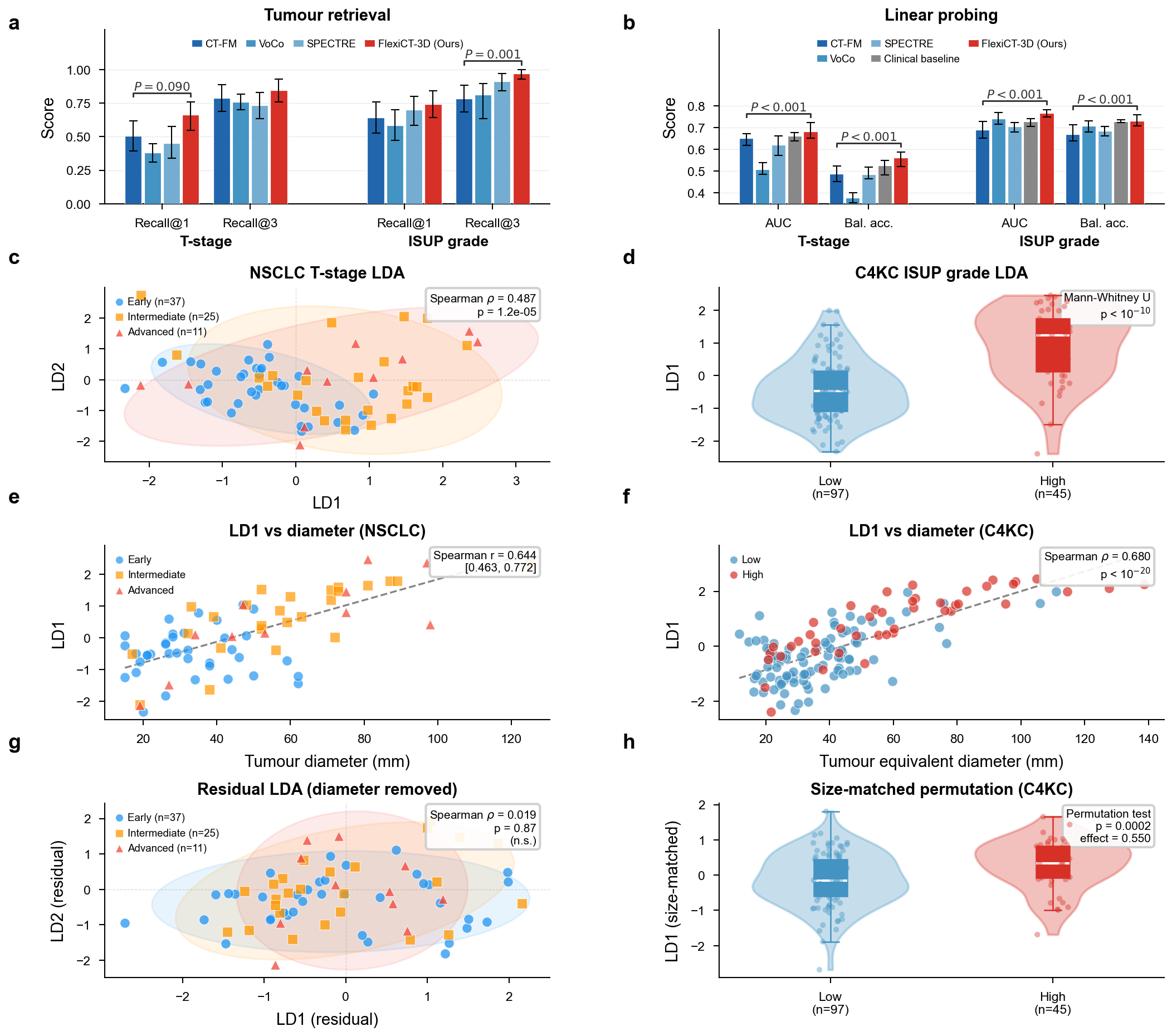}
\caption{\textbf{FlexiCT-3D embeddings organize tumors along clinical severity gradients without staging supervision.}
\textbf{a}, Zero-shot tumor retrieval (Recall@1, Recall@3) for T-stage (NSCLC-Radiogenomics) and ISUP grade (C4KC-KiTS), comparing CT-FM, VoCo, SPECTRE and FlexiCT-3D.
\textbf{b}, Linear probing (AUC, balanced accuracy) on frozen embeddings for T-stage and ISUP grade, including a tumor-diameter-only clinical baseline (grey).
\textbf{c}, LDA projection of FlexiCT-3D embeddings for NSCLC T-stage (Early $n = 37$, Intermediate $n = 25$, Advanced $n = 11$). Tumors form an ordered severity gradient along LD1 (Spearman $\rho = 0.487$, $P = 1.2 \times 10^{-5}$); shaded regions are per-class covariance ellipses.
\textbf{d}, LDA projection for C4KC-KiTS ISUP grade (Low $n = 97$, High $n = 45$). Low and high grade renal tumors separate along LD1 (Mann--Whitney $U$, $P < 10^{-9}$); violins show the full distribution and boxplots mark the median and interquartile range.
\textbf{e}, NSCLC LD1 plotted against tumor diameter (Spearman $\rho = 0.644$, 95\% CI: 0.463--0.772); points are coloured by T-stage as in \textbf{c}.
\textbf{f}, C4KC-KiTS LD1 plotted against tumor equivalent diameter ($\rho = 0.680$, $P < 10^{-20}$); points are coloured by ISUP group.
\textbf{g}, NSCLC T-stage gradient after regressing tumor diameter out of LD1 and LD2 ($\rho = 0.019$, $P = 0.87$, n.s.).
\textbf{h}, C4KC-KiTS ISUP separation under a size-matched permutation test ($P = 0.0002$, effect size $= 0.550$).
In \textbf{a} and \textbf{b}, bars denote point estimates and error bars indicate 95\% bootstrap CIs ($n = 10{,}000$ resamples).}     
\label{fig:phenotype}
\end{figure}

Phase~1 encodes anatomy and disease texture at the slice level, but clinical staging decisions, such as stratifying renal tumor aggressiveness, require 3D volumetric information. Our Phase~2 volumetric embeddings address this limitation. We investigate whether these embeddings organize tumors along clinically meaningful axes using tumor-similarity retrieval directly from the embedding space (Fig.~\ref{fig:phenotype}). Specifically, we use known tumor cases as queries and retrieve staging cases with nearby latent representations using cosine similarity.

On the NSCLC-Radiogenomics cohort~\cite{nsclcradiogenomics}, zero-shot retrieval of T-stage (grouped as Early, Intermediate, and Advanced) yielded Recall@1 of 0.662 (95\% CI: 0.549-0.761) for FlexiCT-3D, compared to 0.507 (95\% CI: 0.394-0.620) for CT-FM~\cite{harvardctfm2024} and 0.451 (95\% CI: 0.338-0.577) for SPECTRE~\cite{SPECTRE2024}. A two-sided permutation test against CT-FM with 10,000 permutations gave $P = 0.090$ for Recall@1 comparison, indicating a directional but not conventionally significant improvement. On C4KC-KiTS cohort~\cite{Heller2019TheSO}, retrieval of International Society of Urological Pathology (ISUP) grade in clear cell renal cell carcinoma (low grade 1 and 2 versus high grade 3 and 4) reached Recall@3 of 0.971 (95\% CI:
0.929--1.000), compared to 0.914 for SPECTRE and 0.786 for CT-FM. The FlexiCT-3D versus CT-FM difference was significant by the same test ($P = 0.0011$). These results indicate that the volumetric representation captures a clinically meaningful severity continuum.

Linear classifiers on frozen FlexiCT-3D embeddings further quantified how much severity information the representation encodes. For ISUP grade, FlexiCT achieved balanced accuracy of 0.730 (95\% CI: 0.707-0.759) and AUC of 0.765 (95\% CI: 0.749-0.782), compared to 0.685 and 0.705 for SPECTRE and 0.669 and 0.689 for CT-FM. Both CT-FM comparisons were significant by two-sided permutation tests with 10,000 permutations (balanced accuracy, $P = 0.0001$; AUC, $P = 0.0001$). For T-stage, FlexiCT reached balanced accuracy of 0.561 (95\% CI: 0.521--0.588) and AUC of 0.681 (95\% CI: 0.651--0.724), compared with 0.487 and 0.651 for CT-FM. These CT-FM comparisons were also significant (balanced accuracy, $P = 0.0001$; AUC, $P = 0.0004$). The T-stage balanced accuracy exceeds the clinical baseline using tumor diameter (0.525; 95\% CI: 0.483--0.548), though the overlapping confidence intervals indicate comparable performance. These results suggest the representation encodes severity-relevant information that could complement existing protocols.

We next sought to characterize the geometric structure that underlies these retrieval and classification results. We projected FlexiCT-3D embeddings into two dimensions using linear discriminant analysis. In the NSCLC-Radiogenomics cohort ($n = 73$; Early $n = 37$, Intermediate $n = 25$, and Advanced $n = 11$), the primary discriminant axis separated Early from Advanced T-stages with an interpretable gradient (Spearman $\rho = 0.487$, $p = 1.2\times10^{-5}$; Fig.~\ref{fig:phenotype}c). Intermediate stages occupied transitional positions rather than forming isolated clusters. The leading discriminant coordinate correlated strongly with tumor diameter ($r = 0.644$, 95\% CI: 0.463-0.772; Fig.~\ref{fig:phenotype}e), consistent with T-stage being primarily a size-based staging system; after regressing out diameter, the T-stage gradient vanished ($\rho = 0.019$, $p = 0.87$; Fig.~\ref{fig:phenotype}g). In the C4KC-KiTS cohort ($n = 142$), low-grade and high-grade renal tumors separated along the leading discriminant axis (Mann--Whitney $p < 10^{-9}$; Fig.~\ref{fig:phenotype}d). The ISUP grade signal persisted after controlling for tumor size (permutation $p = 0.0002$, effect size $= 0.550$; Fig.~\ref{fig:phenotype}h), indicating that the embedding encodes severity features beyond gross morphology, consistent with the nuclear and architectural atypia that defines ISUP grading.

We also conducted ablations on the effectiveness of agglomerative pretraining on tumor phenotype retrieval. Specifically, Phase~1 initialization improved the C4KC-KiTS phenotype tasks: compared with the randomly initialized Phase~2 variant, FlexiCT-3D improved ISUP Recall@1 retrieval (0.743 versus 0.613) and increased ISUP linear-probe AUC from 0.703 to 0.765 (Supplementary Table~\ref{tab:supp_ablation_phase1}). These comparisons indicate that the severity structure is strengthened by agglomerative transfer from Phase~1 rather than by 3D training alone. The implications for clinical workflows, including tumor-similarity retrieval for treatment planning, are considered in the Discussion.

\subsection{Clinical language enables zero-shot disease classification}
\label{sec:results_zeroshot}

Volumetric embeddings organize disease severity, but their use for classification or retrieval still requires a small number of labelled samples. To address this limitation, Phase~3 evaluates whether report-aligned structure in the embedding space can be used directly for zero-shot disease recognition. Specifically, report-aligned semantic agglomeration enables FlexiCT to classify diseases using text prompts alone, without any labelled training data (Fig.~\ref{fig:vlm}). In clinical practice, this capability could support zero-shot triage by matching scans against text descriptions of suspected pathologies, an application of particular value for rare or emerging conditions for which curated labelled cohorts are unlikely to be assembled. 

FlexiCT-3D-VLM achieved AUC of 0.813 (95\% CI: 0.807--0.820) on CT-RATE~\cite{ctrate2024} for zero-shot multi-abnormality classification across 18 chest CT findings, exceeding CT-CLIP (0.732)~\cite{ctclip2024}, COLIPRI (0.787; 95\% CI: 0.780--0.794)~\cite{coplpri2024}, and SPECTRE (0.567; 95\% CI: 0.558--0.577)~\cite{SPECTRE2024}. On the Merlin abdominal CT benchmark~\cite{merlin2024}, which covers 30 abdominal findings, FlexiCT-3D-VLM reached AUC of 0.872 (95\% CI: 0.862--0.882), outperforming Merlin itself (0.825; 95\% CI: 0.812--0.838) and COLIPRI (0.737; 95\% CI: 0.722--0.752). Outperforming dataset-specific competitors on their own benchmarks is notable: CT-CLIP was trained on CT-RATE data and Merlin on its own paired corpus, yet FlexiCT-3D-VLM trained on a combined dataset exceeds both, suggesting that broader pretraining data and the agglomerative representation strategy produce more generalizable vision--language alignment than dataset-specific training alone.

Disease-level results further validate FlexiCT-3D-VLM's advantage (Fig.~\ref{fig:vlm} c). On CT-RATE, FlexiCT-3D-VLM consistently outperforms baseline models across diverse chest pathologies from parenchymal findings (such as atelectasis, consolidation, and emphysema) to structural abnormalities (including masses, nodules, and pleural pathology). On Merlin, the advantage extends across abdominal findings spanning multiple organ systems. This breadth of per-disease performance indicates that the model has learned generalizable associations between clinical language and imaging patterns. 

FlexiCT-3D-VLM also achieved the highest F1 on CT-RATE (0.509 versus 0.482 for COLIPRI and 0.427 for CT-CLIP). On Merlin, FlexiCT-3D-VLM achieved F1 of 0.725, comparable to Merlin itself (0.735), while substantially exceeding COLIPRI (0.651). A Phase~3 initialization ablation showed the same accumulation pattern for zero-shot classification: initializing the VLM from the full Phase~2 backbone outperformed both random initialization and direct Phase~1-to-VLM initialization on CT-RATE (AUC 0.813 versus 0.761 and 0.789) and Merlin (0.872 versus 0.848 and 0.853; Supplementary Table~\ref{tab:supp_ablation_phase2}). The clinical significance of zero-shot classification lies in its potential to reduce the annotation bottleneck that constrains supervised learning in CT. New disease categories which are difficult to collect large-scale data can be recognized through text prompts. 

\subsection{Semantic report retrieval bridges imaging and clinical text}
\label{sec:results_retrieval_vlm}

\begin{figure}[!htbp]
\centering
\includegraphics[width=\textwidth]{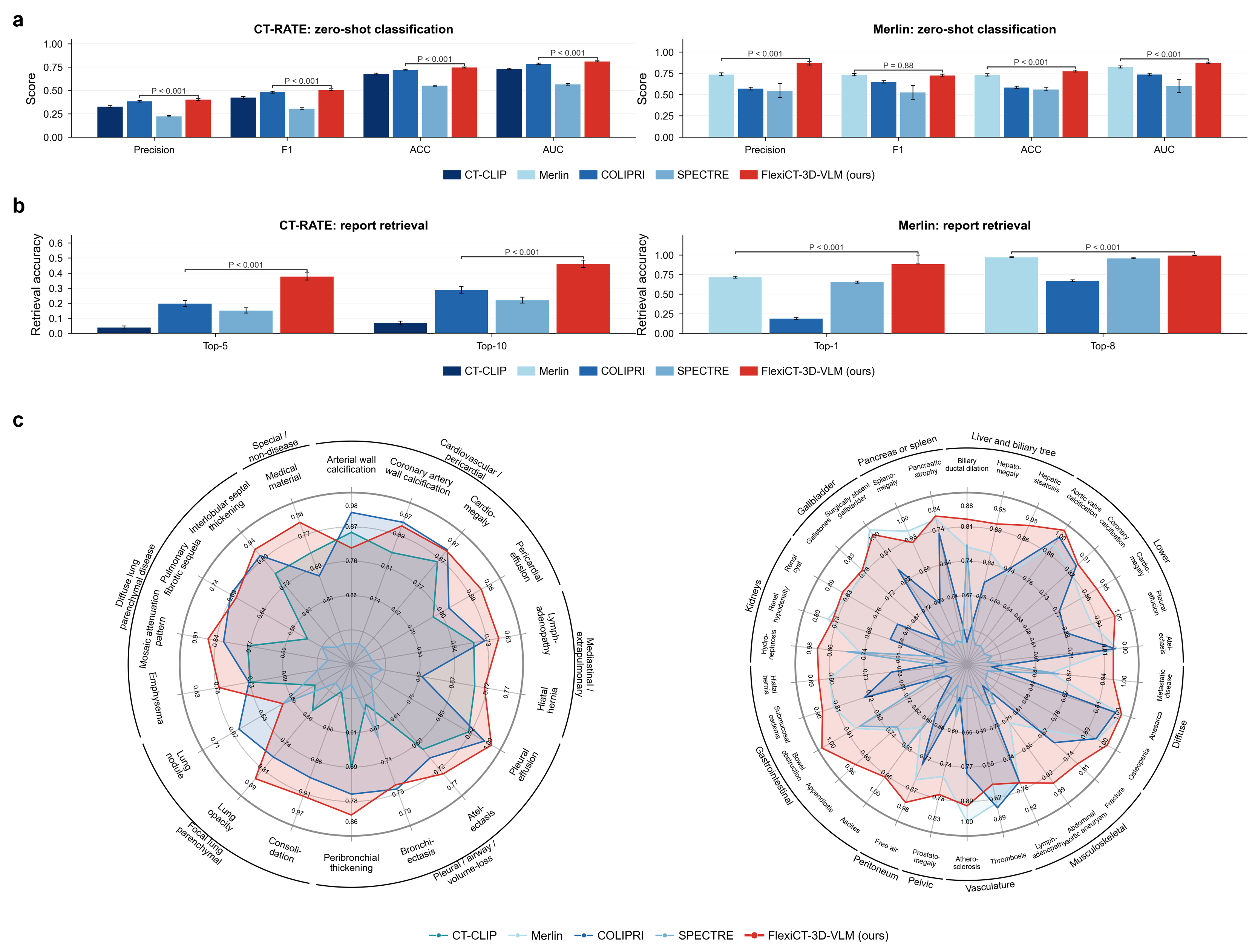}
\caption{\textbf{FlexiCT-3D-VLM supports zero-shot disease classification and report retrieval across chest and abdominal CT.}
\textbf{a}, Zero-shot multi-label disease classification on CT-RATE (left) and Merlin (right), reporting macro-averaged precision, F1, accuracy (ACC) and area under the ROC curve (AUC). Baselines are CT-CLIP, COLIPRI and SPECTRE on CT-RATE; Merlin, COLIPRI and SPECTRE on the Merlin benchmark.
\textbf{b}, Semantic report retrieval across the same two cohorts: Top-5 and Top-10 accuracy on CT-RATE (left) and Top-1 and Top-8 accuracy on Merlin at pool size $n = 32$ (right), with models as in \textbf{a}.
\textbf{c}, Per-disease zero-shot AUC radars for CT-RATE (left; 18 findings) and Merlin (right; 30 findings), grouped by anatomical systems.
In \textbf{a} and \textbf{b}, bars show point estimates with 95\% bias-corrected and accelerated bootstrap CIs ($n = 10{,}000$ resamples). Brackets report two-sided paired permutation-test $P$ values ($n = 10{,}000$ permutations), Holm--Bonferroni corrected within each benchmark, comparing each baseline with FlexiCT-3D-VLM. ACC, accuracy; AUC, area under the receiver-operating-characteristic curve.}
\label{fig:vlm}
\end{figure}

Beyond categorical disease identification, clinical workflows also require the retrieval of relevant text descriptions given a new scan. In routine practice, clinicians consult prior cases with similar imaging findings to inform differential diagnosis, treatment planning, and prognosis estimation. Phase~3 representations support this cross-modal retrieval by matching CT volumes to clinical report descriptions and clinical reports to CT volumes within a shared embedding space (Fig.~\ref{fig:vlm}b).

FlexiCT-3D-VLM achieved Top-5 retrieval accuracy of 0.378 on CT-RATE, nearly twice the value achieved by COLIPRI (0.190) and substantially higher than that of SPECTRE (0.152) and CT-CLIP (0.039). At Top-10, FlexiCT-3D-VLM reached 0.462, compared to 0.289 for COLIPRI and 0.221 for SPECTRE. The magnitude of the improvement on CT-RATE is notable given the difficulty of the benchmark, because the full retrieval pool contains diverse chest CT studies, and matching a volume to its correct clinical description requires encoding both anatomical context and disease-specific findings. On Merlin, Top-1 accuracy at pool size 32 was 0.888, compared to 0.719 for Merlin and 0.655 for SPECTRE, indicating that FlexiCT-3D-VLM can identify the correct report from a pool of 32 candidates in nearly nine out of ten cases.

These retrieval margins indicate that the semantic agglomeration phase produces an embedding space in which visual CT content and clinical text are meaningfully aligned. The Phase~3 initialization ablation showed the same accumulation pattern for retrieval as for zero-shot classification. Specifically, full Phase~2 initialization improved CT-RATE Recall@5 over random and 2D-only initialization (0.378 versus 0.318 and 0.351, respectively) and Merlin Recall@1 at $N = 32$ (0.888 versus 0.811 and 0.865; Supplementary Table~\ref{tab:supp_ablation_phase2}). Together, the zero-shot and retrieval ablations support the premise that report alignment benefits from inherited volumetric structure rather than from contrastive language training alone. Combined with the zero-shot classification results, Phase~3 demonstrates that report-aligned representations enable both categorical disease identification and graded similarity-based retrieval, addressing complementary clinical needs within a single representation.

\section{Discussion}
\label{sec:discussion}

In this study, we developed FlexiCT, a family of CT foundation models trained through agglomerative continual pretraining on 266,227 CT volumes from 56 publicly available datasets, and evaluated its representations across multiple CT analysis tasks. Our main finding is that a single sequentially pretrained model lineage can support segmentation, classification, registration, vision-language analysis and clinical retrieval, while preserving embedding structure associated with disease severity. The agglomerative design appears to preserve complementary representational properties across training stages. We found that dense spatial performance was maintained after report-based language alignment. Existing CT foundation models have shown that self-supervised pretraining can produce robust anatomy-centric features~\cite{voco2024,harvardctfm2024,SPECTRE2024} and that paired CT--report data can introduce clinical semantics~\cite{ctclip2024,merlin2024}. Our results suggest that these capabilities can be integrated within a single sequential training framework. A potential practical implication is that such a pretrained lineage may reduce the need to develop separate models for individual CT analysis tasks, thereby lowering the engineering burden associated with adapting CT AI systems across clinical workflows.

Among the analyses we conducted, three findings have particular clinical relevance. First, the label-efficiency results indicate that frozen FlexiCT-2D features with as few as 5\% of available labels match or exceed the full-data performance of competing encoders, indicating that agglomerative pretraining produces features with higher intrinsic discriminability for disease-relevant variation. This property is clinically consequential because large annotated training sets may never be assembled for rare conditions, emerging diseases, or newly adopted imaging protocols. A foundation model that reaches useful accuracy with minimal supervision therefore broadens the range of conditions for which AI-assisted interpretation is feasible. The consistency of this advantage across four diverse pathologies, including renal tumor subtyping, universal lesion characterization, pulmonary nodule detection, and pneumonia infection detection, indicates that the label efficiency reflects a general property of the learned representation rather than an artifact of any single disease domain.

Second, the training-free registration results merit particular attention because spatial correspondence was never explicitly optimized during pretraining. Self-supervised features trained exclusively on axial CT slices support cross-modal CT-MR alignment, with boundary errors approaching millimeter precision required for upper abdominal radiation therapy planning. This provides strong evidence that Phase~1 encodes genuine anatomical geometry rather than modality-specific texture. Unlike Curia~\cite{curia2024} and DINOv3~\cite{dinov3}, whose features produce substantially lower registration accuracy under the same training-free pipeline, FlexiCT achieves spatial competence as an emergent property of anatomy-focused pretraining. This is clinically significant: cross-modal registration underpins radiation therapy workflows in which diagnostic CT must be aligned with treatment-planning MR for accurate target delineation, and a foundation model that inherently encodes modality-invariant structure could reduce the need for task-specific registration algorithms and the associated engineering effort required to deploy them.

Third, the phenotype retrieval results carry an arguably more consequential implication for clinical translation. The ability to retrieve tumors based on clinical aggressiveness on two cancer types without any staging supervision suggests that the volumetric representation encodes morphological features associated with disease severity, not merely anatomical structure. The ISUP grade signal persists after controlling for tumor size, indicating that the embedding captures information beyond gross morphology. This finding is consistent with the nuclear and architectural atypia that defines ISUP grading in renal cell carcinoma. By contrast, the T-stage gradient in lung cancer collapses after size regression, which is consistent with T-stage being primarily a size-based criterion. This dissociation is informative because it suggests that the representation distinguishes size-dependent from size-independent components of clinical severity, a distinction that could clarify which imaging biomarkers carry independent prognostic value. These findings extend recent work linking self-supervised radiographic features to tumor biology from lesion-level crops to whole-volume representations. Whether such severity organization is unique to agglomerative pretraining remains unexplored. However, CT-FM~\cite{harvardctfm2024} and SPECTRE~\cite{SPECTRE2024} which are both trained at substantial scale, produce lower retrieval accuracy on the same cohorts, suggesting that the agglomerative strategy contributes beyond scale alone.

Report-aligned semantic agglomeration further bridges visual and clinical reasoning. A single FlexiCT-3D-VLM model outperforms dataset-specific competitors on their own benchmarks, suggesting that broader pretraining data combined with the agglomerative strategy produces more generalizable vision-language models. Instead of training separate vision-language models for chest and abdominal CT, we show that a single report-aligned model can serve both anatomical contexts with stronger capabilities. The phase ablations provide direct support for this interpretation (Table \ref{tab:supp_ablation_phase1}, \ref{tab:supp_ablation_phase2}). Specifically, removing Phase~1 initialization weakened WORD segmentation and C4KC-KiTS phenotype tasks, and removing Phase~2 initialization weakened Phase~3 zero-shot and retrieval performance. These comparisons do not eliminate all possible effects of data scale or architecture, but they show that each stage contributes reusable structure to the next.

Despite these advances, several limitations should be acknowledged. First, while the pretraining corpus is drawn from 56 datasets, it may over-represent certain populations, scanner manufacturers, and clinical protocols. This sampling bias could limit generalization to under-represented imaging contexts such as paediatric populations or non-standard contrast protocols. Second, FlexiCT comprises three checkpoints, each intended for different downstream use cases. This requires users to select the appropriate checkpoint for a given task, a practical consideration that may be reduced by future unified architectures. Third, all evaluations are retrospective. The clinical retrieval analysis has not been validated prospectively in treatment planning or decision support workflows, and the phenotype analysis is limited to two cancer types (NSCLC T-staging and RCC ISUP grading). Fourth, the computational requirements of agglomerative pretraining are substantial and may limit reproducibility for groups without comparable infrastructure. Finally, during Phase~3 the text encoder remains frozen while the visual encoder continues to evolve; whether joint fine-tuning or larger language encoders would further improve vision--language alignment is an open question.

Several directions follow from the current work. First, scaling laws for CT foundation models remain largely unexplored. Systematic investigation would inform resource allocation as public CT datasets grow beyond the 263{,}000-volume scale used here. Second, prospective validation of phenotype retrieval in clinical decision support is essential to translate the retrospective severity organization into actionable tools for tumor board review and treatment selection. Third, extending agglomerative pretraining to additional modality pairs (CT-MRI, CT-pathology, or CT-PET) would test whether the hierarchical design generalizes beyond a single modality and could support multimodal workflows such as radiation therapy planning and surgical navigation. Fourth, unifying FlexiCT family into a single architecture that jointly supports slice-level, volume-level, and language-aligned inference would simplify deployment. Fifth, fine-grained phenotype analysis across additional cancer types would clarify whether the severity organization observed here reflects a general property of CT foundation model embeddings.

These results establish that CT foundation learning can progress from anatomical structure through spatial correspondence to clinically meaningful disease severity organization within one transferable representation family. The implication extends beyond benchmark performance: if CT representations can be systematically accumulated across abstraction levels, then the same design principle may apply to other volumetric imaging modalities where hierarchical clinical interpretation is the norm. Agglomerative pretraining offers a principled framework for building representations that encode not only what is visible in an image, but what it means for the patient.

\section{Methods}
\label{sec:methods}

\subsection{Pretraining data curation}
\label{sec:methods_data}

We assembled a pretraining corpus of 266,227 CT volumes from 56 publicly available datasets spanning abdominal, thoracic, pelvic, head-and-neck, and whole-body anatomical regions, with contrast, non-contrast, and CT angiography acquisitions represented. Major contributing sources include NLST (132,985 volumes), CT-RATE (47,149)~\cite{ctrate2024}, Merlin (25,489)~\cite{merlin2024}, INSPECT (23,240), and FLARE'23 (4,100)~\cite{ma2024automatic}. This corpus exceeds prior public CT pretraining collections, including CT-FM (148,000 scans)~\cite{harvardctfm2024}. A complete list of datasets with volume counts, anatomical coverage, and source institutions is provided in Supplementary Table~\ref{tab:supp_pretrain_datasets}, with access URLs in Supplementary Table~\ref{tab:supp_dataset_urls}.

Automated quality control removed volumes with degenerate geometry, anomalous intensity distributions (binary masks, pre-normalized images), and heuristic duplicates across overlapping source datasets (Supplementary Methods~\ref{supp:data_qc}). All volumes were reoriented to a canonical coordinate system (LPS), resampled to 1.5\,mm spacing (in-plane for 2D, isotropic for 3D), intensity-clamped to $[-1000, 1000]$\,HU, and normalized to zero mean and unit standard deviation. For 2D pretraining, individual axial slices were extracted and body-cropped to $256 \times 256$ pixels. Full preprocessing details are provided in Supplementary Methods~\ref{supp:preprocessing}.

\subsection{Phase 1: 2D anatomical pretraining}
\label{sec:methods_phase1}

The encoder is a Vision Transformer (ViT-Base)~\cite{vit2020} adapted for single-channel CT input (embedding dimension 864, 16 blocks, 12 heads, ${\sim}120$\,M parameters). Positional information is encoded with Rotary Position Embeddings (RoPE)~\cite{dinov3}, which decouple position encoding from sequence length and enable resolution flexibility at inference time. Four learnable register tokens~\cite{darcet2023vision} absorb global information and reduce attention map artefacts.

A key design feature is the flexible patch embedding module (PatchEmbedND), which supports runtime adjustment to any target patch size via pseudoinverse-based kernel resampling~\cite{beyer2023flexivit}. This enables alternating between patch-16 (coarser, faster) and patch-8 (finer) tokenizations during training without modifying network parameters. Full architectural specifications are provided in Supplementary Methods~\ref{supp:architecture}.

We train using a DINOv3 self-supervised framework~\cite{dinov3} that combines three complementary objectives within an exponential moving average (EMA) teacher--student architecture: a DINO self-distillation loss on \texttt{[CLS]} token representations, an iBOT masked patch prediction loss~\cite{zhou2021ibot} for dense spatial learning, and a KoLeo regularizer~\cite{Fournier2016OnTK} to encourage uniform embedding-space utilization (Supplementary Methods~\ref{supp:dino_objective}). The multi-crop strategy generates 2 global crops ($256 \times 256$) and 8 local crops ($112 \times 112$) per image (Supplementary Methods~\ref{supp:augmentations}). Unlike standard DINOv3 pretraining on natural images, our pipeline leverages CT-specific data augmentations~\cite{Cardoso2022MONAIAO}: Gaussian noise, random contrast adjustment, simulated low-resolution, and intensity scaling, designed to simulate acquisition variability across scanners and protocols. This domain adaptation allows the model to align with physically meaningful intensity ranges.

We train for $10^{6}$ iterations on 16 NVIDIA B200 GPUs (effective batch size 1,600) using AdamW with cosine learning rate decay, initialized from DINOv3 weights pretrained on ImageNet. Layer-wise learning rate decay (factor 0.9) and a reduced multiplier (0.2) on the patch embedding layer stabilize fine-tuning of the transferred weights.

This is followed by a high-resolution continuation stage ($10^{5}$ iterations) at 384--512\,pixel global crops and 112--224\,pixel local crops, sampled from a multi-resolution schedule. The purpose of this stage is to absorb finer spatial detail (i.e. organ boundaries, small lesions, vasculature) that the base resolution cannot resolve, while a Gram loss~\cite{dinov3} (weight 1.5) with a frozen copy of the base-resolution checkpoint as reference teacher prevents representational drift (Supplementary Methods~\ref{supp:highres}). The resolution-flexible RoPE positional encoding enables this transition without architectural changes. All training hyperparameters are listed in Supplementary Table~\ref{tab:supp_training_config}.

\subsection{Phase 2: 3D volumetric agglomeration}
\label{sec:methods_phase2}

To transfer learned 2D representations into three dimensions, we inflate the 2D patch embedding kernel into a 3D convolutional kernel using the same pseudoinverse-based resampling mechanism, applied with trilinear interpolation along the depth axis. All transformer blocks transfer directly from the 2D checkpoint without modification, as the transformer operates on a flattened sequence of patch tokens regardless of spatial dimensionality. Only the 3D RoPE module is initialized from scratch (Supplementary Methods~\ref{supp:weight_inflation}).

The FlexiCT architecture natively supports both 2D and 3D inputs through dual patch embedding and RoPE modules. Input dimensionality is detected automatically at forward time. After patch embedding and flattening, both paths produce identical token sequence formats, and all transformer blocks process 2D and 3D tokens with self-attention. We deliberately chose full self-attention over windowed or linear alternatives, since full self-attention ensures that every patch can attend to every other patch regardless of spatial distance. This global receptive field is essential for capturing long-range anatomical relationships.

For volumetric pretraining, we replace the 2D multi-crop strategy with 3D random spatial crops: 2 global crops of $160^3$ voxels and 8 local crops of $80^3$ voxels (local scale range 0.1875--0.5 of the global crop dimensions), augmented with random axis flipping along all three spatial axes. Volumes smaller than $160^3$ are padded to the target size. A 3D Region Collaborative Cutout (RCC) masking strategy~\cite{qiu2024mind} partitions each volume into a $3 \times 3 \times 3$ grid of cuboids, within which sub-regions are masked to enforce spatially coherent occlusion patterns that encourage learning of volumetric structure rather than local texture (Supplementary Methods~\ref{supp:augmentations}).

The model is trained for $10^{6}$ iterations (effective batch size 400), initialized from the high-resolution Phase~1 checkpoint. Relative to Phase~1, the DINO global self-distillation loss weight is reduced from 1.0 to 0.5 while the iBOT masked patch prediction loss weight remains at 1.0. This shift emphasizes dense spatial prediction over global representation, learning fine-grained volumetric correspondences during the 3D phase. Full hyperparameters are listed in Supplementary Table~\ref{tab:supp_training_config}.

\subsection{Phase 3: Report-aligned semantic agglomeration}
\label{sec:methods_phase3}

We assembled 95,878 paired CT–report volumes from CT-RATE (47,149)~\cite{ctrate2024}, Merlin (25,489)~\cite{merlin2024}, and INSPECT (23,240). After excluding validation sets used for downstream vision–language evaluation, Phase 3 report-aligned pretraining used 63,710 unique CT–report training pairs. Following the TIPS~\cite{maninis2024tips}, we extend it with a structured negation loss (opposite sentence loss, OSL), which is applied to CT-RATE pairs only. Merlin and INSPECT reports are used in their original unmodified form in accordance with their respective data use agreements. 

\paragraph{Vision-language architecture.}
The vision-language model wraps the Phase~2 backbone with a text encoder (Qwen3-Embedding-0.6B~\cite{Zhang2025Qwen3EA}, 0.6\,B parameters) and contrastive projection heads that map both modalities into a shared 1024-dimensional embedding space. On the vision side, global and patch-level features are concatenated and projected. Text features are obtained via last-token pooling and linear projection. Following the TIPS design, Phase~3 uses a single global crop per volume rather than two, reducing vision compute while retaining local views for patch-level learning. Architectural details are provided in Supplementary Methods~\ref{supp:vlm_architecture}.

\paragraph{Report preprocessing.}
CT-RATE reports were standardized in two stages: (1) a large language model (GPT-5.2) restructured each free-text report into eight anatomical sections with zero-omission constraints, and (2) a second model (Qwen3-30B~\cite{Yang2025Qwen3TR}) extracted per-section positive and negative finding captions that form the training pairs for the opposite sentence loss (Supplementary Methods~\ref{supp:report_preprocessing}). Merlin and INSPECT reports were used in their original form without LLM-based rewriting, as their data use agreements prohibit modification of the released data.

\paragraph{Alignment objectives.}
The primary objective is a symmetric CLIP-style contrastive loss~\cite{clip2021} between paired CT volumes and report text, implemented with memory-efficient ring-topology communication across GPUs (Supplementary Methods~\ref{supp:clip_loss}). For CT-RATE volumes, each sample is paired with a caption randomly sampled from the full structured report or from the extracted positive or negative finding captions. For Merlin and INSPECT volumes, the original unmodified report text is used directly as the caption.

To address the prevalence of negated findings in radiology reports (``no pleural effusion''), which standard contrastive learning struggles to distinguish from their affirmed counterparts, we introduce an opposite sentence loss (OSL). For each CT-RATE sample, we construct pairs of true and rule-negated findings, and train the model to select the factually correct statement given the image embedding. The OSL operates as a binary classification loss over cosine similarity differences between positive and negated text embeddings; it is not applied to Merlin or INSPECT samples, which lack extracted caption pairs. The total Phase~3 objective combines iBOT, CLIP, and OSL losses with weights 1.0, 1.0, and 0.5, respectively (Supplementary Methods~\ref{supp:osl},~\ref{supp:phase3_loss}). The DINO global self-distillation loss is omitted because Phase~3 uses a single global crop per volume, which precludes the cross-view consistency required by the DINO objective; the iBOT masked patch prediction loss, which operates within individual views, remains active to preserve dense spatial learning.

\paragraph{Training.}
We train for $5 \times 10^{5}$ iterations (effective batch size 1,024 on 16 B200 GPUs) with both the vision backbone and text encoder fully trainable. The model is initialized from the Phase~2 checkpoint (vision) and pretrained Qwen3-Embedding-0.6B weights (text). Full hyperparameters are listed in Supplementary Table~\ref{tab:supp_training_config}.

\subsection{Downstream task adaptation}
\label{sec:methods_downstream}

\paragraph{Segmentation.}
All segmentation experiments used the nnU-Net framework~\cite{nnunet2021} with the FlexiCT backbone as the encoder. Features were extracted from four intermediate transformer layers (blocks 3, 7, 11, and 15) and concatenated along the channel dimension to provide multi-scale representations. A lightweight convolutional decoder (PatchDecode) progressively upsampled the features using transposed convolutions to produce dense predictions. For 2D segmentation, we used a multi-scale decoder variant (Primus\_Multiscale) that directly processes the concatenated features; for 3D segmentation, a variant (Primus\_v2) that includes an additional convolutional projection before decoding. The backbone was fine-tuned end-to-end with a dual learning-rate scheme. We report Dice coefficient and surface Dice (SDC) as primary metrics. Per-dataset training configurations are provided in Supplementary Methods~\ref{supp:seg_config}.

\paragraph{Classification.}
We evaluated FlexiCT-2D as a frozen feature extractor for cancer and disease classification on KiTS~\cite{kits2023}, DeepLesion~\cite{deeplesion2018}, LUNA16~\cite{luna16}, and COVIDx-CT~\cite{covidxct2021}, following the Curia evaluation protocol~\cite{curia2024}. For each image, the \texttt{[CLS]} token and mean-pooled patch token features were concatenated and passed through a single linear classification layer trained with SGD and cosine annealing. For 3D datasets (LUNA16), multi-slice features were aggregated using a learned single-head attention module. Feature caching was used to pre-compute backbone embeddings offline. Training hyperparameters per dataset are listed in Supplementary Methods~\ref{supp:cls_config}.

\paragraph{Registration.}
We adopted the training-free DINO-Reg framework~\cite{dinoreg2024} for zero-shot registration. Features were extracted from the last four transformer layers of the FlexiCT-2D backbone via \texttt{get\_intermediate\_layers()}, concatenated along the channel dimension (yielding $864 \times 4 = 3{,}456$ dimensions per spatial position), and reduced to 24 dimensions via PCA fitted on training data and reused across all test cases. The resulting dense feature maps encode anatomically meaningful correspondences that transfer across imaging modalities without explicit multi-modal training.

Registration was optimized using ConvexAdam~\cite{Siebert2024ConvexAdamSD}, a two-stage optimizer that first solves a coupled convex relaxation to obtain a coarse displacement field, then refines it with instance-wise Adam optimization. The objective minimizes the sum of squared differences (SSD) between fixed and moving feature maps with a smoothness regularizer to penalize non-diffeomorphic deformations. Deformation regularity was assessed via the standard deviation of the log Jacobian determinant.

The approach is entirely training-free: the backbone is frozen, no registration-specific fine-tuning or deformation supervision is applied, and only the displacement field is optimized at test time. This makes registration quality a direct indicator of the anatomical correspondence encoded in the pretrained features. We evaluated on both intra-modal (CT--CT) and cross-modal (CT--MR) abdominal registration benchmarks, measuring Dice overlap and 95th-percentile Hausdorff distance (HD95) on organ labels (Supplementary Methods~\ref{supp:reg_config}).

\paragraph{Retrieval and linear probing.}
For patient-level retrieval, \texttt{[CLS]} and mean-pooled patch token features were extracted from FlexiCT-3D, concatenated, projected through the VLM projection head to a 1,024-dimensional space, and $\ell_2$-normalized. To capture both peri-tumoral context and fine-grained lesion detail, two crop sizes were extracted per tumor: a small ROI ($32^3$ voxels) and a large ROI ($64^3$ voxels), both centred on the lesion. Retrieval rankings from the two scales were combined via reciprocal rank fusion with $k = 60$, which assigns each candidate a fused score inversely proportional to its rank under each crop, thereby leveraging complementary information without requiring scale-specific tuning. We report Recall@$K$ and mean average precision (mAP). For linear probing, an $\ell_2$-regularized logistic regression classifier was trained on the extracted features using repeated stratified 5-fold cross-validation with balanced class weighting and regularization strength $C$ selected via grid search. For explainability, linear discriminant analysis (LDA) projected the features into a two-dimensional space, and we measured class separability via silhouette scores and Spearman correlations between discriminant axes and clinical variables such as tumor diameter and histological grade (Supplementary Methods~\ref{supp:retrieval_config}).

\paragraph{VLM zero-shot inference.}
For zero-shot disease classification, we encoded positive and negative text prompts (``\{class name\}.'' and ``No \{class name\}.'') alongside CT volume embeddings, computed cosine similarities against both prompts, and applied a softmax to obtain per-class probabilities. For report retrieval, we computed cosine similarity between volume and text embeddings and reported Recall@$K$ at pool sizes of 32, 64, and 128 in both image-to-text and text-to-image directions. Optimal per-class thresholds were determined from ROC curves on the validation set (Supplementary Methods~\ref{supp:vlm_config}).

\subsection{Evaluation protocols}
\label{sec:methods_eval}

Segmentation performance was measured using the Dice coefficient and the normalized surface Dice coefficient (SDC) at a 2\,mm tolerance. Classification was evaluated by area under the receiver operating characteristic curve (AUC). For vision-language tasks, we report weighted precision, weighted F1 score, accuracy, and macro-averaged AUC (one-vs-rest). Registration quality was assessed by Dice overlap and the 95th-percentile Hausdorff distance (HD95) on organ labels. Retrieval performance was quantified by Recall@$K$ and mean average precision (mAP).

\subsection{Statistical analysis}
\label{sec:methods_stats}
All reported confidence intervals are 95\% bias-corrected and accelerated (BCa) bootstrap intervals computed from 10,000 resampling iterations. BCa intervals account for both bias and skewness in the bootstrap distribution, providing more accurate coverage than percentile intervals, which is particularly important for metrics such as Dice and AUC that exhibit asymmetric sampling distributions near boundary values. Statistical comparisons between FlexiCT and each baseline method used two-sided paired permutation tests (10,000 permutations), in which per-sample metric differences were randomly sign-flipped to construct the null distribution. When multiple comparisons were performed within a task family (e.g., FlexiCT versus each of several baselines across multiple datasets), raw $p$-values were adjusted using the Holm--Bonferroni step-down procedure to control the family-wise error rate at $\alpha = 0.05$. Exact adjusted P values are reported in the corresponding figures, legends or source data. For classification tasks with imbalanced class distributions (e.g., DeepLesion with 8 lesion types, COVIDx-CT with varying prevalence), macro-averaged metrics were used to give equal weight to each class regardless of prevalence.

\section*{Acknowledgements}
This research is supported in part by the National Institutes of Health under Award Number R01EB032680, R01DE033512, R01CA272991, and U54CA274513. The authors acknowledge University of Florida Information Technology Research Computing for computational resources and support provided through the HiPerGator supercomputing cluster.

\clearpage





\section*{Supplementary Information}

\setcounter{section}{0}
\setcounter{subsection}{0}
\renewcommand{\thesection}{S\arabic{section}}
\renewcommand{\thesubsection}{S\arabic{section}.\arabic{subsection}}
\renewcommand{\thetable}{S\arabic{table}}
\renewcommand{\thefigure}{S\arabic{figure}}
\renewcommand{\theequation}{S\arabic{equation}}

\subsection{Data quality control}
\label{supp:data_qc}

Three levels of automated filtering were applied to the assembled pretraining corpus. First, volumes with degenerate geometry were excluded: fewer than eight axial slices, in-plane spacing exceeding 10\,mm, or an axial-to-coronal dimension ratio below $1/3$ (typically scout or localizer acquisitions). Second, a voxel-intensity audit flagged and removed volumes appearing to be binary masks, pre-normalized images, or constant-valued arrays---artefacts of inconsistent upstream processing across heterogeneous public sources. Third, heuristic deduplication identified datasets partially derived from other included sources.

For body cropping during 2D slice extraction, segmentation labels were used to anchor the crop when available for a dataset; otherwise, an intensity-based foreground detector identified the body boundary by thresholding Hounsfield unit values and extracting the largest connected component.

Supplementary Table~\ref{tab:supp_pretrain_datasets} enumerates all 56 constituent datasets grouped by size tier, with volume counts after quality control, anatomical region, country of origin, and source institution. Access URLs and licensing details for each dataset are provided separately in Supplementary Table~\ref{tab:supp_dataset_urls}.

\clearpage
{\footnotesize
\setlength{\LTcapwidth}{0.95\linewidth}
\begin{xltabular}{\textwidth}{@{}l r l l >{\raggedright\arraybackslash}X@{}}
\caption{Pretraining corpus: 56 publicly available CT datasets spanning diverse anatomy, geography, and clinical context.
Volume counts are after deduplication and quality control filtering (see Supplementary Methods~\ref{supp:data_qc}).
Datasets marked with $\dagger$ are multi-site collections.
Anatomical regions are listed where annotated in the source; datasets without region annotation span variable anatomy.}
\label{tab:supp_pretrain_datasets} \\
\toprule
Dataset & Volumes & Region & Country & Institution \\
\midrule
\endfirsthead
\multicolumn{5}{l}{\tablename~\thetable{} \textit{(continued)}} \\
\toprule
Dataset & Volumes & Region & Country & Institution \\
\midrule
\endhead
\midrule
\multicolumn{5}{r}{\textit{Continued on next page}} \\
\endfoot
\bottomrule
\endlastfoot
\multicolumn{5}{l}{\textit{Major contributing sources ($n > 5{,}000$ volumes)}} \\
\addlinespace[2pt]
NLST$^\dagger$ & 132{,}985 & Chest & US & NCI / ACRIN (33 centers) \\
CT-RATE & 47{,}149 & Chest & Turkey & Istanbul Medipol Univ.\ Hosp. \\
Merlin & 25{,}489 & Abdominal & US & Stanford Univ.\ Med.\ Ctr. \\
INSPECT & 23{,}240 & Chest & US & Stanford Univ.\ Hosp. \\
DeepLesion & 5{,}000 & Mixed & US & NIH Clinical Center \\
\addlinespace[4pt]
\multicolumn{5}{l}{\textit{Medium-scale sources ($500$--$5{,}000$ volumes)}} \\
\addlinespace[2pt]
FLARE'23$^\dagger$ & 4{,}100 & Mixed & Canada & Univ.\ of Toronto / UHN \\
StonyBrookChestCT & 2{,}316 & Chest & US & Stony Brook Univ.\ Hosp. \\
Panorama$^\dagger$ & 2{,}238 & Abdominal & Netherlands & Radboud UMC / UMCG \\
AbdominalTrauma$^\dagger$ & 2{,}029 & Mixed & US & RSNA (23 sites, 14 countries) \\
STOIC$^\dagger$ & 2{,}000 & Chest & France & AP-HP (20 univ.\ hospitals) \\
AMOS$^\dagger$ & 1{,}850 & Abdominal & China & Longgang Dist.\ Central Hosp. \\
CT Colonography$^\dagger$ & 1{,}730 & Chest/Abd./Pelv. & US & ACRIN (15 centers) \\
TotalSegmentator V2 & 1{,}203 & Whole body & Switzerland & Univ.\ Hosp.\ Basel \\
HNSCC & 1{,}071 & Head \& neck & US & MD Anderson Cancer Ctr. \\
AbdomenCT-1K$^\dagger$ & 1{,}000 & Abdominal & China & Nanjing Univ. \\
Qin-Headneck & 898 & Head \& neck & US & Univ.\ of Iowa \\
LUNA16$^\dagger$ & 843 & Chest & US & LIDC-IDRI (7 centers) \\
TCGA-KIRC$^\dagger$ & 812 & Mixed & US & NCI TCGA (multi-site) \\
ULS-Radboud-Bone & 744 & Abdominal & Netherlands & Radboud UMC \\
HECTOR$^\dagger$ & 680 & Head \& neck & Switzerland & Univ.\ of Geneva / HES-SO \\
RibFrac & 660 & Mixed & China & Huadong Hosp., Fudan Univ. \\
OPC-Radiomics & 606 & Oropharyngeal & Canada & Princess Margaret Cancer Ctr. \\
CADS-CT-Tri & 585 & Mixed & Germany & TU Munich \\
\addlinespace[4pt]
\multicolumn{5}{l}{\textit{Smaller sources ($< 500$ volumes)}} \\
\addlinespace[2pt]
CADS-BrainCT & 484 & Head & Turkey & Istanbul Medipol Univ. \\
TCGA-BLCA$^\dagger$ & 409 & Mixed & US & NCI TCGA (multi-site) \\
CPTAC-UCEC$^\dagger$ & 393 & Mixed & US & NCI CPTAC (multi-site) \\
TCGA-OV$^\dagger$ & 384 & Mixed & US & NCI TCGA (multi-site) \\
Pediatric-CT-SEG & 358 & Mixed & US & Children's Wisconsin \\
Lung-PET-CT-Dx & 347 & Mixed & China & Harbin Medical Univ. \\
TCGA-UCEC$^\dagger$ & 330 & Mixed & US & NCI TCGA (multi-site) \\
CPTAC-PDA$^\dagger$ & 305 & Mixed & US & NCI CPTAC (multi-site) \\
ACRIN-FLT-Breast$^\dagger$ & 279 & Mixed & US & ACRIN (multi-site) \\
Anti-PD1-Lung & 265 & Mixed & US & MD Anderson Cancer Ctr. \\
CPTAC-ccRCC$^\dagger$ & 258 & Mixed & US & NCI CPTAC (multi-site) \\
CMB-CRC$^\dagger$ & 251 & Mixed & US & NCI Moonshot (NCORP) \\
TCGA-LIHC$^\dagger$ & 242 & Mixed & US & NCI TCGA (multi-site) \\
TCGA-STAD$^\dagger$ & 237 & Mixed & US & NCI TCGA (multi-site) \\
RIDER-Lung-PET-CT & 235 & Mixed & US & Univ.\ of Washington \\
StageII-CRC-CT & 230 & Mixed & China & Fudan Univ.\ Shanghai Cancer Ctr. \\
CRC-Liver-Mets & 197 & Mixed & US & Memorial Sloan Kettering \\
TCGA-LUAD$^\dagger$ & 183 & Mixed & US & NCI TCGA (multi-site) \\
CT-Lymph-Nodes & 174 & Mixed & US & NIH Clinical Center \\
MIDRC-RICORD$^\dagger$ & 163 & Mixed & US & RSNA/STR MIDRC (4 sites) \\
CPTAC-LSCC$^\dagger$ & 159 & Mixed & US & NCI CPTAC (multi-site) \\
CT-ORG$^\dagger$ & 140 & Mixed & US & Stanford Univ. \\
TCGA-LUSC$^\dagger$ & 133 & Mixed & US & NCI TCGA (multi-site) \\
Prostate-Edge-Cases & 131 & Mixed & US & OHSU / VA Portland \\
NSCLC-Radiomics & 131 & Mixed & Netherlands & MAASTRO Clinic \\
ULS-Radboud-Pancreas & 124 & Abdominal & Netherlands & Radboud UMC \\
HCC-TACE-Seg & 103 & Mixed & US & MD Anderson Cancer Ctr. \\
Pancreatic-CT-CBCT & 93 & Mixed & US & Memorial Sloan Kettering \\
BTCV & 47 & Mixed & US & Vanderbilt UMC \\
TCIA-Pancreas-CT & 42 & Mixed & US & NIH Clinical Center \\
CHAOS & 20 & Abdominal & Turkey & Dokuz Eylul Univ.\ Hosp. \\
TCGA-KIRP$^\dagger$ & 19 & Mixed & US & NCI TCGA (multi-site) \\
CPTAC-LUAD & 133 & Lung & US & NCI Clinical Proteomic Tumor Analysis Consortium (CPTAC) \\
\midrule
\textbf{Total (56 datasets)} & \textbf{266{,}227} & & & \\
\end{xltabular}
}

\subsection{Preprocessing details}
\label{supp:preprocessing}

All volumes were reoriented to a canonical anatomical coordinate system (left--posterior--superior, LPS) and resampled using linear interpolation. For 2D pretraining, the through-plane spacing was standardized to 1.5\,mm while preserving native in-plane resolution; individual axial slices were then extracted, body-cropped, and resized to $256 \times 256$ pixels. For 3D pretraining, volumes were resampled to 1.5\,mm isotropic spacing. Voxel intensities were clamped to $[-1000, 1000]$\,HU and normalized per image to zero mean and unit standard deviation at training time.

\subsection{FlexiCT architecture details}
\label{supp:architecture}

The FlexiCT encoder is a Vision Transformer (ViT-Base)~\cite{vit2020} with the following specifications: embedding dimension 864, 16 transformer blocks, 12 attention heads, and a feed-forward expansion ratio of 4, totalling approximately 120\,M parameters. The model uses LayerNorm, GELU activations, stochastic depth with a drop-path rate of 0.2, and layer-scale initialization at $10^{-5}$. Four learnable register tokens~\cite{darcet2023vision} are appended after the \texttt{[CLS]} token.

The flexible patch embedding module (PatchEmbedND) operates at a base patch size of 8. Runtime adjustment to any target patch size $p'$ proceeds via pseudoinverse-based kernel resampling~\cite{beyer2023flexivit}: the base kernel $W \in \mathbb{R}^{C_\text{out} \times C_\text{in} \times p \times p}$ is resampled to $W' \in \mathbb{R}^{C_\text{out} \times C_\text{in} \times p' \times p'}$ by constructing an interpolation matrix $R$ (bicubic for 2D, trilinear for 3D) and computing $W' = W R^{\dagger}$, where $R^{\dagger}$ denotes the Moore--Penrose pseudoinverse. This allows alternating between patch-16 and patch-8 tokenizations during training without modifying network parameters.

For 2D inputs, positional information is encoded with two-axis Rotary Position Embeddings (RoPE)~\cite{dinov3}. For 3D inputs, the RoPE module extends to three spatial axes (requiring the embedding dimension to be divisible by $6 \times$ the number of heads for proper axis allocation). Both 2D and 3D RoPE modules are independent; the 3D module is initialized from scratch when transitioning from Phase~1 to Phase~2.

\subsection{DINOv3 self-supervised objective}
\label{supp:dino_objective}

The DINOv3 framework~\cite{dinov3} uses an exponential moving average (EMA) teacher--student architecture. The teacher is updated with a momentum of 0.994. Three complementary objectives are combined:

\paragraph{DINO loss} Student and teacher \texttt{[CLS]} features are projected through a 3-layer MLP head (hidden dimension 2048, bottleneck dimension 256, output dimension 65{,}536 prototypes). A cross-entropy loss is computed between the student's softmax logits and the teacher's sharpened probability distribution. The teacher output is centered using the Sinkhorn--Knopp algorithm with a temperature that warms from 0.04 to 0.07 over 30 epochs.

\paragraph{iBOT loss} An iBOT masked patch prediction objective~\cite{zhou2021ibot} is applied in parallel: for each global crop, a random subset of 10--50\% of patch tokens is masked (with probability 0.5 per sample), replaced by a learnable mask token, and the student must predict the teacher's patch-level representations through a separate projection head with identical architecture.

\paragraph{KoLeo regularizer} A KoLeo regularizer~\cite{dinov2} with weight 0.1 is applied to the student's pre-head \texttt{[CLS]} features to encourage uniform utilization of the embedding space. Both DINO and iBOT losses are weighted equally (weight 1.0).

\subsection{Multi-crop strategy and augmentations}
\label{supp:augmentations}

\paragraph{2D augmentations.} Each training image yields 2 global crops ($256 \times 256$, scale range 0.32--1.0) and 8 local crops ($112 \times 112$, scale range 0.05--0.32), generated via random resized cropping with bicubic interpolation. Horizontal flipping is applied with probability 0.5. CT-specific intensity augmentations~\cite{Cardoso2022MONAIAO} are applied to the full image before geometric cropping: Gaussian noise ($p = 0.1$, $\sigma = 0.1$), Gaussian smoothing ($p = 0.2$, $\sigma \in [0.5, 1.0]$), random intensity scaling ($p = 0.15$, factor $\in [-0.25, 0.25]$), simulated low resolution ($p = 0.25$, zoom $\in [0.5, 1.0]$), and random contrast adjustment ($p = 0.1$, $\gamma \in [0.7, 1.5]$). Standard colour jittering and solarization are omitted, as they are not meaningful for single-channel CT data.

\paragraph{3D augmentations.} Each training volume yields 2 global crops of size $160^3$ voxels and 8 local crops of size $80^3$ voxels, with local crop scale ranging from 0.1875 to 0.5 of the global crop dimensions. Global crops are generated via random spatial cropping, resized to $160^3$ via trilinear interpolation, and augmented with random axis flipping (probability 0.5 per axis). Volumes smaller than $160^3$ are padded to the target size using the minimum voxel intensity.

\paragraph{Masking.} Both 2D and 3D pretraining use Region Collaborative Cutout (RCC) masking~\cite{qiu2024mind}. In 2D, a $3 \times 3$ grid of bounding boxes is sampled, and sub-regions within each box are masked to reach a target masking ratio. In 3D, this extends to a $3 \times 3 \times 3$ grid of cuboids, with larger boxes processed first and over-cut boxes recovered to maintain spatial coherence.

\subsection{Training hyperparameters}
\label{supp:training_hyperparams}

Supplementary Table~\ref{tab:supp_training_config} summarizes the training configuration for all three pretraining phases.

\begin{table}[h]
\centering
\caption{Training hyperparameters for each pretraining phase.}
\label{tab:supp_training_config}
\footnotesize
\setlength{\tabcolsep}{4pt}
\renewcommand{\arraystretch}{1.05}
\begin{tabularx}{\textwidth}{@{}l *{3}{>{\centering\arraybackslash}X}@{}}
\toprule
Hyperparameter & Phase 1 (2D) & Phase 2 (3D) & Phase 3 (VLM) \\
\midrule
Total iterations & $10^{6}$ & $10^{6}$ & $5 \times 10^{5}$ \\
Optimizer & AdamW & AdamW & AdamW \\
$\beta_1, \beta_2$ & 0.9, 0.999 & 0.9, 0.999 & 0.9, 0.999 \\
Peak learning rate & $2{\times}10^{-4}$ & $2{\times}10^{-4}$ & $2{\times}10^{-4}$ \\
LR schedule & Cosine & Cosine & Cosine (to $2{\times}10^{-5}$) \\
Warmup epochs & 30 & 30 & 30 \\
Weight decay & 0.04 & 0.04 & 0.04\,$\to$\,0.4 \\
Gradient clipping & 3.0 & 3.0 & 3.0 \\
Per-GPU batch size & 100 & 25 & 64 \\
Number of GPUs & 16$\times$B200 & 16$\times$B200 & 16$\times$B200 \\
Effective batch size & 1{,}600 & 400 & 1{,}024 \\
Precision & bfloat16 & bfloat16 & bfloat16 \\
EMA momentum & 0.994 & 0.994 & 0.994\,$\to$\,1.0 \\
DINO loss weight & 1.0 & 0.5 & --- \\
iBOT loss weight & 1.0 & 1.0 & 1.0 \\
CLIP loss weight & --- & --- & 1.0 \\
OSL loss weight & --- & --- & 0.5 \\
KoLeo weight & 0.1 & 0.1 & --- \\
Global crops & $2\,{\times}\,256^{2}$ & $2\,{\times}\,160^{3}$ & $1\,{\times}\,160^{3}$ \\
Local crops & $8\,{\times}\,112^{2}$ & $8\,{\times}\,80^{3}$ & $8\,{\times}\,80^{3}$ \\
Patch embed.\ LR mult. & 0.2 & 0.2 & 0.2 \\
Layer-wise LR decay & 0.9 & 0.9 & 0.9 \\
Initialization & ImageNet pretrained & Phase 1 checkpoint & Phase 2 checkpoint \\
\bottomrule
\end{tabularx}
\end{table}

\subsection{High-resolution continuation (Phase 1)}
\label{supp:highres}

After the initial 2D pretraining, a high-resolution continuation stage runs for $10^{5}$ iterations (100 epochs). Global crop sizes increase to 384--512 pixels and local crop sizes span 112--224 pixels, sampled from a multi-resolution schedule. A Gram loss~\cite{dinov3} with weight 1.5 is introduced using a frozen copy of the initial-phase checkpoint as a reference teacher. The Gram loss operates at the image level with $\ell_2$-normalized features, encouraging the high-resolution model to preserve the representational structure learned at lower resolution. The Gram teacher receives crops at 192--256 pixels without intensity distortions. The learning rate follows a cosine schedule from 0 to $5 \times 10^{-5}$ over the first 10 epochs. EMA momentum is increased to 0.999 and horizontal flipping is disabled.

\subsection{Weight inflation procedure}
\label{supp:weight_inflation}

To transfer 2D representations into three dimensions, the 2D patch embedding kernel $W \in \mathbb{R}^{C_\text{out} \times 1 \times p \times p}$ is inflated to $W' \in \mathbb{R}^{C_\text{out} \times 1 \times p \times p \times p}$ using the same pseudoinverse-based resampling mechanism as PatchEmbedND, applied with trilinear interpolation along the depth axis. All transformer blocks---self-attention layers, feed-forward networks, layer norms, and the \texttt{[CLS]} and register tokens---transfer directly from the 2D checkpoint without modification, as the transformer operates on a flattened sequence of patch tokens regardless of spatial dimensionality. The DINO and iBOT projection heads are also transferred. The 3D RoPE module is initialized from scratch.

\subsection{Vision-language architecture details}
\label{supp:vlm_architecture}

The text encoder is Qwen3-Embedding-0.6B~\cite{Zhang2025Qwen3EA}, a 0.6-billion-parameter decoder-only language model. Text inputs are tokenized with left-padding, truncated to 512 tokens, and processed with flash attention in bfloat16 precision. Token features are pooled via last-token pooling and projected to the shared embedding space through a learned linear layer.

On the vision side, the \texttt{[CLS]} token and the mean of all patch tokens from the student backbone are concatenated along the channel dimension, producing a $2 \times 864 = 1{,}728$-dimensional feature vector, which is then projected to the shared VLM embedding dimension of 1024 via a bias-free linear layer. Both image and text features are $\ell_2$-normalized before computing similarity. A learnable logit scale parameter, initialized to $\ln(1/0.07) \approx 2.66$, controls the temperature of the contrastive loss.

\subsection{Contrastive alignment loss}
\label{supp:clip_loss}

The primary alignment objective is a symmetric CLIP-style contrastive loss~\cite{clip2021}. For a batch of $B$ image--text pairs distributed across $N$ GPUs:
\begin{equation}
\mathcal{L}_{\text{CLIP}} = -\frac{1}{2B}\sum_{i=1}^{B}\left[\log\frac{\exp(\tau \cdot \mathbf{v}_i^\top \mathbf{t}_i)}{\sum_{j}\exp(\tau \cdot \mathbf{v}_i^\top \mathbf{t}_j)} + \log\frac{\exp(\tau \cdot \mathbf{t}_i^\top \mathbf{v}_i)}{\sum_{j}\exp(\tau \cdot \mathbf{t}_i^\top \mathbf{v}_j)}\right]
\label{eq:supp_clip_loss}
\end{equation}
where $\mathbf{v}_i$ and $\mathbf{t}_i$ are the $\ell_2$-normalized image and text embeddings for pair $i$, $\tau = \exp(s)$ is the learnable logit scale, and the summation over $j$ ranges over all $NB$ samples across GPUs. A memory-efficient implementation~\cite{dinov3} based on ring-topology peer-to-peer communication avoids materializing the full $NB \times NB$ similarity matrix.

\subsection{Opposite sentence loss formulation}
\label{supp:osl}

For each CT-RATE training sample, $K = 8$ sentence pairs $(s^{+}_k, s^{-}_k, y_k)$ are constructed, where $y_k \in \{0,1\}$ indicates whether the positive sentence $s^{+}_k$ is a true finding for this patient. The OSL is not applied to Merlin or INSPECT samples, which lack LLM-extracted caption pairs (see Section~\ref{supp:report_preprocessing}). True pairs ($y_k = 1$) are drawn from the patient's own positive findings in a given anatomical section. False pairs ($y_k = 0$) are drawn from a global database of positive findings belonging to other patients in the same section. Each finding sentence is rule-negated to produce its opposite (e.g., ``Pleural effusion.'' $\rightarrow$ ``No pleural effusion.'') using pattern-based templates that handle common radiology constructions.

The OSL is formulated as a binary classification loss:
\begin{equation}
\mathcal{L}_{\text{OSL}} = -\frac{1}{|\mathcal{V}|}\sum_{k \in \mathcal{V}} \left[y_k \log \sigma(\tau(\mathbf{v}^\top \mathbf{t}^{+}_k - \mathbf{v}^\top \mathbf{t}^{-}_k)) + (1 - y_k)\log \sigma(\tau(\mathbf{v}^\top \mathbf{t}^{-}_k - \mathbf{v}^\top \mathbf{t}^{+}_k))\right]
\label{eq:supp_osl_loss}
\end{equation}
where $\sigma$ is the sigmoid function, $\mathcal{V}$ is the set of valid (non-padded) pairs, and $\tau$ is the shared logit scale. The model selects between the positive and negated sentence based on the image embedding, with the target indicating which is factually correct for this patient.

\subsection{Report preprocessing pipeline}
\label{supp:report_preprocessing}

The LLM-based report preprocessing pipeline described below was applied exclusively to CT-RATE~\cite{ctrate2024} reports, whose CC-BY-NC-SA~4.0 license permits adaptation and creation of derivative works. Merlin~\cite{merlin2024} and INSPECT reports were used in their original unmodified form, as their respective Stanford University Dataset Research Use Agreements prohibit modification and creation of derivative works from the released data.

\paragraph{Stage 1: Report restructuring (CT-RATE only).} Each free-text CT-RATE report was restructured into a standardized eight-section format (image quality, lungs and airways, pleura, mediastinum and hila, cardiovascular structures, bones and soft tissues, tubes and devices, upper abdomen) using GPT-5.2 with a system prompt enforcing zero-omission: every clinical statement from the original report must appear exactly once in the appropriate subsection.

\paragraph{Stage 2: Caption extraction (CT-RATE only).} The structured findings were processed by Qwen3-30B to extract per-section lists of concise positive and negative finding captions (e.g., ``Septal thickenings.'' or ``No pulmonary abnormalities detected.''). These extracted captions form the training pairs for the opposite sentence loss.

\paragraph{Training-time caption sampling.} For CT-RATE volumes, each sample is paired with a text caption randomly sampled from three sources: the structured findings section of the full report, or the concatenated positive or negative short captions. Captions are randomly section-shuffled with probability 0.5 to reduce order dependence. For Merlin and INSPECT volumes, the original report text is used directly as the caption without restructuring or caption extraction.

\subsection{Combined Phase 3 objective}
\label{supp:phase3_loss}

The total Phase~3 loss combines self-supervised and vision-language objectives:
\begin{equation}
\mathcal{L}_{\text{Phase\,3}} = \lambda_{\text{iBOT}}\,\mathcal{L}_{\text{iBOT}} + \lambda_{\text{CLIP}}\,\mathcal{L}_{\text{CLIP}} + \lambda_{\text{OSL}}\,\mathcal{L}_{\text{OSL}}
\label{eq:supp_phase3_total}
\end{equation}
with $\lambda_{\text{iBOT}} = \lambda_{\text{CLIP}} = 1.0$ and $\lambda_{\text{OSL}} = 0.5$. For Merlin and INSPECT samples, which lack extracted caption pairs, $\mathcal{L}_{\text{OSL}}$ is masked to zero and only the iBOT and CLIP terms contribute. The DINO global self-distillation loss is omitted because only a single global crop is used per volume, precluding cross-view self-distillation at the global level. KoLeo regularization is disabled during this phase.


\subsection{Segmentation configuration}
\label{supp:seg_config}

All segmentation experiments were implemented in the nnUNet framework~\cite{nnunet2021} using a custom FlexiCT trainer. The segmentation network used FlexiCT as the encoder, initialized from the corresponding 2D or 3D FlexiCT checkpoint. Spatial feature maps were extracted from transformer blocks 3, 7, 11, and 15 and concatenated along the channel dimension. The decoder was a lightweight PatchDecode head: for runtime patch size $p$, it applies $\log_2 p$ upsampling stages, each consisting of a stride-2 transposed convolution, layer normalization, and GELU activation. A final $1 \times 1$ convolution maps the upsampled feature map to the segmentation logits. Deep supervision was disabled for all FlexiCT segmentation runs.
  
\paragraph{2D segmentation.} For 2D experiments, we used the multi-scale decoder. In this configuration, the concatenated intermediate features are passed directly to PatchDecode, so the decoder operates on the full multi-layer feature representation. The backbone and decoder were optimized jointly with AdamW (betas 0.9, 0.98), using a learning rate of $5 \times 10^{-5}$ for both
parameter groups, weight decay $3 \times 10^{-5}$, polynomial learning-rate decay  with power 1.0, 1{,}000 epochs of 250 training iterations each, 50 validation iterations per epoch, and foreground oversampling at 33\%.

\paragraph{3D segmentation.} For 3D experiments, we generally used a lighter decoder. This variant first projects the concatenated feature tensor back to the encoder embedding dimension using a $1 \times 1 \times 1$ convolution followed by layer normalization, and then applies PatchDecode with 3D transposed convolutions. Unless otherwise specified, the optimization, learning-rate schedule, training length, and foreground oversampling settings matched the 2D configuration.


\subsection{Classification configuration}
\label{supp:cls_config}

Classification used the FlexiCT-2D backbone as a frozen feature extractor. For each input, the \texttt{[CLS]} token and mean-pooled patch tokens were concatenated to produce a feature vector of dimensionality $2 \times 864 = 1{,}728$, which was passed through a single linear layer. The backbone was frozen (\texttt{requires\_grad\_(False)}) and features were cached offline to accelerate training. All datasets used images resized to $512 \times 512$ pixels, HU clipping to $[-1000, 1000]$, and patch size 16 at inference. The classification head was trained with SGD (momentum 0.9) and cosine annealing. Supplementary Table~\ref{tab:supp_cls_config} lists per-dataset hyperparameters.

\begin{table}[h]
\centering
\caption{Per-dataset classification training hyperparameters.}
\label{tab:supp_cls_config}
\small
\begin{tabular}{lccccl}
\toprule
Dataset & Classes & Epochs & Batch size & Base LR & Dim \\
\midrule
KiTS & 2 & 50 & 64 & 0.002 & 2D \\
DeepLesion & 8 & 10 & 64 & 0.0005 & 2D \\
LUNA16 & 2 & 50 & 64 & 0.02 & 3D, attention aggregation \\
COVIDx-CT & 3 & 60 & 64 & 0.002 & 2D \\
\bottomrule
\end{tabular}
\end{table}

For LUNA16, which requires volumetric reasoning, multi-slice features were aggregated using a single-head attention module before the classification layer. Learning rates were scaled by batch size ($\text{LR}_\text{scaled} = \text{LR}_\text{base} \times B / 256$).

\subsection{Registration configuration}
\label{supp:reg_config}

Zero-shot registration followed the DINO-Reg framework~\cite{dinoreg2024}. The FlexiCT-2D backbone was run in inference mode with frozen weights. Features from the last four transformer layers were extracted and concatenated along the channel dimension ($864 \times 4 = 3{,}456$ dimensions), and reduced to 24 dimensions using PCA (fitted on training data and reused across test cases). Intensity preprocessing clipped CT values to $[-1000, 1000]$\,HU followed by zero-to-one normalization; for MR images, standard zero-to-one normalization was applied.

Spatial correspondence was optimized using ConvexAdam~\cite{Siebert2024ConvexAdamSD}, which combines coupled convex optimization with instance-wise Adam optimization to minimize the sum of squared differences (SSD) between fixed and moving feature maps. ConvexAdam hyperparameters: learning rate 3, smoothness weight 2, 1{,}000 iterations, smoothing kernel size 7, smoothing passes 5. Deformation regularity was assessed via the standard deviation of the log Jacobian determinant ($\text{LogJacDetStd}$).

We evaluated on CT-MR abdominal registration (AbdomenMRCT, organ labels: liver, spleen, left kidney, right kidney) and CT-CT abdominal registration (AbdomenCTCT). Quality was measured by Dice overlap and HD95 on organ labels.

\subsection{Retrieval and linear probing configuration}
\label{supp:retrieval_config}

\paragraph{Embedding extraction.} For each patient volume, the FlexiCT-3D vision-language model extracted \texttt{[CLS]} and mean-pooled patch token features, which were concatenated, projected through the VLM projection head to 1{,}024 dimensions, and $\ell_2$-normalized. Two crop sizes were used per tumor: a small ROI (32 voxels) and a large ROI (64 voxels), centred on the lesion.

\paragraph{Retrieval.} Cosine similarity was computed between all normalized feature pairs. Rankings from the small and large crops were fused using reciprocal rank fusion (RRF, $k=60$):
\begin{equation}
\text{score}(i,j) = \sum_{\text{sys}} \frac{1}{k_{\text{RRF}} + \text{rank}_{\text{sys}}(i,j)}
\label{eq:rrf}
\end{equation}
Metrics include Recall@$K$ ($K \in \{1, 3, 5, 10\}$), mAP, and per-group retrieval rates with 95\% bootstrap confidence intervals (10{,}000 iterations).

\paragraph{Linear probing.} An $\ell_2$-regularized logistic regression classifier (scikit-learn, LBFGS solver, \texttt{max\_iter=20{,}000}, balanced class weighting) was trained on the extracted features with the regularization parameter $C$ selected via grid search over $\{10^{-4}, 10^{-3}, \ldots, 10^{2}\}$. Features were standardized (zero mean, unit variance) before fitting. Evaluation used repeated stratified 5-fold cross-validation (50/50 train--test split) and reports balanced accuracy, macro F1, macro AUC (one-vs-rest), and macro PR-AUC with 95\% confidence intervals across folds.

\paragraph{LDA explainability.} Linear discriminant analysis projected features into a two-dimensional space ($n_\text{components}=2$). For VLM representations, PCA whitening ($n_\text{components} \in [16, 30]$) was applied before LDA. Separability was quantified using silhouette scores, between/within scatter ratios, and Spearman correlations between discriminant axes and clinical variables (tumor diameter, volume, Gleason grade percentage) with 95\% bootstrap confidence intervals.

\subsection{VLM inference configuration}
\label{supp:vlm_config}

\paragraph{Zero-shot classification.} For each disease class, positive and negative text prompts were constructed as ``\{class name\}.'' and ``No \{class name\}.'' (lowercase). Both prompts were tokenized (max length 768 tokens) and encoded via the text tower. Image embeddings were obtained from the vision projection head (\texttt{[CLS]} + mean patch tokens $\rightarrow$ projection $\rightarrow$ $\ell_2$ normalization). For each class independently, a softmax was applied over the stacked positive and negative cosine similarities (scaled by the learned logit temperature) to obtain the probability of the positive class. Optimal per-class decision thresholds were determined by minimizing the distance to the top-left corner of the ROC curve on the validation set.

\paragraph{Report retrieval.} Volume and text embeddings were $\ell_2$-normalized and cosine similarity was computed for all image--text pairs. Retrieval was evaluated using a pool-based protocol: samples were partitioned into non-overlapping pools of size $N \in \{32, 64, 128\}$, and Recall@$K$ ($K \in \{1, 8\}$) was computed per pool and averaged (both macro and micro). Both image-to-text (I$\rightarrow$T) and text-to-image (T$\rightarrow$I) retrieval directions were evaluated.

Evaluation metrics include weighted precision, weighted F1, per-class accuracy, and macro-averaged AUC (one-vs-rest) for classification; macro/micro Recall@$K$ for retrieval. All metrics include 95\% bootstrap confidence intervals from 10{,}000 resampling iterations.


\begin{table}[h]
\centering
\caption{KiTS23 3D segmentation results by class. Dice similarity coefficient (DSC) and surface Dice coefficient (SDC) with 95\% BCa bootstrap confidence intervals. Overall is the per-case mean across the three classes. Best result per metric in bold.}
\label{tab:supp_kits23}
\footnotesize
\resizebox{\textwidth}{!}{%
\begin{tabular}{llcc}
\toprule
Class & Model & DSC (95\% CI) & SDC (95\% CI) \\
\midrule
\multirow{5}{*}{Kidney} 
        & nnUNet            & 0.974 (0.967--0.977) & 0.953 (0.941--0.960) \\
        & Primus-M          & 0.977 (0.975--0.979) & 0.956 (0.947--0.963) \\
        & Voco              & 0.975 (0.961--0.979) & 0.952 (0.937--0.960) \\
        & CT-FM             & 0.969 (0.948--0.976) & 0.942 (0.916--0.955) \\
        & FlexiCT-3D (Ours) & \textbf{0.979 (0.977--0.980)} & \textbf{0.967 (0.959--0.972)} \\
\midrule
\multirow{5}{*}{Mass}
        & nnUNet            & 0.839 (0.795--0.868) & 0.747 (0.706--0.779) \\
        & Primus-M          & 0.849 (0.809--0.875) & 0.757 (0.711--0.786) \\
        & Voco              & 0.845 (0.805--0.873) & 0.744 (0.705--0.778) \\
        & CT-FM             & 0.823 (0.772--0.858) & 0.726 (0.680--0.764) \\
        & FlexiCT-3D (Ours) & \textbf{0.861 (0.819--0.887)} & \textbf{0.781 (0.742--0.810)} \\
\midrule
\multirow{5}{*}{Tumor}
        & nnUNet            & 0.788 (0.726--0.833) & 0.695 (0.639--0.738) \\
        & Primus-M          & 0.809 (0.757--0.847) & 0.717 (0.666--0.756) \\
        & Voco              & 0.805 (0.751--0.843) & 0.704 (0.656--0.748) \\
        & CT-FM             & 0.757 (0.692--0.809) & 0.663 (0.606--0.714) \\
        & FlexiCT-3D (Ours) & \textbf{0.820 (0.761--0.860)} & \textbf{0.736 (0.677--0.775)} \\
\midrule
\multirow{5}{*}{Overall}
        & nnUNet            & 0.867 (0.835--0.889) & 0.798 (0.771--0.824) \\
        & Primus-M          & 0.878 (0.853--0.898) & 0.810 (0.784--0.832) \\
        & Voco              & 0.875 (0.844--0.895) & 0.800 (0.771--0.823) \\
        & CT-FM             & 0.850 (0.811--0.877) & 0.777 (0.742--0.806) \\
        & FlexiCT-3D (Ours) & \textbf{0.887 (0.859--0.906)} & \textbf{0.828 (0.798--0.848)} \\
\bottomrule
\end{tabular}%
}
\end{table}
        
\begin{table}[p]
        \centering
        \caption{WORD 3D segmentation results by organ. Dice similarity coefficient (DSC) and surface Dice coefficient (SDC) are reported with 95\% BCa bootstrap confidence intervals. Overall is the per-case mean across all 16 organs. Each organ appears once per metric to reduce table length; best result per organ per metric is in bold.}
        \label{tab:supp_word}
        \begingroup
        \setlength{\tabcolsep}{2pt}
        \renewcommand{\arraystretch}{1.5}
        \resizebox{\textwidth}{!}{%
        \begin{tabular}{@{}lccccc@{}}
        \toprule
        Metric / organ & nnUNet & Primus-M & Voco & CT-FM & FlexiCT-3D (Ours) \\
        \midrule
        \multicolumn{6}{@{}l}{\textit{DSC (95\% CI)}} \\
        Liver & \textbf{0.966 (0.964--0.969)} & 0.964 (0.960--0.967) & 0.966 (0.963--0.969) & 0.965 (0.962--0.968) & 0.964 (0.961--0.967) \\
        Spleen & \textbf{0.961 (0.956--0.964)} & 0.954 (0.950--0.957) & 0.959 (0.955--0.962) & 0.959 (0.955--0.962) & 0.956 (0.951--0.959) \\
        Left kidney & \textbf{0.959 (0.954--0.962)} & 0.953 (0.949--0.957) & 0.957 (0.952--0.960) & 0.955 (0.950--0.959) & 0.955 (0.950--0.959) \\
        Right kidney & \textbf{0.958 (0.954--0.962)} & 0.953 (0.950--0.957) & 0.957 (0.953--0.960) & 0.956 (0.952--0.960) & 0.954 (0.949--0.957) \\
        Stomach & 0.919 (0.894--0.935) & 0.921 (0.897--0.932) & 0.923 (0.899--0.937) & 0.918 (0.880--0.932) & \textbf{0.930 (0.916--0.940)} \\
        Gallbladder & \textbf{0.768 (0.629--0.837)} & 0.662 (0.519--0.742) & 0.737 (0.595--0.811) & 0.751 (0.616--0.828) & 0.743 (0.615--0.823) \\
        Esophagus & \textbf{0.802 (0.768--0.827)} & 0.775 (0.743--0.799) & 0.793 (0.757--0.818) & 0.785 (0.753--0.808) & 0.791 (0.767--0.810) \\
        Pancreas & \textbf{0.848 (0.827--0.865)} & 0.822 (0.801--0.840) & 0.841 (0.823--0.857) & 0.839 (0.813--0.856) & 0.837 (0.813--0.853) \\
        Duodenum & 0.683 (0.618--0.730) & 0.674 (0.601--0.720) & \textbf{0.702 (0.652--0.740)} & 0.668 (0.611--0.723) & 0.699 (0.632--0.743) \\
        Colon & \textbf{0.836 (0.783--0.868)} & 0.799 (0.740--0.837) & 0.825 (0.754--0.859) & 0.821 (0.727--0.858) & 0.834 (0.771--0.865) \\
        Intestine & \textbf{0.870 (0.840--0.887)} & 0.841 (0.805--0.861) & 0.863 (0.827--0.881) & 0.864 (0.834--0.881) & 0.869 (0.839--0.884) \\
        Adrenal & \textbf{0.724 (0.677--0.758)} & 0.669 (0.619--0.705) & 0.721 (0.674--0.756) & 0.714 (0.667--0.747) & 0.684 (0.628--0.722) \\
        Rectum & 0.760 (0.642--0.807) & 0.754 (0.685--0.795) & 0.769 (0.686--0.809) & 0.752 (0.635--0.795) & \textbf{0.775 (0.700--0.805)} \\
        Bladder & 0.925 (0.828--0.952) & 0.923 (0.861--0.946) & 0.919 (0.814--0.950) & 0.928 (0.865--0.950) & \textbf{0.929 (0.850--0.951)} \\
        Head of femur (L) & 0.578 (0.457--0.683) & \textbf{0.880 (0.781--0.915)} & 0.577 (0.445--0.688) & 0.680 (0.643--0.723) & 0.872 (0.784--0.912) \\
        Head of femur (R) & 0.674 (0.563--0.751) & \textbf{0.898 (0.858--0.921)} & 0.741 (0.635--0.801) & 0.468 (0.357--0.573) & 0.879 (0.823--0.911) \\
        Overall & 0.827 (0.793--0.845) & 0.840 (0.814--0.857) & 0.828 (0.798--0.845) & 0.814 (0.780--0.831) & \textbf{0.854 (0.829--0.870)} \\
        \midrule
        \multicolumn{6}{@{}l}{\textit{SDC (95\% CI)}} \\
        Liver & \textbf{0.772 (0.747--0.795)} & 0.751 (0.727--0.773) & 0.766 (0.741--0.791) & 0.758 (0.735--0.782) & 0.746 (0.724--0.769) \\
        Spleen & \textbf{0.876 (0.855--0.893)} & 0.844 (0.825--0.861) & 0.867 (0.847--0.883) & 0.870 (0.852--0.885) & 0.843 (0.824--0.860) \\
        Left kidney & \textbf{0.867 (0.844--0.885)} & 0.837 (0.812--0.855) & 0.858 (0.835--0.874) & 0.854 (0.833--0.872) & 0.841 (0.816--0.858) \\
        Right kidney & \textbf{0.861 (0.840--0.878)} & 0.840 (0.822--0.859) & 0.856 (0.839--0.871) & 0.853 (0.835--0.872) & 0.836 (0.817--0.853) \\
        Stomach & \textbf{0.704 (0.667--0.739)} & 0.668 (0.627--0.701) & 0.703 (0.667--0.734) & 0.694 (0.654--0.726) & 0.690 (0.659--0.720) \\
        Gallbladder & \textbf{0.695 (0.559--0.766)} & 0.562 (0.448--0.638) & 0.639 (0.514--0.710) & 0.663 (0.520--0.734) & 0.655 (0.529--0.727) \\
        Esophagus & \textbf{0.719 (0.680--0.759)} & 0.668 (0.628--0.701) & 0.705 (0.667--0.743) & 0.683 (0.644--0.716) & 0.680 (0.639--0.713) \\
        Pancreas & \textbf{0.682 (0.652--0.711)} & 0.627 (0.602--0.657) & 0.665 (0.635--0.694) & 0.654 (0.620--0.683) & 0.650 (0.625--0.673) \\
        Duodenum & 0.541 (0.489--0.587) & 0.495 (0.438--0.537) & \textbf{0.546 (0.499--0.586)} & 0.521 (0.466--0.566) & 0.524 (0.471--0.564) \\
        Colon & \textbf{0.668 (0.599--0.708)} & 0.577 (0.517--0.622) & 0.643 (0.569--0.685) & 0.639 (0.559--0.678) & 0.636 (0.564--0.675) \\
        Intestine & \textbf{0.717 (0.673--0.746)} & 0.654 (0.604--0.689) & 0.699 (0.653--0.730) & 0.699 (0.657--0.727) & 0.689 (0.646--0.717) \\
        Adrenal & \textbf{0.718 (0.663--0.766)} & 0.647 (0.583--0.692) & 0.716 (0.660--0.762) & 0.697 (0.640--0.741) & 0.651 (0.589--0.698) \\
        Rectum & \textbf{0.552 (0.480--0.595)} & 0.516 (0.463--0.562) & 0.543 (0.473--0.589) & 0.519 (0.449--0.565) & 0.541 (0.484--0.584) \\
        Bladder & \textbf{0.747 (0.664--0.782)} & 0.707 (0.639--0.745) & 0.730 (0.643--0.763) & 0.729 (0.652--0.761) & 0.734 (0.664--0.764) \\
        Head of femur (L) & 0.449 (0.342--0.536) & 0.686 (0.620--0.726) & 0.477 (0.379--0.564) & 0.523 (0.484--0.569) & \textbf{0.690 (0.612--0.738)} \\
        Head of femur (R) & 0.527 (0.450--0.588) & \textbf{0.703 (0.652--0.747)} & 0.581 (0.514--0.632) & 0.377 (0.296--0.457) & 0.694 (0.639--0.739) \\
        Overall & 0.693 (0.658--0.714) & 0.674 (0.645--0.696) & 0.687 (0.655--0.708) & 0.671 (0.631--0.692) & \textbf{0.694 (0.666--0.715)} \\
        \bottomrule
        \end{tabular}%
        }
        \endgroup
        \end{table}
        
\begin{table}[h]
\centering
\caption{MSD-Liver 3D segmentation results by class. Dice similarity coefficient (DSC) and surface Dice coefficient (SDC) with 95\% BCa bootstrap confidence intervals. Overall is the per-case mean across the two classes. Best result per class per metric in bold.}
\label{tab:supp_msd_liver}
\footnotesize
\resizebox{\textwidth}{!}{%
\begin{tabular}{llcc}
\toprule
Class & Model & DSC (95\% CI) & SDC (95\% CI) \\
\midrule
\multirow{5}{*}{Liver}
        & nnUNet            & 0.945 (0.913--0.958) & 0.683 (0.634--0.724) \\
        & Primus-M          & 0.955 (0.933--0.963) & 0.696 (0.649--0.736) \\
        & Voco              & 0.950 (0.921--0.960) & 0.683 (0.637--0.723) \\
        & CT-FM             & 0.934 (0.903--0.950) & 0.661 (0.616--0.703) \\
        & FlexiCT-3D (Ours) & \textbf{0.960 (0.951--0.965)} & \textbf{0.700 (0.651--0.744)} \\
\midrule
\multirow{5}{*}{Cancer}
        & nnUNet            & 0.657 (0.538--0.737) & 0.528 (0.434--0.610) \\
        & Primus-M          & 0.639 (0.514--0.728) & 0.496 (0.397--0.590) \\
        & Voco              & 0.631 (0.508--0.720) & 0.496 (0.397--0.586) \\
        & CT-FM             & 0.650 (0.555--0.720) & 0.494 (0.411--0.561) \\
        & FlexiCT-3D (Ours) & \textbf{0.775 (0.673--0.870)} & \textbf{0.639 (0.584--0.701)} \\
\midrule
\multirow{5}{*}{Overall}
        & nnUNet            & 0.801 (0.739--0.838) & 0.605 (0.550--0.656) \\
        & Primus-M          & 0.803 (0.740--0.847) & 0.601 (0.548--0.655) \\
        & Voco              & 0.791 (0.725--0.832) & 0.590 (0.536--0.642) \\
        & CT-FM             & 0.797 (0.744--0.834) & 0.581 (0.531--0.628) \\
        & FlexiCT-3D (Ours) & \textbf{0.867 (0.812--0.917)} & \textbf{0.670 (0.617--0.723)} \\
\bottomrule
\end{tabular}%
}
\end{table}

\begin{table}[h]
\centering
\caption{MSD-Lung 3D segmentation results. Dice similarity coefficient (DSC) and surface Dice coefficient (SDC) with 95\% BCa bootstrap confidence intervals. Best result per metric in bold.}
\label{tab:supp_msd_lung}
\footnotesize
\resizebox{\textwidth}{!}{%
\begin{tabular}{lcc}
\toprule
Model & DSC (95\% CI) & SDC (95\% CI) \\
\midrule
nnUNet & 0.715 (0.607--0.805) & 0.497 (0.401--0.585) \\
Primus-M & 0.704 (0.528--0.834) & 0.501 (0.360--0.620) \\
Voco & 0.656 (0.507--0.780) & 0.463 (0.348--0.564) \\
CT-FM & 0.701 (0.525--0.835) & 0.525 (0.376--0.656) \\
FlexiCT-3D (Ours) & \textbf{0.738 (0.604--0.832)} & \textbf{0.533 (0.417--0.637)} \\
\bottomrule
\end{tabular}%
}
\end{table}

\begin{table}[h]
\centering
\caption{MSD-Pancreas 3D segmentation results by class. Dice similarity coefficient (DSC) and surface Dice coefficient (SDC) with 95\% BCa bootstrap confidence intervals. Overall is the per-case mean across the two classes. Best result per class per metric in bold.}
\label{tab:supp_msd_pancreas}
\footnotesize
\resizebox{\textwidth}{!}{%
\begin{tabular}{llcc}
\toprule
Class & Model & DSC (95\% CI) & SDC (95\% CI) \\
\midrule
\multirow{5}{*}{Pancreas}
        & nnUNet            & 0.824 (0.797--0.841) & 0.613 (0.580--0.640) \\
        & Primus-M          & 0.797 (0.768--0.816) & 0.555 (0.522--0.586) \\
        & Voco              & 0.811 (0.788--0.830) & 0.584 (0.549--0.614) \\
        & CT-FM             & 0.818 (0.791--0.838) & 0.590 (0.560--0.616) \\
        & FlexiCT-3D (Ours) & \textbf{0.839 (0.825--0.854)} & \textbf{0.629 (0.599--0.652)} \\
\midrule
\multirow{5}{*}{Cancer}
        & nnUNet            & 0.499 (0.417--0.568) & 0.341 (0.278--0.419) \\
        & Primus-M          & 0.434 (0.350--0.513) & 0.296 (0.237--0.367) \\
        & Voco              & 0.433 (0.348--0.513) & 0.296 (0.234--0.371) \\
        & CT-FM             & 0.420 (0.338--0.502) & 0.271 (0.213--0.343) \\
        & FlexiCT-3D (Ours) & \textbf{0.568 (0.491--0.637)} & \textbf{0.405 (0.348--0.471)} \\
\midrule
\multirow{5}{*}{Overall}
        & nnUNet            & 0.661 (0.619--0.705) & 0.477 (0.435--0.517) \\
        & Primus-M          & 0.615 (0.568--0.659) & 0.425 (0.387--0.473) \\
        & Voco              & 0.622 (0.575--0.667) & 0.440 (0.402--0.485) \\
        & CT-FM             & 0.619 (0.572--0.662) & 0.431 (0.396--0.468) \\
        & FlexiCT-3D (Ours) & \textbf{0.703 (0.658--0.746)} & \textbf{0.517 (0.473--0.561)} \\
\bottomrule
\end{tabular}%
}
\end{table}

\begin{table}[h]
\centering
\caption{AutoPET~II 3D segmentation results (metabolically active tumor). Dice similarity coefficient (DSC) and surface Dice coefficient (SDC) with 95\% BCa bootstrap confidence intervals. Best result per metric in bold.}
\label{tab:supp_autopetii}
\footnotesize
\resizebox{\textwidth}{!}{%
\begin{tabular}{lcc}
\toprule
Model & DSC (95\% CI) & SDC (95\% CI) \\
\midrule
        nnUNet & 0.336 (0.282--0.390) & 0.202 (0.167--0.237) \\
        Primus-M & 0.382 (0.323--0.441) & 0.231 (0.192--0.269) \\
        Voco & 0.207 (0.166--0.249) & 0.097 (0.077--0.118) \\
        CT-FM & 0.326 (0.273--0.381) & 0.187 (0.153--0.221) \\
        FlexiCT-3D (Ours) & \textbf{0.605 (0.538--0.671)} & \textbf{0.369 (0.323--0.416)} \\
\bottomrule
\end{tabular}%
}
\end{table}

\clearpage
{\footnotesize
\setlength{\tabcolsep}{4pt}
\begin{longtable}{@{}l l c c@{}}
\caption{AMOS22 2D segmentation results by organ. Dice similarity coefficient (DSC) and surface Dice coefficient (SDC) with 95\% BCa bootstrap confidence intervals. Overall is the per-case mean across all 15 organs. Best result per organ per metric in bold.}\label{tab:supp_seg_amos22}\\
\toprule
Organ & Model & DSC (95\% CI) & SDC (95\% CI) \\
\midrule
\endfirsthead
\multicolumn{4}{l}{\textit{Supplementary Table 10 (continued)}}\\
\toprule
Organ & Model & DSC (95\% CI) & SDC (95\% CI) \\
\midrule
\endhead
\midrule
\multicolumn{4}{r}{\textit{Continued on next page}}\\
\endfoot
\bottomrule
\endlastfoot
        \multirow{5}{*}{Spleen}
         & UNet              & 0.940 (0.882--0.966) & 0.862 (0.805--0.893) \\
         & DINOv3            & 0.939 (0.880--0.966) & 0.861 (0.806--0.891) \\
         & Curia             & 0.939 (0.883--0.965) & 0.862 (0.807--0.893) \\
         & BiomedCLIP        & 0.928 (0.864--0.954) & 0.837 (0.775--0.871) \\
         & FlexiCT-2D (Ours) & \textbf{0.945 (0.883--0.972)} & \textbf{0.888 (0.832--0.918)} \\
        \midrule
        \multirow{5}{*}{Right kidney}
         & UNet              & 0.948 (0.902--0.960) & 0.858 (0.819--0.884) \\
         & DINOv3            & 0.950 (0.926--0.960) & 0.856 (0.817--0.879) \\
         & Curia             & 0.950 (0.927--0.960) & 0.860 (0.825--0.885) \\
         & BiomedCLIP        & 0.942 (0.912--0.956) & 0.845 (0.802--0.875) \\
         & FlexiCT-2D (Ours) & \textbf{0.965 (0.958--0.969)} & \textbf{0.885 (0.850--0.904)} \\
        \midrule
        \multirow{5}{*}{Left kidney}
         & UNet              & 0.937 (0.882--0.957) & 0.849 (0.803--0.879) \\
         & DINOv3            & 0.941 (0.888--0.960) & 0.856 (0.806--0.882) \\
         & Curia             & 0.945 (0.876--0.961) & 0.862 (0.820--0.889) \\
         & BiomedCLIP        & 0.932 (0.882--0.954) & 0.843 (0.792--0.874) \\
         & FlexiCT-2D (Ours) & \textbf{0.949 (0.885--0.967)} & \textbf{0.879 (0.831--0.905)} \\
        \midrule
        \multirow{5}{*}{Gallbladder}
         & UNet              & 0.811 (0.751--0.849) & 0.691 (0.629--0.745) \\
         & DINOv3            & 0.799 (0.736--0.840) & 0.675 (0.614--0.731) \\
         & Curia             & 0.783 (0.722--0.829) & 0.658 (0.589--0.715) \\
         & BiomedCLIP        & 0.738 (0.668--0.795) & 0.616 (0.548--0.679) \\
         & FlexiCT-2D (Ours) & \textbf{0.849 (0.795--0.881)} & \textbf{0.733 (0.671--0.786)} \\
        \midrule
        \multirow{5}{*}{Esophagus}
         & UNet              & 0.830 (0.799--0.850) & 0.764 (0.718--0.800) \\
         & DINOv3            & 0.811 (0.782--0.833) & 0.731 (0.687--0.768) \\
         & Curia             & 0.822 (0.795--0.841) & 0.747 (0.700--0.782) \\
         & BiomedCLIP        & 0.806 (0.780--0.826) & 0.719 (0.670--0.756) \\
         & FlexiCT-2D (Ours) & \textbf{0.836 (0.813--0.854)} & \textbf{0.764 (0.722--0.799)} \\
        \midrule
        \multirow{5}{*}{Liver}
         & UNet              & 0.970 (0.961--0.975) & 0.832 (0.797--0.856) \\
         & DINOv3            & 0.972 (0.966--0.975) & 0.832 (0.803--0.854) \\
         & Curia             & 0.973 (0.967--0.976) & 0.840 (0.814--0.864) \\
         & BiomedCLIP        & 0.968 (0.956--0.973) & 0.823 (0.787--0.849) \\
         & FlexiCT-2D (Ours) & \textbf{0.978 (0.972--0.980)} & \textbf{0.864 (0.842--0.883)} \\
        \midrule
        \multirow{5}{*}{Stomach}
         & UNet              & 0.862 (0.805--0.895) & 0.674 (0.621--0.721) \\
         & DINOv3            & 0.867 (0.813--0.897) & 0.676 (0.620--0.719) \\
         & Curia             & 0.876 (0.821--0.903) & 0.684 (0.633--0.732) \\
         & BiomedCLIP        & 0.853 (0.802--0.886) & 0.647 (0.590--0.696) \\
         & FlexiCT-2D (Ours) & \textbf{0.896 (0.845--0.923)} & \textbf{0.737 (0.680--0.778)} \\
        \midrule
        \multirow{5}{*}{Aorta}
         & UNet              & 0.950 (0.942--0.956) & 0.885 (0.859--0.906) \\
         & DINOv3            & 0.945 (0.937--0.951) & 0.869 (0.842--0.890) \\
         & Curia             & 0.946 (0.939--0.952) & 0.865 (0.839--0.886) \\
         & BiomedCLIP        & 0.943 (0.934--0.950) & 0.862 (0.833--0.885) \\
         & FlexiCT-2D (Ours) & \textbf{0.953 (0.946--0.959)} & \textbf{0.895 (0.871--0.915)} \\
        \midrule
        \multirow{5}{*}{Postcava}
         & UNet              & 0.883 (0.867--0.897) & 0.717 (0.686--0.745) \\
         & DINOv3            & 0.870 (0.852--0.884) & 0.689 (0.654--0.719) \\
         & Curia             & 0.876 (0.858--0.890) & 0.703 (0.673--0.732) \\
         & BiomedCLIP        & 0.863 (0.843--0.878) & 0.670 (0.636--0.701) \\
         & FlexiCT-2D (Ours) & \textbf{0.893 (0.878--0.905)} & \textbf{0.732 (0.699--0.759)} \\
        \midrule
        \multirow{5}{*}{Pancreas}
         & UNet              & 0.825 (0.791--0.848) & 0.663 (0.625--0.699) \\
         & DINOv3            & 0.808 (0.765--0.836) & 0.640 (0.595--0.677) \\
         & Curia             & 0.819 (0.777--0.842) & 0.658 (0.615--0.694) \\
         & BiomedCLIP        & 0.798 (0.757--0.825) & 0.625 (0.579--0.660) \\
         & FlexiCT-2D (Ours) & \textbf{0.852 (0.827--0.872)} & \textbf{0.706 (0.664--0.739)} \\
        \midrule
        \multirow{5}{*}{Right adrenal}
         & UNet              & 0.735 (0.682--0.767) & 0.767 (0.711--0.808) \\
         & DINOv3            & 0.709 (0.660--0.741) & 0.743 (0.691--0.784) \\
         & Curia             & 0.720 (0.671--0.750) & 0.753 (0.701--0.793) \\
         & BiomedCLIP        & 0.675 (0.627--0.710) & 0.707 (0.655--0.752) \\
         & FlexiCT-2D (Ours) & \textbf{0.738 (0.688--0.765)} & \textbf{0.774 (0.724--0.812)} \\
        \midrule
        \multirow{5}{*}{Left adrenal}
         & UNet              & 0.749 (0.712--0.778) & 0.764 (0.720--0.802) \\
         & DINOv3            & 0.723 (0.686--0.752) & 0.737 (0.692--0.777) \\
         & Curia             & 0.734 (0.697--0.764) & 0.749 (0.702--0.790) \\
         & BiomedCLIP        & 0.677 (0.625--0.716) & 0.693 (0.638--0.741) \\
         & FlexiCT-2D (Ours) & \textbf{0.757 (0.720--0.785)} & \textbf{0.775 (0.730--0.813)} \\
        \midrule
        \multirow{5}{*}{Duodenum}
         & UNet              & 0.767 (0.733--0.795) & 0.601 (0.559--0.644) \\
         & DINOv3            & 0.744 (0.710--0.772) & 0.572 (0.528--0.613) \\
         & Curia             & 0.751 (0.714--0.780) & 0.581 (0.538--0.621) \\
         & BiomedCLIP        & 0.725 (0.685--0.756) & 0.544 (0.503--0.584) \\
         & FlexiCT-2D (Ours) & \textbf{0.799 (0.767--0.823)} & \textbf{0.646 (0.603--0.686)} \\
        \midrule
        \multirow{5}{*}{Bladder}
         & UNet              & 0.873 (0.821--0.901) & 0.772 (0.728--0.805) \\
         & DINOv3            & 0.883 (0.835--0.909) & 0.786 (0.745--0.818) \\
         & Curia             & 0.885 (0.836--0.911) & 0.791 (0.747--0.823) \\
         & BiomedCLIP        & 0.877 (0.824--0.905) & 0.774 (0.729--0.808) \\
         & FlexiCT-2D (Ours) & \textbf{0.920 (0.877--0.938)} & \textbf{0.843 (0.800--0.867)} \\
        \midrule
        \multirow{5}{*}{Prostate/Uterus}
         & UNet              & 0.847 (0.791--0.869) & 0.634 (0.589--0.671) \\
         & DINOv3            & 0.848 (0.787--0.872) & 0.643 (0.596--0.678) \\
         & Curia             & 0.851 (0.794--0.873) & 0.647 (0.596--0.683) \\
         & BiomedCLIP        & 0.835 (0.775--0.861) & 0.614 (0.568--0.657) \\
         & FlexiCT-2D (Ours) & \textbf{0.875 (0.812--0.894)} & \textbf{0.694 (0.649--0.725)} \\
        \midrule
        \multirow{5}{*}{Overall}
         & UNet              & 0.861 (0.843--0.876) & 0.753 (0.720--0.780) \\
         & DINOv3            & 0.853 (0.835--0.867) & 0.741 (0.709--0.767) \\
         & Curia             & 0.857 (0.840--0.871) & 0.748 (0.716--0.775) \\
         & BiomedCLIP        & 0.836 (0.815--0.854) & 0.718 (0.686--0.749) \\
         & FlexiCT-2D (Ours) & \textbf{0.879 (0.865--0.891)} & \textbf{0.784 (0.750--0.807)} \\
\end{longtable}
}

\begin{table}[h]
\centering
\caption{TotalSegmentator 2D segmentation results by anatomical group. Dice similarity coefficient (DSC) and surface Dice coefficient (SDC) with 95\% BCa bootstrap confidence intervals. Per-case macro aggregation within each group (mean over classes present in each case, then mean over cases). Best result per group per metric in bold.\textsuperscript{a}}
\label{tab:supp_seg_totalseg}
\footnotesize
\resizebox{\textwidth}{!}{%
\begin{tabular}{llcc}
\toprule
Group & Model & DSC (95\% CI) & SDC (95\% CI) \\
\midrule
\multirow{5}{*}{Organs (24)}
        & UNet\textsuperscript{a} & 0.788 (0.764--0.805) & 0.523 (0.501--0.542) \\
        & DINOv3            & 0.766 (0.745--0.784) & 0.460 (0.443--0.475) \\
        & Curia             & 0.812 (0.790--0.827) & 0.548 (0.528--0.566) \\
        & BiomedCLIP        & 0.766 (0.742--0.784) & 0.462 (0.444--0.479) \\
        & FlexiCT-2D (Ours) & \textbf{0.840 (0.822--0.855)} & \textbf{0.588 (0.567--0.606)} \\
\midrule
\multirow{5}{*}{Vertebrae (26)}
        & UNet\textsuperscript{a} & 0.798 (0.769--0.822) & 0.662 (0.636--0.685) \\
        & DINOv3            & 0.768 (0.742--0.790) & 0.557 (0.535--0.576) \\
        & Curia             & 0.813 (0.784--0.835) & 0.674 (0.649--0.695) \\
        & BiomedCLIP        & 0.782 (0.756--0.805) & 0.592 (0.569--0.613) \\
        & FlexiCT-2D (Ours) & \textbf{0.846 (0.822--0.868)} & \textbf{0.724 (0.701--0.746)} \\
\midrule
\multirow{5}{*}{Cardiac (18)}
        & UNet\textsuperscript{a} & 0.782 (0.759--0.804) & 0.550 (0.529--0.568) \\
        & DINOv3            & 0.732 (0.709--0.752) & 0.445 (0.427--0.461) \\
        & Curia             & 0.804 (0.781--0.822) & 0.570 (0.552--0.589) \\
        & BiomedCLIP        & 0.726 (0.698--0.748) & 0.446 (0.426--0.464) \\
        & FlexiCT-2D (Ours) & \textbf{0.830 (0.804--0.848)} & \textbf{0.611 (0.589--0.630)} \\
\midrule
\multirow{5}{*}{Musculoskeletal (23)}
        & UNet\textsuperscript{a} & 0.787 (0.762--0.808) & 0.521 (0.500--0.539) \\
        & DINOv3            & 0.753 (0.728--0.775) & 0.451 (0.434--0.468) \\
        & Curia             & 0.812 (0.786--0.832) & 0.553 (0.534--0.570) \\
        & BiomedCLIP        & 0.746 (0.721--0.769) & 0.468 (0.448--0.487) \\
        & FlexiCT-2D (Ours) & \textbf{0.851 (0.826--0.872)} & \textbf{0.625 (0.604--0.644)} \\
\midrule
\multirow{5}{*}{Ribs (26)}
        & UNet\textsuperscript{a} & 0.790 (0.755--0.820) & 0.723 (0.689--0.753) \\
        & DINOv3            & 0.764 (0.731--0.792) & 0.655 (0.627--0.682) \\
        & Curia             & 0.798 (0.763--0.829) & 0.726 (0.693--0.753) \\
        & BiomedCLIP        & 0.762 (0.729--0.791) & 0.655 (0.623--0.681) \\
        & FlexiCT-2D (Ours) & \textbf{0.820 (0.785--0.851)} & \textbf{0.763 (0.730--0.793)} \\
\midrule
\multirow{5}{*}{Overall (117)}
        & UNet\textsuperscript{a} & 0.793 (0.775--0.809) & 0.600 (0.580--0.616) \\
        & DINOv3            & 0.762 (0.744--0.777) & 0.521 (0.504--0.535) \\
        & Curia             & 0.811 (0.792--0.827) & 0.618 (0.600--0.633) \\
        & BiomedCLIP        & 0.762 (0.741--0.778) & 0.531 (0.513--0.547) \\
        & FlexiCT-2D (Ours) & \textbf{0.842 (0.825--0.857)} & \textbf{0.668 (0.650--0.682)} \\
\bottomrule
\multicolumn{4}{l}{\textsuperscript{a}\scriptsize UNet trained on $n=229$ labelled volumes; all foundation model methods use the standard split.} \\
\end{tabular}%
}
\end{table}

\begin{table}[h]
\centering
\caption{CT--CT intra-modal registration per-organ Dice similarity coefficient (DSC) aggregated across all 45 validation pairs (5-fold CV). Mean (95\% bootstrap CI). Best per organ in bold.}
\label{tab:supp_reg_ctct_dsc}
\footnotesize
\resizebox{\textwidth}{!}{%
\begin{tabular}{lccccc}
\toprule
Organ & VoxelMorph & DINOv2 & DINOv3 & Curia & FlexiCT-2D (Ours) \\
\midrule
Spleen       & 0.412 (0.371--0.451) & 0.689 (0.637--0.737) & 0.447 (0.399--0.491) & 0.481 (0.440--0.522) & \textbf{0.836 (0.812--0.858)} \\
R. kidney    & 0.344 (0.289--0.399) & 0.718 (0.667--0.762) & 0.342 (0.289--0.396) & 0.384 (0.331--0.444) & \textbf{0.723 (0.679--0.762)} \\
L. kidney    & 0.349 (0.306--0.392) & 0.718 (0.674--0.757) & 0.356 (0.310--0.402) & 0.402 (0.358--0.448) & \textbf{0.785 (0.750--0.816)} \\
Gallbladder\textsuperscript{a} & 0.038 (0.011--0.070) & 0.129 (0.082--0.186) & 0.053 (0.014--0.105) & 0.043 (0.014--0.075) & \textbf{0.187 (0.127--0.252)} \\
Esophagus    & 0.225 (0.170--0.278) & 0.402 (0.349--0.453) & 0.255 (0.197--0.315) & 0.278 (0.220--0.337) & \textbf{0.471 (0.423--0.516)} \\
Liver        & 0.621 (0.593--0.648) & 0.845 (0.826--0.862) & 0.680 (0.649--0.709) & 0.692 (0.665--0.716) & \textbf{0.901 (0.894--0.909)} \\
Stomach      & 0.241 (0.199--0.284) & 0.440 (0.383--0.493) & 0.290 (0.241--0.338) & 0.284 (0.236--0.330) & \textbf{0.560 (0.508--0.609)} \\
Aorta        & 0.323 (0.286--0.360) & 0.612 (0.576--0.648) & 0.314 (0.269--0.364) & 0.371 (0.333--0.413) & \textbf{0.665 (0.633--0.695)} \\
IVC          & 0.350 (0.317--0.387) & 0.538 (0.513--0.562) & 0.386 (0.342--0.431) & 0.407 (0.371--0.441) & \textbf{0.680 (0.663--0.697)} \\
PSV          & 0.048 (0.030--0.069) & 0.204 (0.168--0.240) & 0.068 (0.049--0.090) & 0.074 (0.051--0.098) & \textbf{0.363 (0.325--0.402)} \\
Pancreas     & 0.150 (0.118--0.183) & 0.256 (0.213--0.298) & 0.157 (0.124--0.190) & 0.177 (0.145--0.210) & \textbf{0.381 (0.337--0.424)} \\
R. adrenal   & 0.075 (0.042--0.112) & 0.237 (0.204--0.270) & 0.109 (0.076--0.144) & 0.101 (0.072--0.134) & \textbf{0.332 (0.303--0.362)} \\
L. adrenal   & 0.083 (0.058--0.110) & 0.213 (0.177--0.249) & 0.075 (0.049--0.103) & 0.099 (0.075--0.123) & \textbf{0.319 (0.280--0.359)} \\
\midrule
\textbf{Overall} & 0.257 (0.239--0.275) & 0.472 (0.449--0.493) & 0.278 (0.260--0.297) & 0.299 (0.281--0.318) & \textbf{0.565 (0.544--0.586)} \\
\bottomrule
\multicolumn{6}{l}{\textsuperscript{a}\scriptsize Gallbladder is absent in 17/45 pairs; n=28.} \\
\end{tabular}%
}
\end{table}
    
\begin{table}[h]
    \centering
    \caption{CT--CT intra-modal registration per-organ Hausdorff distance at the 95th percentile (HD95, mm) aggregated across all 45 validation pairs (5-fold CV). Mean (95\% bootstrap CI). Best per organ in bold (lower is better).}
    \label{tab:supp_reg_ctct_hd95}
    \footnotesize
    \resizebox{\textwidth}{!}{%
    \begin{tabular}{lccccc}
    \toprule
    Organ & VoxelMorph & DINOv2 & DINOv3 & Curia & FlexiCT-2D (Ours) \\
    \midrule
    Spleen       & 13.45 (11.96--15.01) & 11.28 (8.78--13.89) & 13.47 (11.73--15.23) & 12.31 (10.85--14.01) & \textbf{3.88 (2.77--5.09)} \\
    R. kidney    & 14.01 (11.97--16.27) & \textbf{6.26 (4.26--8.74)} & 14.57 (12.39--17.01) & 12.92 (10.88--15.01) & 6.88 (5.25--8.87) \\
    L. kidney    & 13.57 (11.81--15.48) & 6.79 (5.00--8.85) & 12.69 (11.06--14.37) & 11.94 (10.24--13.72) & \textbf{3.63 (2.67--4.61)} \\
    Gallbladder\textsuperscript{a} & 26.60 (22.93--30.46) & 16.15 (13.15--19.27) & 22.39 (19.08--25.52) & 23.77 (20.31--26.95) & \textbf{15.27 (12.33--18.31)} \\
    Esophagus    & 12.00 (10.10--13.82) & 9.09 (7.19--11.07) & 9.86 (8.10--11.72) & 11.23 (9.24--13.32) & \textbf{8.41 (6.77--10.15)} \\
    Liver        & 15.77 (14.32--17.32) & 6.39 (4.84--8.01) & 14.17 (12.40--16.00) & 13.19 (11.73--14.70) & \textbf{1.98 (1.54--2.49)} \\
    Stomach      & 26.06 (23.38--28.60) & 19.11 (16.78--21.61) & 23.69 (21.36--26.25) & 23.87 (21.35--26.45) & \textbf{14.46 (12.44--16.48)} \\
    Aorta        & 22.89 (19.25--26.46) & 17.09 (13.75--20.59) & 22.82 (19.42--26.40) & 21.95 (18.32--25.71) & \textbf{15.01 (11.43--18.62)} \\
    IVC          & 12.50 (9.86--15.25)  & 8.20 (6.55--10.09)  & 14.23 (11.05--17.64) & 11.62 (9.15--14.39)  & \textbf{5.16 (4.39--5.98)} \\
    PSV          & 19.60 (17.66--21.85) & 12.77 (11.32--14.19) & 17.49 (15.88--19.16) & 17.69 (15.83--19.71) & \textbf{10.08 (8.97--11.31)} \\
    Pancreas     & 20.84 (17.93--24.14) & 15.42 (13.34--17.63) & 19.18 (16.54--22.01) & 19.41 (16.47--22.51) & \textbf{12.76 (10.99--14.59)} \\
    R. adrenal   & 12.69 (11.06--14.37) & 7.43 (6.64--8.23)    & 10.30 (8.91--11.92)  & 10.91 (9.43--12.52)  & \textbf{6.42 (5.62--7.27)} \\
    L. adrenal   & 14.16 (12.33--16.16) & 9.50 (8.18--10.93)   & 12.74 (11.20--14.35) & 12.69 (10.87--14.64) & \textbf{6.22 (5.59--6.93)} \\
    \midrule
    \textbf{Overall} & 16.96 (16.19--17.78) & 11.04 (10.34--11.73) & 15.78 (15.06--16.52) & 15.41 (14.68--16.14) & \textbf{8.27 (7.65--8.84)} \\
    \bottomrule
    \multicolumn{6}{l}{\textsuperscript{a}\scriptsize Gallbladder is absent in 17/45 pairs; n=28.} \\
    \end{tabular}%
    }
    \end{table}
    
\begin{table}[h]
\centering
\caption{CT--MR cross-modal registration per-organ Dice similarity coefficient (DSC) aggregated across all validation pairs (5-fold CV, n=19 pairs). Mean (95\% bootstrap CI). Best per organ in bold.}
\label{tab:supp_reg_ctmr_dsc}
\footnotesize
\resizebox{\textwidth}{!}{%
\begin{tabular}{lccccc}
\toprule
Organ & VoxelMorph & DINOv2 & DINOv3 & Curia & FlexiCT-2D (Ours) \\
\midrule
Liver        & 0.543 (0.447--0.637) & 0.618 (0.477--0.746) & 0.571 (0.425--0.723) & 0.609 (0.498--0.727) & \textbf{0.797 (0.734--0.854)} \\
Spleen       & 0.336 (0.219--0.448) & 0.380 (0.226--0.541) & 0.411 (0.249--0.549) & 0.463 (0.280--0.637) & \textbf{0.641 (0.490--0.770)} \\
R. kidney    & 0.240 (0.125--0.350) & 0.352 (0.159--0.570) & 0.400 (0.273--0.526) & 0.404 (0.244--0.554) & \textbf{0.548 (0.375--0.704)} \\
L. kidney\textsuperscript{a} & 0.243 (0.161--0.320) & 0.418 (0.173--0.654) & 0.424 (0.343--0.533) & 0.413 (0.272--0.554) & \textbf{0.624 (0.491--0.747)} \\
\midrule
\textbf{Overall} & 0.346 (0.276--0.416) & 0.443 (0.344--0.546) & 0.453 (0.381--0.524) & 0.476 (0.388--0.555) & \textbf{0.654 (0.573--0.723)} \\
\bottomrule \\
\end{tabular}%
}
\end{table}
\begin{table}[h]
    \centering
    \caption{CT--MR cross-modal registration per-organ Hausdorff distance at the 95th percentile (HD95, mm) aggregated across all validation pairs (5-fold CV, n=19 pairs). Mean (95\% bootstrap CI). Best per organ in bold (lower is better).}
    \label{tab:supp_reg_ctmr_hd95}
    \footnotesize
    \resizebox{\textwidth}{!}{%
    \begin{tabular}{lccccc}
    \toprule
    Organ & VoxelMorph & DINOv2 & DINOv3 & Curia & FlexiCT-2D (Ours) \\
    \midrule
    Liver        & 13.20 (9.80--16.44)  & 15.65 (9.89--21.00)  & 16.88 (8.63--25.26)  & 11.71 (7.03--16.84)  & \textbf{6.19 (3.52--9.32)} \\
    Spleen       & 17.19 (10.22--24.89) & 17.76 (9.96--25.70)  & 16.36 (11.24--21.70) & 15.84 (8.20--23.78)  & \textbf{10.36 (4.90--16.61)} \\
    R. kidney    & 18.03 (11.82--24.87) & 19.74 (10.72--29.49) & 14.20 (8.96--19.32)  & 14.69 (7.98--22.04)  & \textbf{11.93 (6.90--16.17)} \\
    L. kidney\textsuperscript{a} & 18.38 (11.51--27.97) & 15.57 (7.88--23.00) & 14.85 (10.32--18.83) & 14.80 (8.02--22.46) & \textbf{11.50 (7.32--14.88)} \\
    \midrule
    \textbf{Overall} & 16.60 (13.61--20.10) & 17.27 (13.27--21.19) & 15.62 (12.64--18.80) & 14.23 (10.73--17.86) & \textbf{9.91 (7.61--12.34)} \\
    \bottomrule
    \end{tabular}%
    }
    \end{table}
            
\begin{table}[htbp]
    \centering
    \footnotesize
    \setlength{\tabcolsep}{3pt}
    \caption{KiTS classification results across label fractions. AUC and accuracy with 95\% BCa bootstrap confidence intervals. Best result per fraction in bold.}
    \label{tab:supp_cls_kits}
    \resizebox{\textwidth}{!}{%
    \begin{tabular}{@{}clcc@{}}
    \toprule
    Fraction & Model & AUC (95\% CI) & ACC (95\% CI) \\
    \midrule
    \multirow{4}{*}{0.01} & DINOv3 & 0.612 (0.491--0.731) & 0.466 (0.424--0.500) \\
     & Curia & 0.582 (0.463--0.701) & 0.457 (0.383--0.530) \\
     & BiomedCLIP & 0.495 (0.376--0.613) & 0.443 (0.391--0.488) \\
     & FlexiCT-2D (Ours) & \textbf{0.664 (0.548--0.777)} & \textbf{0.484 (0.429--0.537)} \\
    \cmidrule(lr){1-4}
    \multirow{4}{*}{0.05} & DINOv3 & 0.535 (0.416--0.652) & 0.522 (0.451--0.592) \\
     & Curia & 0.464 (0.344--0.583) & 0.472 (0.373--0.571) \\
     & BiomedCLIP & 0.477 (0.357--0.597) & 0.550 (0.462--0.638) \\
     & FlexiCT-2D (Ours) & \textbf{0.538 (0.403--0.669)} & \textbf{0.550 (0.446--0.650)} \\
    \cmidrule(lr){1-4}
    \multirow{4}{*}{0.10} & DINOv3 & 0.578 (0.460--0.693) & \textbf{0.565 (0.465--0.666)} \\
     & Curia & 0.583 (0.463--0.699) & 0.482 (0.421--0.541) \\
     & BiomedCLIP & 0.485 (0.367--0.602) & 0.486 (0.407--0.566) \\
     & FlexiCT-2D (Ours) & \textbf{0.683 (0.567--0.793)} & 0.453 (0.400--0.500) \\
    \cmidrule(lr){1-4}
    \multirow{4}{*}{0.25} & DINOv3 & 0.620 (0.505--0.733) & \textbf{0.558 (0.461--0.654)} \\
     & Curia & 0.604 (0.487--0.719) & 0.531 (0.460--0.602) \\
     & BiomedCLIP & 0.508 (0.386--0.627) & 0.500 (0.500--0.500) \\
     & FlexiCT-2D (Ours) & \textbf{0.747 (0.637--0.848)} & 0.502 (0.437--0.567) \\
    \cmidrule(lr){1-4}
    \multirow{4}{*}{0.50} & DINOv3 & 0.428 (0.312--0.543) & 0.415 (0.316--0.514) \\
     & Curia & 0.620 (0.505--0.732) & \textbf{0.558 (0.475--0.638)} \\
     & BiomedCLIP & 0.441 (0.326--0.559) & 0.506 (0.451--0.564) \\
     & FlexiCT-2D (Ours) & \textbf{0.809 (0.716--0.890)} & 0.489 (0.463--0.500) \\
    \cmidrule(lr){1-4}
    \multirow{4}{*}{1.00} & DINOv3 & 0.514 (0.393--0.637) & 0.500 (0.500--0.500) \\
     & Curia & 0.690 (0.574--0.800) & 0.481 (0.414--0.545) \\
     & BiomedCLIP & 0.546 (0.424--0.665) & 0.489 (0.462--0.500) \\
     & FlexiCT-2D (Ours) & \textbf{0.851 (0.760--0.934)} & \textbf{0.544 (0.482--0.607)} \\
    \bottomrule
    \end{tabular}%
    }
    \end{table}

\begin{table}[htbp]
    \centering
    \footnotesize
    \setlength{\tabcolsep}{3pt}
    \caption{DeepLesion classification results across label fractions. AUC and accuracy with 95\% BCa bootstrap confidence intervals. Best result per fraction in bold.}
    \label{tab:supp_cls_deeplesion}
    \resizebox{\textwidth}{!}{%
    \begin{tabular}{@{}clcc@{}}
    \toprule
    Fraction & Model & AUC (95\% CI) & ACC (95\% CI) \\
    \midrule
    \multirow{4}{*}{0.01} & DINOv3 & 0.613 (0.591--0.634) & 0.125 (0.125--0.125) \\
     & Curia & 0.832 (0.813--0.850) & 0.282 (0.260--0.303) \\
     & BiomedCLIP & 0.819 (0.799--0.838) & 0.259 (0.245--0.274) \\
     & FlexiCT-2D (Ours) & \textbf{0.922 (0.909--0.935)} & \textbf{0.335 (0.315--0.355)} \\
    \cmidrule(lr){1-4}
    \multirow{4}{*}{0.05} & DINOv3 & 0.777 (0.756--0.798) & 0.125 (0.125--0.125) \\
     & Curia & 0.924 (0.911--0.936) & 0.479 (0.444--0.514) \\
     & BiomedCLIP & 0.963 (0.955--0.970) & \textbf{0.525 (0.494--0.557)} \\
     & FlexiCT-2D (Ours) & \textbf{0.979 (0.971--0.986)} & 0.487 (0.465--0.511) \\
    \cmidrule(lr){1-4}
    \multirow{4}{*}{0.10} & DINOv3 & 0.848 (0.831--0.864) & 0.125 (0.125--0.125) \\
     & Curia & 0.968 (0.960--0.975) & 0.615 (0.575--0.654) \\
     & BiomedCLIP & 0.977 (0.973--0.982) & 0.590 (0.558--0.622) \\
     & FlexiCT-2D (Ours) & \textbf{0.988 (0.982--0.992)} & \textbf{0.635 (0.602--0.668)} \\
    \cmidrule(lr){1-4}
    \multirow{4}{*}{0.25} & DINOv3 & 0.866 (0.850--0.881) & 0.143 (0.137--0.150) \\
     & Curia & 0.988 (0.984--0.991) & 0.834 (0.800--0.866) \\
     & BiomedCLIP & 0.986 (0.982--0.989) & 0.736 (0.698--0.775) \\
     & FlexiCT-2D (Ours) & \textbf{0.994 (0.991--0.996)} & \textbf{0.859 (0.828--0.890)} \\
    \cmidrule(lr){1-4}
    \multirow{4}{*}{0.50} & DINOv3 & 0.923 (0.910--0.935) & 0.246 (0.231--0.261) \\
     & Curia & 0.991 (0.988--0.993) & 0.840 (0.807--0.871) \\
     & BiomedCLIP & 0.989 (0.986--0.991) & 0.778 (0.740--0.815) \\
     & FlexiCT-2D (Ours) & \textbf{0.996 (0.994--0.997)} & \textbf{0.876 (0.844--0.905)} \\
    \cmidrule(lr){1-4}
    \multirow{4}{*}{1.00} & DINOv3 & 0.957 (0.946--0.967) & 0.390 (0.368--0.412) \\
     & Curia & 0.994 (0.992--0.996) & 0.869 (0.839--0.898) \\
     & BiomedCLIP & 0.991 (0.989--0.994) & 0.817 (0.782--0.852) \\
     & FlexiCT-2D (Ours) & \textbf{0.997 (0.995--0.998)} & \textbf{0.879 (0.848--0.908)} \\
    \bottomrule
    \end{tabular}%
    }
    \end{table}

\begin{table}[htbp]
    \centering
    \footnotesize
    \setlength{\tabcolsep}{3pt}
    \caption{LUNA16 classification results across label fractions. AUC and accuracy with 95\% BCa bootstrap confidence intervals. Best result per fraction in bold.}
    \label{tab:supp_cls_luna16}
    \resizebox{\textwidth}{!}{%
    \begin{tabular}{@{}clcc@{}}
    \toprule
    Fraction & Model & AUC (95\% CI) & ACC (95\% CI) \\
    \midrule
    \multirow{4}{*}{0.01} & DINOv3 & 0.573 (0.480--0.663) & 0.500 (0.500--0.500) \\
    & Curia & 0.542 (0.449--0.635) & 0.500 (0.500--0.500) \\
    & BiomedCLIP & 0.500 (0.404--0.596) & 0.500 (0.500--0.500) \\
    & FlexiCT-2D (Ours) & \textbf{0.635 (0.544--0.726)} & 0.500 (0.500--0.500) \\
    \cmidrule(lr){1-4}
    \multirow{4}{*}{0.05} & DINOv3 & 0.545 (0.449--0.635) & 0.500 (0.500--0.500) \\
     & Curia & 0.669 (0.581--0.752) & 0.607 (0.528--0.684) \\
     & BiomedCLIP & 0.531 (0.437--0.624) & 0.500 (0.500--0.500) \\
     & FlexiCT-2D (Ours) & \textbf{0.907 (0.852--0.952)} & \textbf{0.815 (0.753--0.873)} \\
    \cmidrule(lr){1-4}
    \multirow{4}{*}{0.10} & DINOv3 & 0.559 (0.462--0.651) & 0.513 (0.432--0.593) \\
     & Curia & 0.700 (0.612--0.781) & 0.500 (0.500--0.500) \\
     & BiomedCLIP & 0.834 (0.766--0.896) & 0.783 (0.713--0.848) \\
     & FlexiCT-2D (Ours) & \textbf{0.916 (0.869--0.956)} & \textbf{0.822 (0.767--0.877)} \\
    \cmidrule(lr){1-4}
    \multirow{4}{*}{0.25} & DINOv3 & 0.610 (0.517--0.700) & 0.508 (0.500--0.524) \\
     & Curia & 0.889 (0.831--0.938) & 0.575 (0.534--0.619) \\
     & BiomedCLIP & 0.815 (0.744--0.879) & 0.659 (0.595--0.724) \\
     & FlexiCT-2D (Ours) & \textbf{0.953 (0.920--0.979)} & \textbf{0.865 (0.809--0.915)} \\
    \cmidrule(lr){1-4}
    \multirow{4}{*}{0.50} & DINOv3 & 0.674 (0.586--0.760) & 0.599 (0.524--0.671) \\
     & Curia & 0.920 (0.873--0.959) & 0.816 (0.752--0.878) \\
     & BiomedCLIP & 0.844 (0.777--0.904) & 0.774 (0.705--0.841) \\
     & FlexiCT-2D (Ours) & \textbf{0.954 (0.920--0.981)} & \textbf{0.895 (0.842--0.941)} \\
    \cmidrule(lr){1-4}
    \multirow{4}{*}{1.00} & DINOv3 & 0.755 (0.675--0.831) & 0.663 (0.588--0.739) \\
     & Curia & 0.942 (0.905--0.973) & \textbf{0.879 (0.825--0.928)} \\
     & BiomedCLIP & 0.884 (0.828--0.933) & 0.782 (0.715--0.845) \\
     & FlexiCT-2D (Ours) & \textbf{0.961 (0.932--0.983)} & 0.864 (0.805--0.917) \\
    \bottomrule
    \end{tabular}%
    }
    \end{table}

\begin{table}[htbp]
    \centering
    \footnotesize
    \setlength{\tabcolsep}{3pt}
    \caption{COVIDx-CT classification results across label fractions. AUC and accuracy with 95\% BCa bootstrap confidence intervals. Best result per fraction in bold.}
    \label{tab:supp_cls_covidx}
    \resizebox{\textwidth}{!}{%
    \begin{tabular}{@{}clcc@{}}
    \toprule
    Fraction & Model & AUC (95\% CI) & ACC (95\% CI) \\
    \midrule
    \multirow{4}{*}{0.01} & DINOv3 & 0.539 (0.523--0.554) & 0.349 (0.336--0.362) \\
     & Curia & 0.650 (0.637--0.663) & 0.379 (0.373--0.385) \\
     & BiomedCLIP & 0.680 (0.669--0.692) & 0.334 (0.333--0.334) \\
     & FlexiCT-2D (Ours) & \textbf{0.889 (0.880--0.897)} & \textbf{0.697 (0.682--0.711)} \\
    \cmidrule(lr){1-4}
    \multirow{4}{*}{0.05} & DINOv3 & 0.525 (0.511--0.540) & 0.333 (0.333--0.333) \\
     & Curia & 0.901 (0.893--0.910) & 0.678 (0.663--0.693) \\
     & BiomedCLIP & 0.887 (0.877--0.896) & 0.432 (0.424--0.439) \\
     & FlexiCT-2D (Ours) & \textbf{0.964 (0.959--0.969)} & \textbf{0.835 (0.822--0.847)} \\
    \cmidrule(lr){1-4}
    \multirow{4}{*}{0.10} & DINOv3 & 0.543 (0.529--0.557) & 0.333 (0.333--0.333) \\
     & Curia & 0.943 (0.936--0.950) & 0.797 (0.783--0.809) \\
     & BiomedCLIP & 0.924 (0.916--0.931) & 0.567 (0.557--0.577) \\
     & FlexiCT-2D (Ours) & \textbf{0.974 (0.969--0.978)} & \textbf{0.868 (0.856--0.879)} \\
    \cmidrule(lr){1-4}
    \multirow{4}{*}{0.25} & DINOv3 & 0.626 (0.613--0.640) & 0.337 (0.335--0.340) \\
     & Curia & 0.957 (0.950--0.963) & 0.835 (0.823--0.847) \\
     & BiomedCLIP & 0.938 (0.931--0.945) & 0.688 (0.674--0.702) \\
     & FlexiCT-2D (Ours) & \textbf{0.978 (0.974--0.982)} & \textbf{0.877 (0.866--0.888)} \\
    \cmidrule(lr){1-4}
    \multirow{4}{*}{0.50} & DINOv3 & 0.719 (0.707--0.731) & 0.373 (0.367--0.378) \\
     & Curia & 0.970 (0.966--0.975) & 0.846 (0.834--0.857) \\
     & BiomedCLIP & 0.952 (0.946--0.958) & 0.758 (0.744--0.772) \\
     & FlexiCT-2D (Ours) & \textbf{0.981 (0.977--0.984)} & \textbf{0.881 (0.870--0.892)} \\
    \cmidrule(lr){1-4}
    \multirow{4}{*}{1.00} & DINOv3 & 0.803 (0.792--0.814) & 0.464 (0.455--0.474) \\
     & Curia & 0.977 (0.973--0.981) & 0.865 (0.854--0.876) \\
     & BiomedCLIP & 0.956 (0.950--0.961) & 0.792 (0.777--0.806) \\
     & FlexiCT-2D (Ours) & \textbf{0.983 (0.980--0.987)} & \textbf{0.890 (0.879--0.900)} \\
    \bottomrule
    \end{tabular}%
    }
    \end{table}

\begin{table}[h]
\centering
\caption{Linear probing results for tumor phenotype prediction. Balanced accuracy, F1 score, area under the ROC curve (AUC), and average precision (PR) with 95\% BCa bootstrap confidence intervals. Best result per task and metric in bold.}
\label{tab:supp_linear_probing}
\footnotesize
\resizebox{\textwidth}{!}{%
\begin{tabular}{llcccc}
\toprule
Task & Model & Balanced ACC (95\% CI) & F1 (95\% CI) & AUC (95\% CI) & PR (95\% CI) \\
\midrule
\multirow{5}{*}{\shortstack[l]{T stage\\(NSCLC Radiogenomics)}}
    & Baseline (tumor diameter + GG) & 0.525 (0.483--0.548) & 0.523 (0.480--0.547) & 0.662 (0.639--0.678) & 0.526 (0.480--0.549) \\
    & CT-FM & 0.487 (0.453--0.523) & 0.481 (0.393--0.525) & 0.651 (0.618--0.672) & 0.470 (0.435--0.495) \\
    & Spectre & 0.485 (0.463--0.517) & 0.479 (0.446--0.515) & 0.620 (0.573--0.662) & 0.461 (0.417--0.509) \\
    & Voco & 0.376 (0.354--0.402) & 0.336 (0.300--0.374) & 0.508 (0.485--0.539) & 0.364 (0.346--0.386) \\
    & FlexiCT-3D (Ours) & \textbf{0.561 (0.521--0.588)} & \textbf{0.564 (0.522--0.594)} & \textbf{0.681 (0.651--0.724)} & \textbf{0.528 (0.498--0.583)} \\
\midrule
\multirow{5}{*}{\shortstack[l]{ISUP grade\\(C4KC-KiTS)}}
    & Baseline (tumor diameter) & 0.727 (0.723--0.734) & \textbf{0.728 (0.724--0.737)} & 0.728 (0.705--0.741) & 0.678 (0.652--0.692) \\
    & CT-FM & 0.669 (0.637--0.713) & 0.671 (0.635--0.713) & 0.689 (0.651--0.729) & 0.680 (0.641--0.715) \\
    & Spectre & 0.685 (0.661--0.705) & 0.672 (0.650--0.691) & 0.705 (0.680--0.723) & 0.688 (0.667--0.703) \\
    & Voco & 0.708 (0.680--0.729) & 0.699 (0.668--0.722) & 0.741 (0.715--0.768) & 0.711 (0.671--0.749) \\
    & FlexiCT-3D (Ours) & \textbf{0.730 (0.707--0.759)} & 0.728 (0.704--0.758) & \textbf{0.765 (0.749--0.782)} & \textbf{0.743 (0.713--0.767)} \\
\bottomrule
\end{tabular}%
}
\end{table}

\begin{table}[h]
\centering
\caption{tumor phenotype retrieval results. Recall@1, Recall@3, and mean average precision (mAP) with 95\% BCa bootstrap confidence intervals. Best result per task and metric in bold.}
\label{tab:supp_retrieval}
\footnotesize
\resizebox{\textwidth}{!}{%
\begin{tabular}{llccc}
\toprule
Task & Model & Recall@1 (95\% CI) & Recall@3 (95\% CI) & mAP (95\% CI) \\
\midrule
\multirow{4}{*}{\shortstack[l]{T stage\\(NSCLC Radiogenomics)}}
    & CT-FM & 0.507 (0.394--0.620) & 0.789 (0.690--0.887) & 0.504 (0.463--0.543) \\
    & Spectre & 0.451 (0.338--0.578) & 0.732 (0.634--0.831) & 0.456 (0.422--0.493) \\
    & Voco & 0.380 (0.313--0.447) & 0.760 (0.702--0.817) & 0.408 (0.395--0.422) \\
    & FlexiCT-3D (Ours) & \textbf{0.662 (0.549--0.761)} & \textbf{0.845 (0.761--0.930)} & \textbf{0.524 (0.488--0.557)} \\
\midrule
\multirow{4}{*}{\shortstack[l]{ISUP grade\\(C4KC-KiTS)}}
    & CT-FM & 0.643 (0.529--0.757) & 0.786 (0.686--0.886) & \textbf{0.650 (0.591--0.700)} \\
    & Spectre & 0.700 (0.586--0.800) & 0.914 (0.843--0.971) & 0.647 (0.605--0.686) \\
    & Voco & 0.586 (0.471--0.700) & 0.814 (0.636--0.896) & 0.617 (0.390--0.723) \\
    & FlexiCT-3D (Ours) & \textbf{0.743 (0.643--0.843)} & \textbf{0.971 (0.929--1.000)} & \textbf{0.658 (0.612--0.702)}  \\
\bottomrule
\end{tabular}%
}
\end{table}

\begin{table}[htbp]
        \centering
        \caption{VLM zero-shot disease classification results. Precision, F1 score, accuracy (ACC), and area under the ROC curve (AUC) with 95\% BCa bootstrap confidence intervals. Best result per dataset and metric in bold.}
        \label{tab:supp_vlm}\label{tab:supp_vlm_zeroshot}
        \footnotesize
        \resizebox{\textwidth}{!}{%
        \begin{tabular}{llcccc}
        \toprule
        Dataset & Model & Precision (95\% CI) & F1 (95\% CI) & ACC (95\% CI) & AUC (95\% CI) \\
        \midrule
        \multirow{4}{*}{CT-RATE} 
        & CT-CLIP & 0.329 (0.320--0.337) & 0.427 (0.417--0.436) & 0.682 (0.676--0.687) & 0.732 (0.724--0.739) \\
        & COPLPRI & 0.385 (0.376--0.394) & 0.482 (0.472--0.492) & 0.724 (0.719--0.729) & 0.787 (0.780--0.794) \\
        & SPECTRE & 0.225 (0.218--0.232) & 0.307 (0.299--0.315) & 0.553 (0.548--0.559) & 0.567 (0.558--0.577) \\
        & FlexiCT-3D-VLM & \textbf{0.403 (0.393--0.412)} & \textbf{0.509 (0.499--0.519)} & \textbf{0.748 (0.743--0.753)} & \textbf{0.813 (0.807--0.820)} \\
        \midrule
        \multirow{4}{*}{Merlin} 
        & Merlin & 0.739 (0.721--0.756) & \textbf{0.735 (0.721--0.749)} & 0.732 (0.719--0.745) & 0.825 (0.812--0.838) \\
        & COPLPRI & 0.572 (0.555--0.588) & 0.651 (0.637--0.665) & 0.585 (0.571--0.599) & 0.737 (0.722--0.752) \\
        & SPECTRE & 0.546 (0.465--0.628) & 0.526 (0.445--0.605) & 0.561 (0.545--0.586) & 0.601 (0.525--0.677) \\
        & FlexiCT-3D-VLM & \textbf{0.869 (0.848--0.889)} & 0.725 (0.709--0.740) & \textbf{0.776 (0.765--0.788)} & \textbf{0.872 (0.862--0.882)} \\
        \bottomrule
        \end{tabular}%
        }
        \end{table}

\begin{table}[htbp]
\centering
\caption{VLM report retrieval results. Recall at rank $K$ with 95\% BCa bootstrap confidence intervals. CT-RATE retrieval uses Recall@5 and Recall@10; Merlin retrieval uses Recall@1 and Recall@8 at gallery size $N{=}32$. Best result per dataset and metric in bold.}
\label{tab:supp_vlm_retrieval}
\footnotesize
\resizebox{\textwidth}{!}{%
\begin{tabular}{llcc}
\toprule
Dataset & Model & Recall@$K_1$ (95\% CI) & Recall@$K_2$ (95\% CI) \\
\midrule
\multicolumn{4}{l}{\textit{CT-RATE report retrieval (Recall@5, Recall@10)}} \\
\midrule
\multirow{4}{*}{CT-RATE} 
& CT-CLIP & 0.039 (0.029--0.049) & 0.068 (0.056--0.081) \\
& COPLPRI & 0.199 (0.179--0.219) & 0.290 (0.267--0.312) \\
& SPECTRE & 0.152 (0.135--0.171) & 0.221 (0.201--0.241) \\
& FlexiCT-3D-VLM & \textbf{0.378 (0.354--0.403)} & \textbf{0.462 (0.438--0.487)} \\
\midrule
\multicolumn{4}{l}{\textit{Merlin report retrieval (Recall@1, Recall@8; $N{=}32$)}} \\
\midrule
\multirow{4}{*}{Merlin} 
& Merlin & 0.719 (0.706--0.731) & 0.974 (0.970--0.979) \\
& COPLPRI & 0.191 (0.180--0.201) & 0.673 (0.660--0.686) \\
& SPECTRE & 0.655 (0.642--0.668) & 0.959 (0.954--0.965) \\
& FlexiCT-3D-VLM & \textbf{0.888 (0.888--1.000)} & \textbf{0.996 (0.996--1.000)} \\
\bottomrule
\end{tabular}%
}
\end{table}


\begin{landscape}
{%
\setlength{\textwidth}{\dimexpr\paperheight-2in\relax}
\setlength{\linewidth}{\textwidth}
\setlength{\hsize}{\textwidth}
\scriptsize
\setlength{\LTcapwidth}{\textwidth}
\setlength{\tabcolsep}{4pt}
\begin{longtable}{@{}>{\raggedright\arraybackslash}p{4.1cm} >{\raggedright\arraybackslash}p{3.8cm} p{1.5cm} r p{1.7cm} >{\raggedright\arraybackslash}p{4.5cm} >{\raggedright\arraybackslash}p{4.5cm}@{}}
\caption{Pretraining dataset access information. Primary source URL, secondary reference, anatomical region, and volume count for all 56 pretraining datasets used in FlexiCT.}
\label{tab:supp_dataset_urls} \\
\toprule
Dataset & Full Name & Region & Volumes & Country & Primary URL & Secondary URL \\
\midrule
\endfirsthead

\multicolumn{7}{l}{\textit{Supplementary Table 20 continued}} \\
\toprule
Dataset & Full Name & Region & Volumes & Country & Primary URL & Secondary URL \\
\midrule
\endhead

\midrule
\multicolumn{7}{r}{\textit{Continued on next page}} \\
\endfoot

\bottomrule
\endlastfoot


0013\_ribfrac & RibFrac & & 660 & China & \url{https://ribfrac.grand-challenge.org/} & \url{https://pmc.ncbi.nlm.nih.gov/articles/PMC7670192/} \\

0019\_tcia\_ct\_lymph\_nodes & CT Lymph Nodes & & 174 & USA & \url{https://wiki.cancerimagingarchive.net/pages/viewpage.action?pageId=19726546} & \url{https://www.cancerimagingarchive.net/collection/ct-lymph-nodes/} \\

0020\_tcia\_cptac\_ccrcc & CPTAC Clear Cell Renal Cell Carcinoma & & 258 & USA & \url{https://wiki.cancerimagingarchive.net/pages/viewpage.action?pageId=33948213} & \url{https://www.cancerimagingarchive.net/collection/cptac-ccrcc/} \\

0021\_tcia\_cptac\_luad & CPTAC Lung Adenocarcinoma & & 133 & USA & \url{https://wiki.cancerimagingarchive.net/pages/viewpage.action?pageId=33948253} & \url{https://www.cancerimagingarchive.net/collection/cptac-luad/} \\

0023\_tcia\_nsclc\_radiomics & NSCLC Radiomics & & 131 & Netherlands & \url{https://www.cancerimagingarchive.net/collection/nsclc-radiomics/} & \url{https://doi.org/10.1038/ncomms5006} \\

0025\_pancreatic\_ct\_cbct\_seg & Pancreatic CT-CBCT Segmentation & & 93 & USA & \url{https://wiki.cancerimagingarchive.net/pages/viewpage.action?pageId=93258557} & \url{https://doi.org/10.1038/s41597-022-01758-9} \\

0029\_tcia\_tcga\_kirp & TCGA Kidney Renal Papillary Cell Carcinoma & & 19 & USA & \url{https://wiki.cancerimagingarchive.net/pages/viewpage.action?pageId=11829555} & \url{https://www.cancerimagingarchive.net/collection/tcga-kirp/} \\

0030\_tcia\_tcga\_lihc & TCGA Liver Hepatocellular Carcinoma & & 242 & USA & \url{https://wiki.cancerimagingarchive.net/pages/viewpage.action?pageId=6885436} & \url{https://www.cancerimagingarchive.net/collection/tcga-lihc/} \\

0042\_new\_brainct\_1mm & CADS Brain CT 1\,mm & & 484 & Turkey & \url{https://huggingface.co/datasets/mrmrx/CADS-dataset/blob/main/0042_new_brainct_1mm/README_0042_new_brainct_1mm.md} & \url{https://arxiv.org/abs/2507.22953} \\

AbdomenCT-1K & AbdomenCT-1K & & 1{,}000 & China & \url{https://arxiv.org/abs/2010.14808} & \url{https://github.com/JunMa11/AbdomenCT-1K} \\

AbdominalTraumaDetection & RSNA Abdominal Trauma Detection & & 2{,}029 & USA & \url{https://pubs.rsna.org/doi/10.1148/ryai.240334} & \url{https://arxiv.org/abs/2405.19595} \\

acrin\_flt\_breast & ACRIN [18F]-FLT Breast PET/CT & & 279 & USA & \url{https://wiki.cancerimagingarchive.net/pages/viewpage.action?pageId=30671268} & \url{https://pmc.ncbi.nlm.nih.gov/articles/PMC4737647/} \\

AMOS & Abdominal Multi-Organ Segmentation & Abdominal & 1{,}850 & China & \url{http://www.amos.sribd.cn/about.html} & \url{https://arxiv.org/abs/2206.08023} \\

anti\_pd\_1\_lung & Anti-PD-1 Lung Cancer & & 265 & USA & \url{https://www.cancerimagingarchive.net/collection/anti-pd-1_lung/} & \url{https://wiki.cancerimagingarchive.net/pages/viewpage.action?pageId=41517500} \\

BTCV & Multi-Atlas Labeling Beyond the Cranial Vault & & 47 & USA & \url{https://www.synapse.org/Synapse:syn3193805} & \url{https://github.com/openmedlab/Awesome-Medical-Dataset/blob/main/resources/BTCV.md} \\

CADS\_0043\_new\_ct\_tri & CADS CT Tri-Region Dataset & & 585 & Germany & \url{https://huggingface.co/datasets/mrmrx/CADS-dataset/blob/main/0043_new_ct_tri/README_0043_new_ct_tri.md} & \url{https://arxiv.org/abs/2507.22953} \\

CHAOS & CHAOS Healthy Abdominal Organ Segmentation & & 20 & Turkey & \url{https://chaos.grand-challenge.org/} & \url{https://pubmed.ncbi.nlm.nih.gov/33421920/} \\

cmb\_crc & Cancer Moonshot Biobank Colorectal Cancer & & 251 & USA & \url{https://www.cancerimagingarchive.net/collection/cmb-crc/} & \url{https://www.ncbi.nlm.nih.gov/projects/gap/cgi-bin/study.cgi?study_id=phs002192.v1.p1} \\

Colorectal-Liver-Metastases & Colorectal Liver Metastases & & 197 & USA & \url{https://www.nature.com/articles/s41597-024-02981-2} & \url{https://wiki.cancerimagingarchive.net/pages/viewpage.action?pageId=89096268} \\

cptac\_lscc & CPTAC Lung Squamous Cell Carcinoma & & 159 & USA & \url{https://wiki.cancerimagingarchive.net/pages/viewpage.action?pageId=33948248} & \url{https://www.cancerimagingarchive.net/collection/cptac-lscc/} \\

cptac\_pda & CPTAC Pancreatic Ductal Adenocarcinoma & & 305 & USA & \url{https://wiki.cancerimagingarchive.net/pages/viewpage.action?pageId=33948258} & \url{https://www.cancerimagingarchive.net/collection/cptac-pda/} \\

cptac\_ucec & CPTAC Uterine Corpus Endometrial Carcinoma & & 393 & USA & \url{https://wiki.cancerimagingarchive.net/pages/viewpage.action?pageId=33948263} & \url{https://www.cancerimagingarchive.net/collection/cptac-ucec/} \\

CT Colonography & ACRIN 6664 CT Colonography & Chest, Abd., Pelvic & 1{,}730 & USA & \url{https://wiki.cancerimagingarchive.net/pages/viewpage.action?pageId=3539213} & \url{https://pmc.ncbi.nlm.nih.gov/articles/PMC3144954/} \\

CT-ORG & CT-ORG Multi-Organ Segmentation & & 140 & USA & \url{https://www.nature.com/articles/s41597-020-00715-8} & \url{https://pmc.ncbi.nlm.nih.gov/articles/PMC7658204/} \\

CT-RATE & CT-RATE & Chest & 47{,}149 & Turkey & \url{https://huggingface.co/datasets/ibrahimhamamci/CT-RATE} & \url{https://arxiv.org/abs/2403.17834} \\

DeepLesion & DeepLesion & & 5{,}000 & USA & \url{https://doi.org/10.1117/1.JMI.5.3.036501} & \url{https://nihcc.app.box.com/v/DeepLesion} \\

FLARE'23 & FLARE 23 Challenge & & 4{,}100 & Canada & \url{https://arxiv.org/abs/2408.12534} & \url{https://arxiv.org/html/2408.12534} \\

HCC-TACE-Seg & HCC Transarterial Chemoembolization Seg. & & 103 & USA & \url{https://wiki.cancerimagingarchive.net/pages/viewpage.action?pageId=70230229} & \url{https://www.nature.com/articles/s41597-023-01928-3} \\

HECTOR & HECKTOR Head and Neck Tumor & & 680 & Switzerland & \url{https://hecktor.grand-challenge.org/} & \url{https://pmc.ncbi.nlm.nih.gov/articles/PMC10171217/} \\

HNSCC & Head and Neck Squamous Cell Carcinoma & Head \& neck & 1{,}071 & USA & \url{https://www.cancerimagingarchive.net/collection/hnscc/} & \url{https://www.nature.com/articles/sdata2018173} \\

INSPECT & INSPECT Chest CT-Report Dataset & Chest & 23{,}240 & USA & \url{https://som-shahlab.github.io/inspect-website/} & \url{https://arxiv.org/abs/2311.10798} \\

LUNA16 & LUng Nodule Analysis 2016 & Chest & 843 & USA & \url{https://luna16.grand-challenge.org/Data/} & \url{https://pmc.ncbi.nlm.nih.gov/articles/PMC3041807/} \\

Lung-PET-CT-Dx & Lung PET-CT Diagnosis & & 347 & China & \url{https://wiki.cancerimagingarchive.net/pages/viewpage.action?pageId=70224216} & \url{https://www.cancerimagingarchive.net/collection/lung-pet-ct-dx/} \\

Merlin & Merlin Vision-Language Foundation Model & Abdominal & 25{,}489 & USA & \url{https://arxiv.org/html/2406.06512} & \url{https://www.nature.com/articles/s41586-026-10181-8} \\

midrc\_ricord\_1a & MIDRC RICORD 1a COVID-19 CT & & 163 & USA & \url{https://doi.org/10.1148/radiol.2021203957} & \url{https://pmc.ncbi.nlm.nih.gov/articles/PMC7993245/} \\

NLST & National Lung Screening Trial & Chest & 132{,}985 & USA & \url{https://www.cancerimagingarchive.net/collection/nlst/} & \url{https://pmc.ncbi.nlm.nih.gov/articles/PMC3009383/} \\

OPC-Radiomics & Oropharyngeal Cancer Radiomics & Head \& neck & 606 & Canada & \url{https://wiki.cancerimagingarchive.net/pages/viewpage.action?pageId=33948764} & \url{https://www.cancerimagingarchive.net/collection/opc-radiomics/} \\

Panorama & PANORAMA Abdominal Organ Segmentation & Abdominal & 2{,}238 & Netherlands & \url{https://panorama.grand-challenge.org/datasets-imaging-labels/} & \url{https://zenodo.org/records/11034178} \\

Pediatric-CT-SEG & Pediatric CT Segmentation Dataset & & 358 & USA & \url{https://pubmed.ncbi.nlm.nih.gov/35067940/} & \url{https://www.cancerimagingarchive.net/collection/pediatric-ct-seg/} \\

Prostate-Anatomical-Edge-Cases & Prostate Anatomical Edge Cases & & 131 & USA & \url{https://www.cancerimagingarchive.net/collection/prostate-anatomical-edge-cases/} & \url{https://pubmed.ncbi.nlm.nih.gov/36793398/} \\

Qin-Headneck & QIN Head-Neck Collection & Head \& neck & 898 & USA & \url{https://www.cancerimagingarchive.net/collection/qin-headneck/} & \url{https://peerj.com/articles/2057/} \\

rider\_lung\_pet\_ct & RIDER Lung PET/CT & & 235 & USA & \url{https://doi.org/10.7937/k9/tcia.2015.ofip7tvm} & \url{https://wiki.cancerimagingarchive.net/display/Public/RIDER+Lung+PET-CT} \\

StageII-Colorectal-CT & Stage II Colorectal CT & & 230 & China & \url{https://www.cancerimagingarchive.net/collection/stageii-colorectal-ct/} & \url{https://onlinelibrary.wiley.com/doi/10.1002/ijc.34053} \\

STOIC & STOIC 2021 & Chest & 2{,}000 & France & \url{https://pubs.rsna.org/doi/full/10.1148/radiol.2021210384} & \url{https://stoic2021.grand-challenge.org/} \\

StonyBrookChestCT & COVID-19-NY-SBU Chest CT & Chest & 2{,}316 & USA & \url{https://www.cancerimagingarchive.net/collection/covid-19-ny-sbu/} & \url{https://wiki.cancerimagingarchive.net/pages/viewpage.action?pageId=89096912} \\

tcga\_blca & TCGA Bladder Urothelial Carcinoma & & 409 & USA & \url{https://wiki.cancerimagingarchive.net/pages/viewpage.action?pageId=16056367} & \url{https://www.cancerimagingarchive.net/collection/tcga-blca/} \\

tcga\_kirc & TCGA Kidney Renal Clear Cell Carcinoma & & 812 & USA & \url{https://wiki.cancerimagingarchive.net/pages/viewpage.action?pageId=5800386} & \url{https://www.cancerimagingarchive.net/collection/tcga-kirc/} \\

tcga\_luad & TCGA Lung Adenocarcinoma & & 183 & USA & \url{https://wiki.cancerimagingarchive.net/pages/viewpage.action?pageId=6881474} & \url{https://gdc.cancer.gov/resources-tcga-users/tcga-code-tables/tissue-source-site-codes} \\

tcga\_lusc & TCGA Lung Squamous Cell Carcinoma & & 133 & USA & \url{https://wiki.cancerimagingarchive.net/pages/viewpage.action?pageId=16056484} & \url{https://gdc.cancer.gov/resources-tcga-users/tcga-code-tables/tissue-source-site-codes} \\

tcga\_ov & TCGA Ovarian Cancer & & 384 & USA & \url{https://wiki.cancerimagingarchive.net/pages/viewpage.action?pageId=7569497} & \url{https://www.cancerimagingarchive.net/collection/tcga-ov/} \\

tcga\_stad & TCGA Stomach Adenocarcinoma & & 237 & USA & \url{https://wiki.cancerimagingarchive.net/pages/viewpage.action?pageId=19039400} & \url{https://gdc.cancer.gov/resources-tcga-users/tcga-code-tables/tissue-source-site-codes} \\

tcga\_ucec & TCGA Uterine Corpus Endometrial Carcinoma & & 330 & USA & \url{https://wiki.cancerimagingarchive.net/pages/viewpage.action?pageId=19039602} & \url{https://www.cancerimagingarchive.net/collection/tcga-ucec/} \\

TCIA-Pancreas-CT & TCIA Pancreas CT & & 42 & USA & \url{https://wiki.cancerimagingarchive.net/display/public/pancreas-ct} & \url{https://www.cancerimagingarchive.net/collection/pancreas-ct/} \\

Totalsegmentator V2 & TotalSegmentator V2 & All body & 1{,}203 & Switzerland & \url{https://zenodo.org/records/10047292} & \url{https://pmc.ncbi.nlm.nih.gov/articles/PMC10546353/} \\

ULS\_Radbound\_Bone\_lesion & ULS Radboud Bone Lesion Subset & Abdominal & 744 & Netherlands & \url{https://zenodo.org/records/10035161} & \url{https://arxiv.org/abs/2406.05231} \\

ULS\_Radbound\_Pancreas & ULS Radboud Pancreas Subset & Abdominal & 124 & Netherlands & \url{https://zenodo.org/records/10035161} & \url{https://arxiv.org/abs/2406.05231} \\

\end{longtable}
}%
\end{landscape}

\begin{table}[htbp]
\centering
\caption{Effect of 2D pretraining initialization on 3D volumetric tasks. FlexiCT-3D-NoInit performs Phase~2 self-supervised pretraining from scratch; FlexiCT-3D starts from the Phase~1 2D checkpoint under an otherwise identical recipe and compute budget. $P$ values are from paired permutation tests, two-sided, with 10{,}000 permutations. \textsuperscript{*} denotes significance after Bonferroni correction at $\alpha = 0.05/3 = 0.0167$. Best result per row is shown in bold.}
\label{tab:supp_ablation_phase1}
\footnotesize
\resizebox{\textwidth}{!}{%
\begin{tabular}{llcc}
\toprule
Task family & Dataset / metric & FlexiCT-3D-NoInit & FlexiCT-3D \\
\midrule
Segmentation (Dice)
 & WORD
 & 0.829 (0.808--0.847)
 & \textbf{0.854 (0.834--0.872)} \\
Retrieval (Recall@1)
 & C4KC-KiTS, ISUP
 & 0.613 (0.498--0.710)
 & \textbf{0.743 (0.643--0.843)} \\
Linear probe (AUC)
 & C4KC-KiTS, ISUP
 & 0.703 (0.683--0.726)
 & \textbf{0.765 (0.749--0.782)} \\
\bottomrule
\end{tabular}%
}
\end{table}

\begin{table}[h]
    \centering
    \caption{Effect of initialization on Phase~3 vision--language alignment. All variants use the same Phase~3 contrastive recipe and compute budgets. Random init starts from a randomly initialised 3D ViT; FlexiCT-2D starts from the Phase~1 checkpoint with 3D-inflated patch embeddings; FlexiCT-3D starts from the full Phase~2 backbone. Values are mean with 95\% BCa bootstrap confidence intervals. $^{*}$ denotes significance under paired two-sided permutation test after Bonferroni correction ($\alpha=0.05/4=0.0125$). Best per row in bold.}
    \label{tab:supp_ablation_phase2}
    \footnotesize
    \resizebox{\textwidth}{!}{%
    \begin{tabular}{llccc}
    \toprule
    Task & Dataset (metric) & Random init & FlexiCT-2D & FlexiCT-3D \\
    \midrule
    Zero-shot disease classification
        & CT-RATE (AUC $\uparrow$)
        & 0.761 (0.750--0.765)
        & 0.789 (0.781--0.797)
        & \textbf{0.813 (0.807--0.820)}$^{*}$ \\
        & Merlin (AUC $\uparrow$)
        & 0.848 (0.835--0.859)
        & 0.853 (0.840--0.865)
        & \textbf{0.872 (0.862--0.882)}$^{*}$ \\
    Report retrieval
        & CT-RATE (Recall@5 $\uparrow$)
        & 0.318 (0.295--0.342)
        & 0.351 (0.335--0.379)
        & \textbf{0.378 (0.354--0.403)}$^{*}$ \\
        & Merlin (Recall@1, $N{=}32$, $\uparrow$)
        & 0.811 (0.734--0.886)
        & 0.865 (0.785--0.920)
        & \textbf{0.888 (0.888--1.000)} \\
    \bottomrule
    \end{tabular}%
    }
\end{table}

\begin{landscape}
{\scriptsize
\setlength{\LTcapwidth}{0.95\linewidth}
\renewcommand{\arraystretch}{1.12}
\begin{longtable}{@{}
>{\raggedright\arraybackslash}p{2.2cm}
>{\raggedright\arraybackslash}p{2.5cm}
>{\raggedright\arraybackslash}p{4.0cm}
>{\raggedright\arraybackslash}p{4.6cm}
>{\raggedright\arraybackslash}p{5.0cm}@{}}

\caption{Case-level exposure of downstream benchmarks relative to FlexiCT pretraining. 
For each evaluation benchmark, we record whether the evaluated CT images were included in any FlexiCT pretraining stage, and whether the corresponding downstream annotations were used as pretraining targets. Annotations include segmentation masks, class labels, staging or grading labels, tumor-size measurements, histology labels, deformation fields, and radiology reports. Benchmarks marked ``Yes'' in the evaluation-cases column represent in-domain transfer. 
Benchmarks marked ``No, held-out'' were excluded at the patient level before the relevant pretraining stage. For CT-RATE and Merlin, only training-split reports or derived captions were used during Phase 3 report-aligned pretraining, whereas validation or test reports were reserved for evaluation.}

\label{tab:supp_downstream_pretrain_overlap}\\

\toprule
Task family &
Downstream dataset &
Evaluation task &
Evaluation cases used in pretraining? &
Labels or reports used in pretraining? \\
\midrule
\endfirsthead

\multicolumn{5}{l}{\textit{Supplementary Table 23 continued}} \\
\toprule
Task family &
Downstream dataset &
Evaluation task &
Raw images used in pretraining? &
Labels or reports used in pretraining? \\
\midrule
\endhead

\midrule
\multicolumn{5}{r}{\textit{Continued on next page}} \\
\endfoot
\bottomrule
\endlastfoot

3D segmentation & KiTS23 & Kidney, mass, and tumor segmentation &
No, held-out evaluation split &
No segmentation labels \\

3D segmentation & WORD & Whole-abdominal organ segmentation &
No, held-out evaluation split &
No segmentation labels \\

3D segmentation & MSD-Liver & Liver and liver-tumor segmentation &
No, held-out evaluation split &
No segmentation labels \\

3D segmentation & MSD-Lung & Lung tumor segmentation &
No, held-out evaluation split &
No segmentation labels \\

3D segmentation & MSD-Pancreas & Pancreas and pancreatic-tumor segmentation &
No, held-out evaluation split &
No segmentation labels \\

3D segmentation & AutoPET II & Metabolically active tumor segmentation &
No, held-out evaluation split &
No segmentation labels \\

2D segmentation & TotalSegmentator V2 & Whole-body anatomical segmentation &
Yes &
No segmentation labels \\

2D segmentation & AMOS22 & Abdominal multi-organ segmentation, CT and MR subsets &
Yes &
No segmentation labels \\

Registration & AbdomenCTCT / Learn2Reg CT--CT & Training-free intra-modal abdominal registration &
No, held-out evaluation split &
No registration labels or deformation fields \\

Registration & AbdomenMRCT / Learn2Reg CT--MR & Training-free cross-modal abdominal registration &
No, held-out evaluation split &
No registration labels or deformation fields \\

\midrule

Classification & KiTS & Renal tumor subtyping from frozen features &
No, held-out evaluation split &
No class labels \\

Classification & DeepLesion & Eight-class lesion-site classification &
Yes &
No lesion-site labels \\

Classification & LUNA16 & Pulmonary nodule classification &
Yes &
No nodule labels \\

Classification & COVIDx-CT & Normal, pneumonia, and COVID-19 classification &
No, held-out evaluation split &
No COVID class labels \\

\midrule

tumor phenotype & NSCLC-Radiogenomics & T-stage retrieval &
No, held-out phenotype cohort &
No T-stage labels \\

tumor phenotype & NSCLC-Radiogenomics & T-stage linear probing &
No, held-out phenotype cohort &
No T-stage labels \\

tumor phenotype & NSCLC-Radiogenomics & LDA severity-gradient and diameter-residual analysis &
No, held-out phenotype cohort &
No T-stage, tumor-diameter, or radiogenomic labels \\

tumor phenotype & C4KC-KiTS & ISUP-grade retrieval &
No, held-out phenotype cohort &
No ISUP grade labels \\

tumor phenotype & C4KC-KiTS & ISUP-grade linear probing &
No, held-out phenotype cohort &
No ISUP grade labels \\

tumor phenotype & C4KC-KiTS & LDA severity-gradient and size-matched analysis &
No, held-out phenotype cohort &
No ISUP grade, tumor-size, or histology labels \\

\midrule

Vision--language & CT-RATE & Zero-shot multi-abnormality classification &
No, held-out validation split excluded from Phase 3 training &
Reports and CT-RATE-derived captions used only for Phase 3 training split \\

Vision--language & CT-RATE & Report retrieval &
No, held-out validation split excluded from Phase 3 training &
Reports and CT-RATE-derived captions used only for Phase 3 training split; held-out evaluation reports are used only at evaluation \\

Vision--language & Merlin & Zero-shot multi-abnormality classification &
No, held-out test split excluded from Phase 3 training &
Original Merlin reports used only for Phase 3 training split \\

Vision--language & Merlin & Report retrieval &
No, held-out test split excluded from Phase 3 training &
Original Merlin reports used only for Phase 3 training split; held-out evaluation reports are used only at evaluation \\

\end{longtable}
}
\end{landscape}


\bibliographystyle{plainnat}
\bibliography{references}

@article{voxelmorph,
  title={Voxelmorph: a learning framework for deformable medical image registration},
  author={Balakrishnan, Guha and Zhao, Amy and Sabuncu, Mert R and Guttag, John and Dalca, Adrian V},
  journal={IEEE transactions on medical imaging},
  volume={38},
  number={8},
  pages={1788--1800},
  year={2019},
  publisher={IEEE}
}

@article{voco2024,
  title={Large-scale 3d medical image pre-training with geometric context priors},
  author={Wu, Linshan and Zhuang, Jiaxin and Chen, Hao},
  journal={IEEE Transactions on Pattern Analysis and Machine Intelligence},
  year={2025},
  publisher={IEEE}
}

@article{spectre2024,
  title={Scaling Self-Supervised and Cross-Modal Pretraining for Volumetric CT Transformers},
  author={Claessens, Cris and Viviers, Christiaan and D'Amicantonio, Giacomo and Bondarev, Egor and van der Sommen, Fons},
  journal={arXiv preprint arXiv:2511.17209},
  year={2025}
}

@article{harvardctfm2024,
  title={Vision foundation models for computed tomography},
  author={Pai, Suraj and Hadzic, Ibrahim and Bontempi, Dennis and Bressem, Keno and Kann, Benjamin H and Fedorov, Andriy and Mak, Raymond H and Aerts, Hugo JWL},
  journal={arXiv preprint arXiv:2501.09001},
  year={2025}
}

@inproceedings{qiu2024mind,
  title={Mind your augmentation: The key to decoupling dense self-supervised learning},
  author={Qiu, Congpei and Zhang, Tong and Wu, Yanhao and Ke, Wei and Salzmann, Mathieu and S{\"u}sstrunk, Sabine},
  booktitle={The Twelfth International Conference on Learning Representations},
  year={2024}
}

@article{maninis2024tips,
  title={Tips: Text-image pretraining with spatial awareness},
  author={Maninis, Kevis-Kokitsi and Chen, Kaifeng and Ghosh, Soham and Karpur, Arjun and Chen, Koert and Xia, Ye and Cao, Bingyi and Salz, Daniel and Han, Guangxing and Dlabal, Jan and others},
  journal={arXiv preprint arXiv:2410.16512},
  year={2024}
}

@article{zhou2021ibot,
  title={ibot: Image bert pre-training with online tokenizer},
  author={Zhou, Jinghao and Wei, Chen and Wang, Huiyu and Shen, Wei and Xie, Cihang and Yuille, Alan and Kong, Tao},
  journal={arXiv preprint arXiv:2111.07832},
  year={2021}
}

@inproceedings{beyer2023flexivit,
  title={Flexivit: One model for all patch sizes},
  author={Beyer, Lucas and Izmailov, Pavel and Kolesnikov, Alexander and Caron, Mathilde and Kornblith, Simon and Zhai, Xiaohua and Minderer, Matthias and Tschannen, Michael and Alabdulmohsin, Ibrahim and Pavetic, Filip},
  booktitle={Proceedings of the IEEE/CVF Conference on Computer Vision and Pattern Recognition},
  pages={14496--14506},
  year={2023}
}

@article{darcet2023vision,
  title={Vision transformers need registers},
  author={Darcet, Timoth{\'e}e and Oquab, Maxime and Mairal, Julien and Bojanowski, Piotr},
  journal={arXiv preprint arXiv:2309.16588},
  year={2023}
}

@article{Fournier2016OnTK,
  title={On the Kozachenko-Leonenko entropy estimator},
  author={Nicolas Fournier and Sylvain Delattre},
  journal={arXiv: Statistics Theory},
  year={2016},
  url={https://api.semanticscholar.org/CorpusID:14948397}
}

@article{Yang2025Qwen3TR,
  title={Qwen3 technical report},
  author={Yang, An and Li, Anfeng and Yang, Baosong and Zhang, Beichen and Hui, Binyuan and Zheng, Bo and Yu, Bowen and Gao, Chang and Huang, Chengen and Lv, Chenxu and others},
  journal={arXiv preprint arXiv:2505.09388},
  year={2025}
}

@article{Heller2019TheSO,
  title={The state of the art in kidney and kidney tumor segmentation in contrast-enhanced CT imaging: Results of the KiTS19 Challenge},
  author={N. Heller and Fabian Isensee and Klaus Hermann Maier-Hein and Xiaoshuai Hou and Chunmei Xie and Fengyi Li and Yang Nan and Guangrui Mu and Zhiyong Lin and Miofei Han and Guang Yao and Yaozong Gao and Yao Zhang and Yixin Wang and Feng Hou and Jiawei Yang and Guangwei Xiong and Jiang Tian and Cheng Zhong and Jun Ma and Jack Rickman and Joshua Dean and Bethany Stai and Resha Tejpaul and Makinna Oestreich and Paul Blake and Heather Kaluzniak and Shaneabbas Raza and Joel E Rosenberg and Keenan Moore and Edward Walczak and Zachary Rengel and Zach Edgerton and Ranveer M.S. Vasdev and Matthew Peterson and Sean McSweeney and Sarah Peterson and Arveen A. Kalapara and Niranjan Jude Sathianathen and Christopher J. Weight and Nikolaos Papanikolopoulos},
  journal={Medical image analysis},
  year={2019},
  volume={67},
  pages={
          101821
        },
  url={https://api.semanticscholar.org/CorpusID:208547601}
}

@article{Siebert2024ConvexAdamSD,
  title={ConvexAdam: Self-Configuring Dual-Optimization-Based 3D Multitask Medical Image Registration},
  author={Hanna Siebert and Christoph Gro{\ss}br{\"o}hmer and Lasse Hansen and Mattias P. Heinrich},
  journal={IEEE Transactions on Medical Imaging},
  year={2024},
  volume={44},
  pages={738-748},
  url={https://api.semanticscholar.org/CorpusID:272693057}
}

@article{Cardoso2022MONAIAO,
  title={MONAI: An open-source framework for deep learning in healthcare},
  author={Manuel Jorge Cardoso and Wenqi Li and Richard Brown and Nic Ma and Eric Kerfoot and Yiheng Wang and Benjamin Murrey and Andriy Myronenko and Can Zhao and Dong Yang and V. Nath and Yufan He and Ziyue Xu and Ali Hatamizadeh and Wenjie Zhu and Yun Liu and Mingxin Zheng and Yucheng Tang and Isaac Yang and Michael Zephyr and Behrooz Hashemian and Sachidanand Alle and Mohammad Zalbagi Darestani and Charles. Budd and Marc Modat and Tom Kamiel Magda Vercauteren and Guotai Wang and Yiwen Li and Yipeng Hu and Yunguan Fu and Benjamin L. Gorman and Hans J. Johnson and Brad W. Genereaux and Barbaros Selnur Erdal and Vikash Gupta and Andr{\'e}s Diaz-Pinto and Andre Dourson and Lena Maier-Hein and Paul F. Jaeger and Michael Baumgartner and Jayashree Kalpathy-Cramer and Mona G. Flores and Justin S. Kirby and Lee Alex Donald Cooper and Holger R. Roth and Daguang Xu and David Bericat and Ralf O. Floca and S. Kevin Zhou and Haris Shuaib and Keyvan Farahani and Klaus H. Maier-Hein and Stephen Aylward and Prerna Dogra and S{\'e}bastien Ourselin and Andrew Feng},
  journal={ArXiv},
  year={2022},
  volume={abs/2211.02701},
  url={https://api.semanticscholar.org/CorpusID:253383958}
}

@article{Zhang2025Qwen3EA,
  title={Qwen3 Embedding: Advancing Text Embedding and Reranking Through Foundation Models},
  author={Yanzhao Zhang and Mingxin Li and Dingkun Long and Xin Zhang and Huan Lin and Baosong Yang and Pengjun Xie and An Yang and Dayiheng Liu and Junyang Lin and Fei Huang and Jingren Zhou},
  journal={ArXiv},
  year={2025},
  volume={abs/2506.05176},
  url={https://api.semanticscholar.org/CorpusID:279243736}
}

@article{ma2024automatic,
  title={Automatic organ and pan-cancer segmentation in abdomen ct: the flare 2023 challenge},
  author={Ma, Jun and Zhang, Yao and Gu, Song and Ge, Cheng and Wang, Ershuai and Zhou, Qin and Huang, Ziyan and Lyu, Pengju and He, Jian and Wang, Bo},
  journal={arXiv preprint arXiv:2408.12534},
  year={2024}
}

@article{pai2024foundation,
  title={Foundation model for cancer imaging biomarkers},
  author={Pai, Suraj and Bontempi, Dennis and Hadzic, Ibrahim and Prudente, Vasco and Soka{\v{c}}, Mateo and Chaunzwa, Tafadzwa L and Bernatz, Simon and Hosny, Ahmed and Mak, Raymond H and Birkbak, Nicolai J and others},
  journal={Nature machine intelligence},
  volume={6},
  number={3},
  pages={354--367},
  year={2024},
  publisher={Nature Publishing Group UK London}
}

@article{curia2024,
  title={Curia: A Multi-Modal Foundation Model for Radiology},
  author={Dancette, Corentin and Khlaut, Julien and Saporta, Antoine and Philippe, Helene and Ferreres, Elodie and Callard, Baptiste and Danielou, Th{\'e}o and Alberge, L{\'e}o and Machado, L{\'e}o and Tordjman, Daniel and others},
  journal={arXiv preprint arXiv:2509.06830},
  year={2025}
}

@article{merlin2024,
  title={Merlin: a computed tomography vision--language foundation model and dataset},
  author={Blankemeier, Louis and Kumar, Ashwin and Cohen, Joseph Paul and Liu, Jiaming and Liu, Longchao and Van Veen, Dave and Gardezi, Syed Jamal Safdar and Yu, Hongkun and Paschali, Magdalini and Chen, Zhihong and others},
  journal={Nature},
  pages={1--11},
  year={2026},
  publisher={Nature Publishing Group UK London}
}

@article{ctclip2024,
  title={Generalist foundation models from a multimodal dataset for 3D computed tomography},
  author={Hamamci, Ibrahim Ethem and Er, Sezgin and Wang, Chenyu and Almas, Furkan and Simsek, Ayse Gulnihan and Esirgun, Sevval Nil and Dogan, Irem and Durugol, Omer Faruk and Hou, Benjamin and Shit, Suprosanna and others},
  journal={Nature Biomedical Engineering},
  pages={1--19},
  year={2026},
  publisher={Nature Publishing Group UK London}
}

@article{primus2024,
  title={Primus: Enforcing attention usage for 3d medical image segmentation},
  author={Wald, Tassilo and Roy, Saikat and Isensee, Fabian and Ulrich, Constantin and Ziegler, Sebastian and Trofimova, Dasha and Stock, Raphael and Baumgartner, Michael and K{\"o}hler, Gregor and Maier-Hein, Klaus},
  journal={arXiv preprint arXiv:2503.01835},
  year={2025}
}

@article{coplpri2024,
  title={Comprehensive language-image pre-training for 3D medical image understanding},
  author={Wald, Tassilo and Hamamci, Ibrahim Ethem and Gao, Yuan and Bond-Taylor, Sam and Sharma, Harshita and Ilse, Maximilian and Lo, Cynthia and Melnichenko, Olesya and Schwaighofer, Anton and Codella, Noel CF and others},
  journal={arXiv preprint arXiv:2510.15042},
  year={2025}
}

@article{dinov2,
  title={Dinov2: Learning robust visual features without supervision},
  author={Oquab, Maxime and Darcet, Timoth{\'e}e and Moutakanni, Th{\'e}o and Vo, Huy and Szafraniec, Marc and Khalidov, Vasil and Fernandez, Pierre and Haziza, Daniel and Massa, Francisco and El-Nouby, Alaaeldin and others},
  journal={arXiv preprint arXiv:2304.07193},
  year={2023}
}

@article{dinov3,
  title={Dinov3},
  author={Sim{\'e}oni, Oriane and Vo, Huy V and Seitzer, Maximilian and Baldassarre, Federico and Oquab, Maxime and Jose, Cijo and Khalidov, Vasil and Szafraniec, Marc and Yi, Seungeun and Ramamonjisoa, Micha{\"e}l and others},
  journal={arXiv preprint arXiv:2508.10104},
  year={2025}
}

@article{biomedclip2023,
  title={Biomedclip: a multimodal biomedical foundation model pretrained from fifteen million scientific image-text pairs},
  author={Zhang, Sheng and Xu, Yanbo and Usuyama, Naoto and Xu, Hanwen and Bagga, Jaspreet and Tinn, Robert and Preston, Sam and Rao, Rajesh and Wei, Mu and Valluri, Naveen and others},
  journal={arXiv preprint arXiv:2303.00915},
  year={2023}
}

@article{vit2020,
  title={An image is worth 16x16 words: Transformers for image recognition at scale},
  author={Dosovitskiy, Alexey and Beyer, Lucas and Kolesnikov, Alexander and Weissenborn, Dirk and Zhai, Xiaohua and Unterthiner, Thomas and Dehghani, Mostafa and Minderer, Matthias and Heigold, Georg and Gelly, Sylvain and others},
  journal={arXiv preprint arXiv:2010.11929},
  year={2020}
}

@article{nnunet2021,
  title={nnU-Net: a self-configuring method for deep learning-based biomedical image segmentation},
  author={Isensee, Fabian and Jaeger, Paul F and Kohl, Simon AA and Petersen, Jens and Maier-Hein, Klaus H},
  journal={Nature methods},
  volume={18},
  number={2},
  pages={203--211},
  year={2021},
  publisher={Nature Publishing Group US New York}
}

@inproceedings{dinoreg2024,
  title={Dino-reg: General purpose image encoder for training-free multi-modal deformable medical image registration},
  author={Song, Xinrui and Xu, Xuanang and Yan, Pingkun},
  booktitle={International Conference on Medical Image Computing and Computer-Assisted Intervention},
  pages={608--617},
  year={2024},
  organization={Springer}
}

@misc{kits2023,
  author       = {Heller, Nicholas and Isensee, Fabian and Tejpau, Resha and Wood, Andrew and Papanikolopoulos, Nikolaos and Weight, Christopher},
  title        = {2023 Kidney and Kidney Tumor Segmentation Challenge},
  year         = {2023},
  howpublished = {International Conference on Medical Image Computing and Computer Assisted Intervention (MICCAI 2023)},
  publisher    = {Zenodo},
  doi          = {10.5281/zenodo.7840134},
  url          = {https://doi.org/10.5281/zenodo.7840134}
}

@article{word2022,
  title={WORD: A large scale dataset, benchmark and clinical applicable study for abdominal organ segmentation from CT image},
  author={Luo, Xiangde and Liao, Wenjun and Xiao, Jianghong and Chen, Jieneng and Song, Tao and Zhang, Xiaofan and Li, Kang and Metaxas, Dimitris N and Wang, Guotai and Zhang, Shaoting},
  journal={Medical Image Analysis},
  volume={82},
  pages={102642},
  year={2022},
  publisher={Elsevier}
}

@article{autopet2022,
  title={A whole-body FDG-PET/CT dataset with manually annotated tumor lesions},
  author={Gatidis, Sergios and Hepp, Tobias and Fr{\"u}h, Marcel and La Foug{\`e}re, Christian and Nikolaou, Konstantin and Pfannenberg, Christina and Sch{\"o}lkopf, Bernhard and K{\"u}stner, Thomas and Cyran, Clemens and Rubin, Daniel},
  journal={Scientific Data},
  volume={9},
  number={1},
  pages={601},
  year={2022},
  publisher={Nature Publishing Group UK London}
}

@article{amos2022,
  title={Amos: A large-scale abdominal multi-organ benchmark for versatile medical image segmentation},
  author={Ji, Yuanfeng and Bai, Haotian and Ge, Chongjian and Yang, Jie and Zhu, Ye and Zhang, Ruimao and Li, Zhen and Zhanng, Lingyan and Ma, Wanling and Wan, Xiang and others},
  journal={Advances in neural information processing systems},
  volume={35},
  pages={36722--36732},
  year={2022}
}

@article{wasserthal2023totalsegmentator,
  title={TotalSegmentator: robust segmentation of 104 anatomic structures in CT images},
  author={Wasserthal, Jakob and Breit, Hanns-Christian and Meyer, Manfred T and Pradella, Maurice and Hinck, Daniel and Sauter, Alexander W and Heye, Tobias and Boll, Daniel T and Cyriac, Joshy and Yang, Shan and others},
  journal={Radiology: Artificial Intelligence},
  volume={5},
  number={5},
  pages={e230024},
  year={2023},
  publisher={Radiological Society of North America}
}

@article{deeplesion2018,
  title={DeepLesion: automated mining of large-scale lesion annotations and universal lesion detection with deep learning},
  author={Yan, Ke and Wang, Xiaosong and Lu, Le and Summers, Ronald M},
  journal={Journal of medical imaging},
  volume={5},
  number={3},
  pages={036501--036501},
  year={2018},
  publisher={Society of Photo-Optical Instrumentation Engineers}
}

@article{luna16,
  title={Validation, comparison, and combination of algorithms for automatic detection of pulmonary nodules in computed tomography images: the LUNA16 challenge},
  author={Setio, Arnaud Arindra Adiyoso and Traverso, Alberto and De Bel, Thomas and Berens, Moira SN and Van Den Bogaard, Cas and Cerello, Piergiorgio and Chen, Hao and Dou, Qi and Fantacci, Maria Evelina and Geurts, Bram and others},
  journal={Medical image analysis},
  volume={42},
  pages={1--13},
  year={2017},
  publisher={Elsevier}
}

@article{covidxct2021,
  title={Covid-net ct-2: Enhanced deep neural networks for detection of covid-19 from chest ct images through bigger, more diverse learning},
  author={Gunraj, Hayden and Sabri, Ali and Koff, David and Wong, Alexander},
  journal={Frontiers in Medicine},
  volume={8},
  pages={729287},
  year={2022},
  publisher={Frontiers Media SA}
}

@article{ctrate2024,
  title={Generalist foundation models from a multimodal dataset for 3D computed tomography},
  author={Hamamci, Ibrahim Ethem and Er, Sezgin and Wang, Chenyu and Almas, Furkan and Simsek, Ayse Gulnihan and Esirgun, Sevval Nil and Dogan, Irem and Durugol, Omer Faruk and Hou, Benjamin and Shit, Suprosanna and others},
  journal={Nature Biomedical Engineering},
  pages={1--19},
  year={2026},
  publisher={Nature Publishing Group UK London}
}

@article{nsclcradiogenomics,
  title={A radiogenomic dataset of non-small cell lung cancer},
  author={Bakr, Shaimaa and Gevaert, Olivier and Echegaray, Sebastian and Ayers, Kelsey and Zhou, Mu and Shafiq, Majid and Zheng, Hong and Benson, Jalen Anthony and Zhang, Weiruo and Leung, Ann NC and others},
  journal={Scientific data},
  volume={5},
  number={1},
  pages={180202},
  year={2018},
  publisher={Nature Publishing Group}
}

@inproceedings{clip2021,
  title={Learning transferable visual models from natural language supervision},
  author={Radford, Alec and Kim, Jong Wook and Hallacy, Chris and Ramesh, Aditya and Goh, Gabriel and Agarwal, Sandhini and Sastry, Girish and Askell, Amanda and Mishkin, Pamela and Clark, Jack and others},
  booktitle={International conference on machine learning},
  pages={8748--8763},
  year={2021},
  organization={PmLR}
}

@article{smith2019trends,
  title={Trends in use of medical imaging in US health care systems and in Ontario, Canada, 2000-2016},
  author={Smith-Bindman, Rebecca and Kwan, Marilyn L and Marlow, Emily C and Theis, Mary Kay and Bolch, Wesley and Cheng, Stephanie Y and Bowles, Erin JA and Duncan, James R and Greenlee, Robert T and Kushi, Lawrence H and others},
  journal={Jama},
  volume={322},
  number={9},
  pages={843--856},
  year={2019}
}

@article{brenner2007computed,
  title={Computed tomography—an increasing source of radiation exposure},
  author={Brenner, David J and Hall, Eric J},
  journal={New England journal of medicine},
  volume={357},
  number={22},
  pages={2277--2284},
  year={2007},
  publisher={Mass Medical Soc}
}

@article{power2016computed,
  title={Computed tomography and patient risk: Facts, perceptions and uncertainties},
  author={Power, Stephen P and Moloney, Fiachra and Twomey, Maria and James, Karl and O’Connor, Owen J and Maher, Michael M},
  journal={World journal of radiology},
  volume={8},
  number={12},
  pages={902},
  year={2016}
}

@article{moor2023foundation,
  title={Foundation models for generalist medical artificial intelligence},
  author={Moor, Michael and Banerjee, Oishi and Abad, Zahra Shakeri Hossein and Krumholz, Harlan M and Leskovec, Jure and Topol, Eric J and Rajpurkar, Pranav},
  journal={Nature},
  volume={616},
  number={7956},
  pages={259--265},
  year={2023},
  publisher={Nature Publishing Group UK London}
}

@article{willemink2022toward,
  title={Toward foundational deep learning models for medical imaging in the new era of transformer networks},
  author={Willemink, Martin J and Roth, Holger R and Sandfort, Veit},
  journal={Radiology: Artificial Intelligence},
  volume={4},
  number={6},
  pages={e210284},
  year={2022},
  publisher={Radiological Society of North America}
}

@article{chen2024towards,
  title={Towards a general-purpose foundation model for computational pathology},
  author={Chen, Richard J and Ding, Tong and Lu, Ming Y and Williamson, Drew FK and Jaume, Guillaume and Song, Andrew H and Chen, Bowen and Zhang, Andrew and Shao, Daniel and Shaban, Muhammad and others},
  journal={Nature medicine},
  volume={30},
  number={3},
  pages={850--862},
  year={2024},
  publisher={Nature Publishing Group US New York}
}

@article{vorontsov2024foundation,
  title={A foundation model for clinical-grade computational pathology and rare cancers detection},
  author={Vorontsov, Eugene and Bozkurt, Alican and Casson, Adam and Shaikovski, George and Zelechowski, Michal and Severson, Kristen and Zimmermann, Eric and Hall, James and Tenenholtz, Neil and Fusi, Nicolo and others},
  journal={Nature medicine},
  volume={30},
  number={10},
  pages={2924--2935},
  year={2024},
  publisher={Nature Publishing Group US New York}
}

@article{zhou2023foundation,
  title={A foundation model for generalizable disease detection from retinal images},
  author={Zhou, Yukun and Chia, Mark A and Wagner, Siegfried K and Ayhan, Murat S and Williamson, Dominic J and Struyven, Robbert R and Liu, Timing and Xu, Moucheng and Lozano, Mateo G and Woodward-Court, Peter and others},
  journal={Nature},
  volume={622},
  number={7981},
  pages={156--163},
  year={2023},
  publisher={Nature Publishing Group UK London}
}

@inproceedings{caron2021emerging,
  title={Emerging properties in self-supervised vision transformers},
  author={Caron, Mathilde and Touvron, Hugo and Misra, Ishan and J{\'e}gou, Herv{\'e} and Mairal, Julien and Bojanowski, Piotr and Joulin, Armand},
  booktitle={Proceedings of the IEEE/CVF international conference on computer vision},
  pages={9650--9660},
  year={2021}
}

@article{ardila2019end,
  title={End-to-end lung cancer screening with three-dimensional deep learning on low-dose chest computed tomography},
  author={Ardila, Diego and Kiraly, Atilla P and Bharadwaj, Sujeeth and Choi, Bokyung and Reicher, Joshua J and Peng, Lily and Tse, Daniel and Etemadi, Mozziyar and Ye, Wenxing and Corrado, Greg and others},
  journal={Nature medicine},
  volume={25},
  number={6},
  pages={954--961},
  year={2019},
  publisher={Nature Publishing Group US New York}
}





\end{document}